\documentclass[a4paper,12pt,pdftex]{article}
\usepackage[pdftex]{graphicx} 
\graphicspath{{./FIGS/}}
\usepackage{lineno}
\usepackage{soul} 
\usepackage[numbers,sort&compress]{natbib}      
\usepackage[nottoc]{tocbibind}  

\usepackage[
    letterpaper,
    top=0.75in, 
    bottom=1.0in, 
    right=0.75in, 
    left=0.75in, 
]{geometry}
\usepackage[utf8]{inputenc} 
\usepackage[T1]{fontenc} 
\usepackage[english]{babel}
\usepackage{microtype} 
\usepackage{lmodern}

\usepackage{tikz}
\usetikzlibrary{positioning, calc, decorations.pathreplacing}
\usetikzlibrary{arrows.meta}

\usepackage{amsmath,amsfonts,units,amssymb,amsthm,mathtools,bigstrut,amstext}
\usepackage{empheq}

\usepackage[colorlinks=true]{hyperref} 
\AtBeginDocument{%
\hypersetup{
    linkcolor=blue,    
    filecolor=magenta,      
    urlcolor=blue,     
}
}
\hypersetup{citecolor=blue}
\numberwithin{equation}{subsection}

\usepackage{subcaption}

\usepackage{algorithm} 
\usepackage{algorithmicx} 
\usepackage{algpseudocode}

\algnewcommand\algorithmicinput{\textbf{Input:}}
\algnewcommand\Input{\item[\algorithmicinput]}
\usepackage{dsfont}
\def\del{\partial} %
\def\half{\frac{1}{2}} 
\newcommand{\eq}[1]{\begin{equation}\label{#1}}
\newcommand{\en}{\end{equation}}
\def\nref#1{(\ref{#1})}
\def\RR{\mathbb{R}}
\def\EE{\mathbb{E}}
\def\One{\mathds{1}}
\def\inv{^{-1}}
\def\scl{\mathcal{L}} 
\def\calB{\mathcal{B}}
\def\calD{\mathcal{D}}
\def\calK{\mathcal{K}}
\def\calN{\mathcal{N}}
\def\vvec{\texttt{vec}} 
\def\up#1{^{(#1)}}%
\def\limit{\lim}
\def\diag{\texttt{diag}}
\usepackage{bm}
\usepackage{lscape}
\usepackage{rotating}
\usepackage[]{geometry}
\usepackage{multicol}
\newtheorem{example}{Example}[section]
\usepackage{circledsteps}

\usepackage[group-separator={,}]{siunitx}[=v2]
\ifdefined\unit\else
  \NewDocumentCommand\unit{m}{\si{#1}}
\fi
\usepackage{nicematrix}
\usepackage{booktabs}
\usepackage[labelfont=bf]{caption} 
\usepackage{cancel}

\usepackage{mathbbol}
\usepackage{arydshln}
\usepackage{gensymb}
\usepackage{url}
\usepackage{bbm}

\title{The Numerical Linear Algebra of Large Language Models}

\author{Abdelkader  Baggag%
\thanks{
    Qatar Computing Research Institute, Hamad Bin Khalifa University, HBKU Research Complex, Doha, Qatar.
    e-mail: \texttt{abaggag@hbku.edu.qa}, \url{https://www.linkedin.com/in/abdelkader-baggag-94a08664/}.}
  \and
  Yousef Saad%
  \thanks{Dept. Computer Science and Engineering, University of Minnesota,
  200 Union Street S.E., Minneapolis, MN, USA.
    e-mail: \texttt{saad@umn.edu}, \url{www.cs.umn.edu/~saad}. Work funded by NSF.}}

\usepackage[normalem]{ulem}
\usepackage{soul,xcolor} 

\begin{document}
\maketitle

\begin{abstract}
  Numerical Linear Algebra (NLA) has consistently played a vital role in
  advancing science by providing tools to solve fundamental problems encountered
  in scientific and engineering applications. Over the decades, it has
  continually evolved to meet the demands driven by successive waves of
  scientific discovery.  For instance, during the 1950s and 1960s, substantial
  efforts were devoted to developing methods for solving eigenvalue problems
  that emerged from the rapidly growing field of aerodynamics.
    This led to the discovery of the LR and QR algorithms.  Later
  the attention turned to the solution of sparse linear systems that were common in
  applications like computational aerodynamics.  Today we are experiencing yet
  another wave of major scientific advancement and NLA is once more at the
  heart of its development.  This Machine Learning (ML) wave is proving to be
  utterly disruptive in science and engineering.  Many tools in ML particularly
  Large Language Models (LLMs) are grounded in matrix and tensor methods.  As
  we are approaching Artificial General Intelligence (AGI), it is clear that
  matrix methods will be called to play an even more significant role.
  For the numerical linear practitioner the speed of the current change
  makes it particularly   challenging to adapt. 
  
  This is a survey article that centers on machine learning techniques,
  with a particular focus on large language models. It has two main
  objectives. The first is to clarify the core concepts behind Large Language
  Models in a manner accessible to specialists in numerical methods. The second
  is to examine the key Numerical Linear Algebra concepts employed by LLM
  techniques, while also highlighting several significant recent contributions of NLA to the field.
\end{abstract}


\bigskip

\paragraph*{MSCcodes:}
65-02, 65F30,  65K10, 68T01, 68W25, 90C06, 90C15, 90C30

\tableofcontents

\setstcolor{red} 
    
\medskip
 
\section{Introduction and historical perspective}
When the first electronic computers appeared scientists were confronted with the
question as to whether or not these machines could exhibit human-like
intelligence.  Thus, in 1950 Turing designed a test (the `Turing
test') to answer this very question \cite{Turing50}.  In the more than seven decades since, the
field of Artificial Intelligence (AI) has witnessed several bursts of
significant advancement.  In particular, there was a surge of
excitement surrounding AI research during the 1970s and 1980s. However, it
became evident that the promises made in the field were too ambitious for the
time, leading to a subsequent negative reaction and a decline in interest.
Fortunately, AI resurfaced with force beginning around 2010 thanks in large part to 
improvements in computing power, and since the mid-2010s, it has undergone an astoundingly fast ascent, penetrating most scientific disciplines.   

\subsection{Major factors of the advancement of AI}
\begin{figure}[htb]
\centerline{
\includegraphics[width=0.8\textwidth]{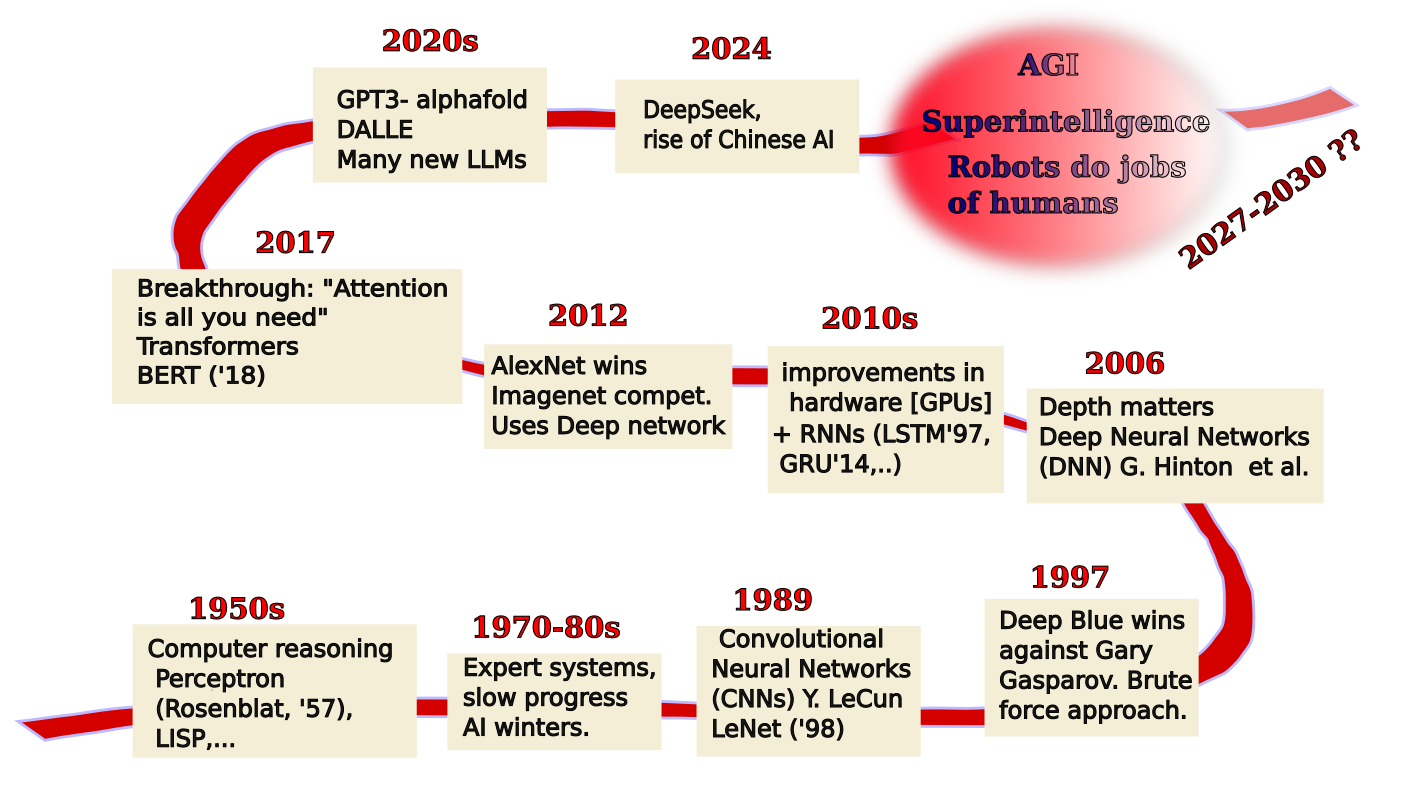}}
\caption{Major milestones in the evolution of artificial intelligence. The timeline highlights selected developments that shaped the field, from early perceptrons and expert systems to deep neural networks, Transformers, and the current era of large language models and generative AI. It is intended as a schematic historical overview rather than an exhaustive chronology.}
\label{fig:evol}
\end{figure}

The fast progress of AI was enabled by a confluence of synergistic factors and it is
worthwhile to list the main milestones of this success story,
see Figure~\ref{fig:evol} for a depiction of the main milestones in the progression of AI.
Neural networks
were invented in the late 1950s \cite{Rosenblatt58} with much of the real
foundational work carried out in the 1990s, see, e.g.,
\cite{suleyman2023,aschenbrenner2024situational,aggarwal23-book,DNN_overview15,DNN_survey17,DNN_history18,AI_brief_History_21} 
for a historical perspective.  It was soon realized that while the approach
could be successful, the training costs were potentially prohibitive.  Starting
around 2010, Neural Networks reappeared and spread quickly thanks in part to the
emergence of cheap computing power made possible by Graphical Processing Units
(GPUs).  Looking back, it is fascinating to see how the potential of GPUs,
originally developed as inexpensive boards for gaming and high-end graphics, was
gradually realized.  The second factor was the improvements of the models
brought about by `Deep Neural Networks' (DNNs).  Deep neural networks use a
large number of hidden layers (hence the term ``deep''), allowing them to learn
complex representations of data. The motivation for DNNs is that the models
employ a large number of parameters to learn hierarchical features, making them
capable of representing and learning more complex patterns in data. However,
this also means that the models became more complex and expensive to train.  In
this regard, the idea of exploiting backward differentiation or back-propagation
\cite{rumelhart1986learning} played a major role in facilitating the training of
multi-layer neural networks.  This important discovery  made
it a simple task to train complex models by providing easy-to-use computational
codes such as TensorFlow \cite{TensorFlow16} and PyTorch \cite{PyTorch}. There
is no doubt that the availability of these codes played a major role in the
proliferation of  AI and in its stunning progression.

\subsection{The big AI wave and a comparison with chip manufacturing} 
A number of recent publications highlight the extraordinary progress made
by AI in recent years, see, e.g., 
\cite{aschenbrenner2024situational,suleyman2023,AI_brief_History_21,DNN_history18}.  Most of
these works stress the exponential nature of the progress made by AI with
a particularly impressive acceleration in the past 6-7 years.  The main
synergistic factors for this advancement are the progress in hardware, openness of
the field, and a massive community of researchers who contribute ideas and software.
These works also raise concerns about the unprecedented risks posed
by AI's unusually fast pace of development.  For example, an emphasis of
\cite{suleyman2023} relates to \emph{containment}, specifically ensuring that
the technology does not spiral out of control.

New technologies often display a similar exponential rate of progress, with a slow adoption at the beginning,
then very rapid progress, followed by a slower behavior due to limits encountered or just saturation.
What is different with AI is its unusually fast rate of progress.  
In this regard,
it is  interesting to draw a comparison with the remarkable progress made in
another field over the past few decades: chip-making.
In this context, Moore's Law \cite{MooresLaw65}, formulated in 1965, 
predicted the exponential progress of chip manufacturing over the following decades.
\emph{Moore's Law} predicted that the number  of transistors that can be placed on a chip will double every two years.
The prediction was fairly accurate until recent years where device manufacturing
started to confront the limits imposed by physics at the nanoscale.  We can ask
the question: Can an analogous observation be made for AI?
     
\begin{figure}[htb] 
  \centerline{\includegraphics[width=0.80\textwidth]{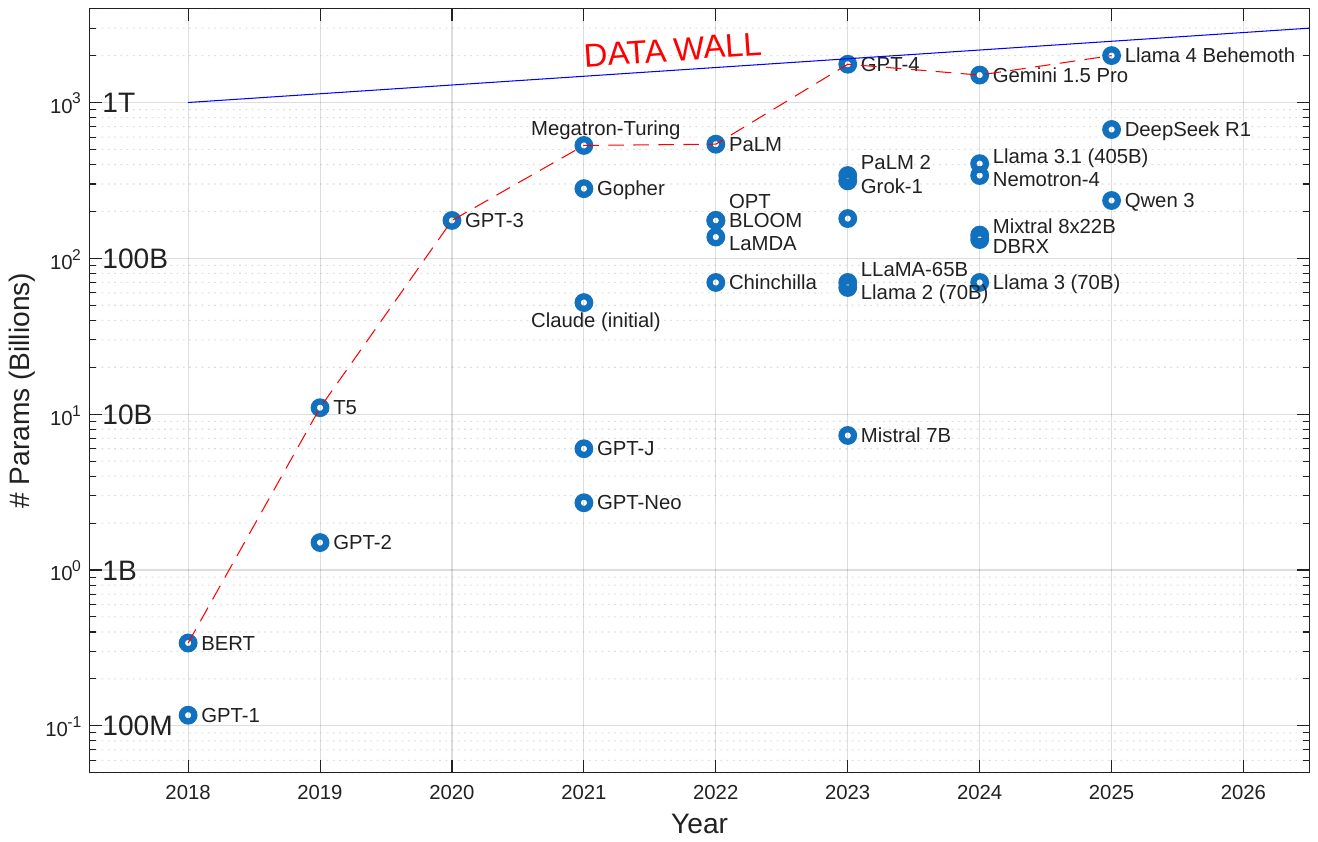}} 
  \caption{Growth of parameter sizes in large language models. The plot shows the rapid increase in parameter counts across representative models over time, from early Transformer-era systems to recent frontier models. The horizontal reference lines and the indicated ``data wall'' emphasize that simple parameter scaling cannot continue indefinitely and may eventually be constrained by data availability and practical training considerations.}
  \label{fig:MooreAI}
\end{figure}
In Artificial Intelligence we can look at the number of parameters in large language models (LLMs) to see if a similar
law can be stipulated.  When considering only the GPT models, the number of parameters
was 177M for GPT-1 (2018), 1.5B for GPT-2 (2019),
175B for GPT-3 (2020), and 1.7T (estimate) for GPT-4 (2023).   
The progression using these four  points suggests
that the number of parameters  \emph{doubles every $\approx$  4.6 months},
which translates to a  \emph{factor of 10 every $\approx$  15 months},  a much faster rate than that
of Moore's Law. It is clear that this rate cannot be sustained.
In fact Figure~\ref{fig:MooreAI}  shows that the number of parameters is already hitting a limit.
The number of parameters is somewhat linked to the corpus of data used, which itself is limited
so it may actually not increase much from current levels. 

One clear distinction that sets chipmaking and AI apart is that computer hardware
technology is extremely competitive and therefore highly protected, whereas AI is
more or less open (so far), allowing contributions of ideas from various horizons. This is a
case where the `bazaar' model, which characterized the Linux bottom-up approach, is
winning over the `Cathedral' top-down model \cite{BazaarBook}.  Such an  environment
is a blessing for research/science but it is also a curse for governments and
law-enforcement agencies as it facilitates misuses of the technology, making
\emph{containment} a necessity.

\begin{figure}[htb]
\centerline{
\includegraphics[width=0.9\textwidth]{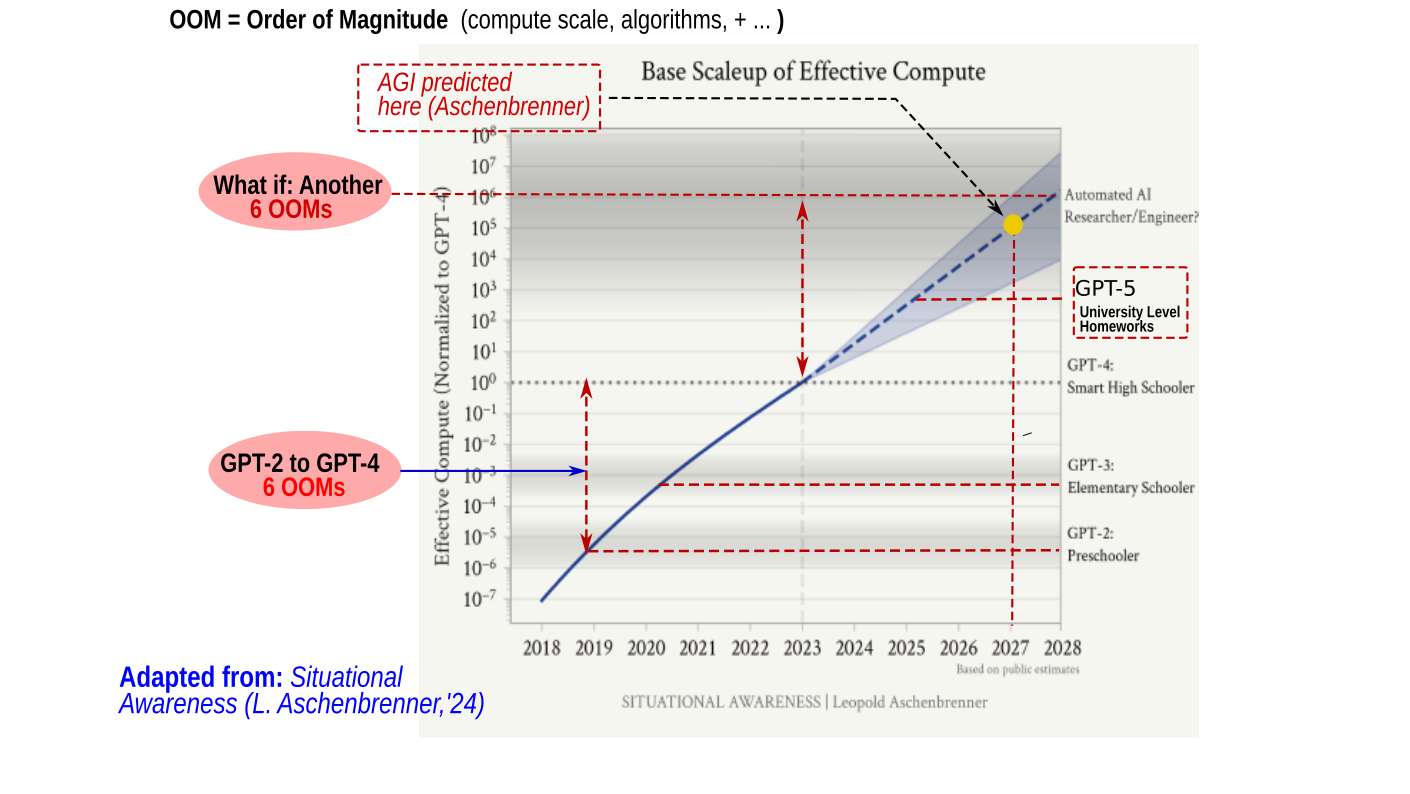}}
\caption{Order-of-magnitude argument for recent AI progress, adapted from \cite{aschenbrenner2024situational}. The figure illustrates the claim that advances in effective compute, algorithmic efficiency, and related ``unhobbling'' factors may combine multiplicatively, leading to rapid capability gains over relatively short time scales. It is included as a schematic argument about scaling trends, not as a precise predictive model.}
\label{fig:OOMa}
\end{figure}

`Artificial General Intelligence' (AGI)
refers to an autonomous AI system that can outperform humans at most
economically valuable work without needing task-specific training. In his 2024
article `Situational Awareness,' Leopold Aschenbrenner
\cite{aschenbrenner2024situational} argued that exponential scaling in computing
power and algorithmic efficiency, and other factors, make the arrival of AGI
highly plausible by 2027–2028.
The graph in Figure~\ref{fig:OOMa} is modified
from \cite{aschenbrenner2024situational}.  The argument made in the plot is that
GPT-2 to GPT-4 took us from pre-schooler to smart high-schooler abilities in
four years from 2019 to 2023. In these four years the author estimated that there
was an advance of six Orders Of Magnitude (OOMs) that combines computing power,
algorithmic efficiency, and what he terms `unhobbling' factors.  Assuming
another 6 OOMs in the 4 years from 2023 to 2027 the author argues that `we should
expect another preschooler-to-high-schooler-sized qualitative jump by
2027'.
The same article went on to discuss the sequel to AGI, namely 
Artificial Super Intelligence (ASI),  a hypothetical future stage of AI
defined as a superior intellect exceeding human capabilities in
creativity, decision-making, and general problem-solving across all domains.
The combined forces of massive
financial investments, intense research focus based on exploiting AGI, and national security motivations strongly suggest that   the emergence of ASI is unavoidable.

\begin{figure}[htb]
\centerline{
\includegraphics[width=0.6\textwidth]{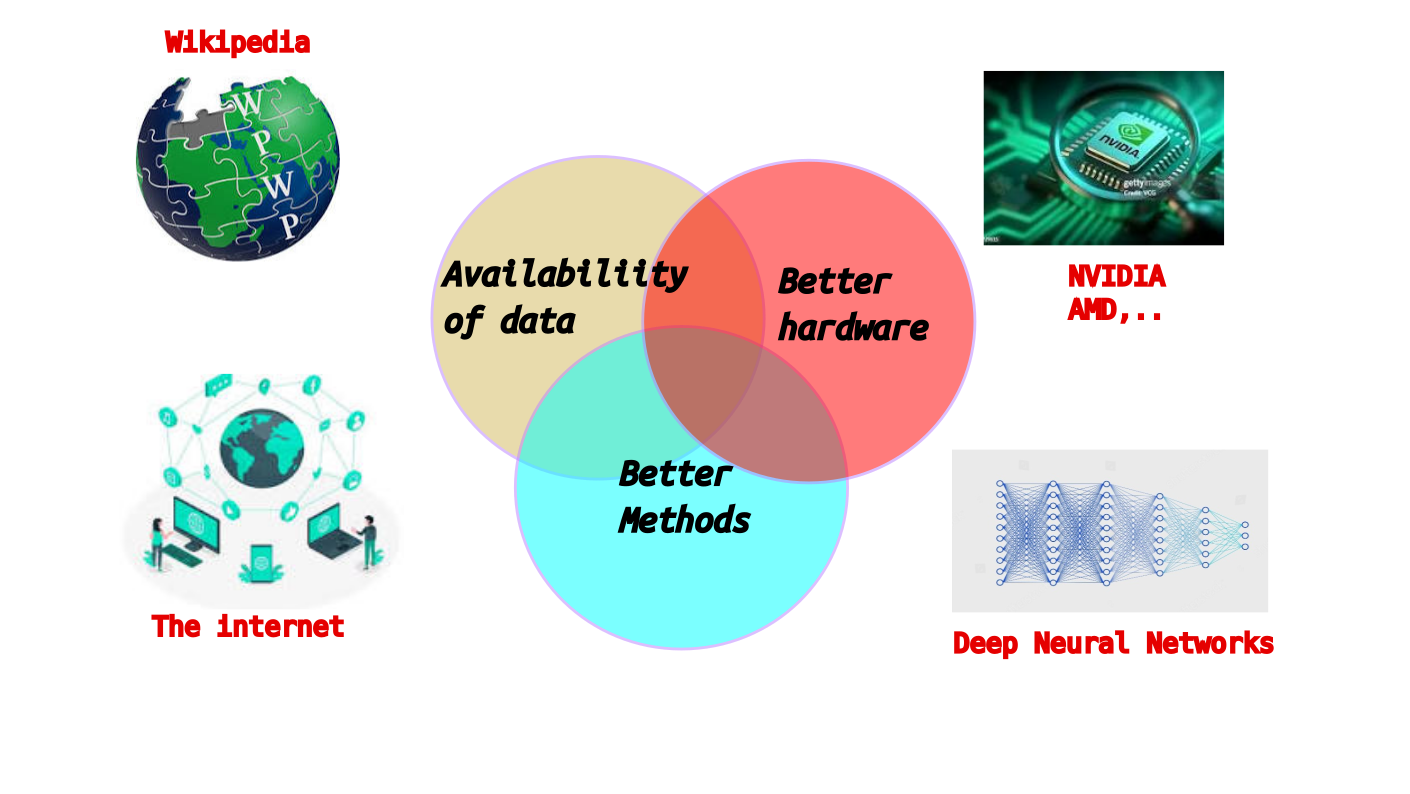}}
\caption{Main ingredients of the recent AI wave. The schematic emphasizes the interaction among three mutually reinforcing factors: improved hardware, the availability of large amounts of data, and advances in methods and algorithms. Together, these ingredients explain much of the rapid progress of modern deep learning and large language models.}
\label{fig:ingredients}
\end{figure}

\subsection{Main ingredients of Deep-Learning}\label{sec:ingr}
As was explained earlier, the recent wave that propelled AI resulted from the combination of
improved hardware (GPUs), the availability of data, and  big progress made in methods and algorithms
(Figure \ref{fig:ingredients}).
Among the `methods and algorithms'
we list  optimization first, although this does not imply any ranking in
importance relative to the other ingredients. 
The most time-consuming part in the development of an LLM
lies in its training phase. We will illustrate this in the next section
for an old-fashioned  neural network, where we will see that the 
methods used are stochastic in nature.  This brings us to the second tool,
namely statistics. Since deep learning methods rely heavily on data,
it is not surprising that statistical theory is at their core. 
Statistics  provide motivation and intuition as to why certain methods are used,
and insight as to why a given approach should work.

Along with optimization stands Numerical Linear Algebra as another key component of modern
AI.  Section~\ref{sec:spectral} shows a few examples of methods exploited in
deep learning that are fundamentally rooted in NLA. 
Beyond these illustrative examples, one must realize that Deep Learning relies heavily on matrix notation and concepts in linear algebra; see, e.g., \cite{aggarwal2023neural, alom2018history, BaggagSaad2025}. 
We will also discuss NLA
further in the next section. 

Improved hardware is  another important ingredient of deep learning. It is a big leap in  
inexpensive computational power that ushered the comeback of AI in the 2010s and
beyond. Today, large machines equipped with GPUs are key to making it possible to
train very large models, along with the availability of data.  Training a large
language model can be quite expensive. A primary reason for this high cost is
that in order to extract information that exploits the \emph{context} in a given
input, it is necessary to deal with a large number $n$ of tokens at once.
However, as will be seen later, Transformers rely on attention modules that
scale quadratically with $n$. This cost can be mitigated by good hardware on the
one hand and better algorithms that aim to deliver good accuracy with a
linear, or close to linear, scaling. Hardware will not
be discussed  any further in this article due to space considerations.

It is important to also mention the impact of software development tools. At
least in its early stages, AI research was widely open with thousands
contributing new methods and algorithmic improvements.  The cumulative effect of
this openness has had an impressive impact. What made this possible are common
programming languages that allowed the exchange of software.  Around 2015,
Google Brain released TensorFlow \cite{TensorFlow16}, the result of a few years of
effort.  This release followed the tradition of sharing code for machine
learning in the Python language \cite{PythonInAI20}. About two years later
PyTorch was released and quickly became the favorite Python-based package for
deep learning~\cite{pytorch17,pytorch19,pytorch21}.  PyTorch was initially a
project of the company Meta (formerly Facebook) but in 2022 it became part of
the Linux Foundation. While PyTorch (and TensorFlow) now dominate the scene of
software development in AI, other languages exist and may gain in popularity in
the years ahead.  Among them is Google JAX \cite{JAX18}, which was introduced
by Google as a replacement to TensorFlow after the latter lost substantial
ground to PyTorch.  A common feature of the software packages just mentioned is
that they are tooled for parallel processing with multiple GPUs.  JAX has aimed
at providing better integration with hardware by facilitating the access of
custom-built chips, namely the Tensor Processing Units (TPUs) of its predecessor
TensorFlow.  Theano \cite{theano16} is another language devoted to deep
learning, one of the oldest, since it was first released in 2007. Torch-7, the
predecessor of PyTorch, was released in 2011, see \cite{PythonInAI20} for
details.



\subsection{Linear Algebra for AI} \label{sec:nla} We cannot overstate the
important role that Numerical Linear Algebra is playing in the development and
deployment of AI. NLA  provides packages
used internally (LAPACK) and  is contributing key ideas and algorithms to reduce
computational time. The matrix and tensor formalisms allow to easily express the various
transformations employed in deep learning. One of the key tools in optimizing models
is Back-Propagation, which can be efficiently carried out because it amounts to
a sequence of matrix-matrix products. In fact, one might  argue that tools from NLA
contribute the most to the improvement of algorithms for training LLMs and for
the subsequent inference. Low-rank approximations are heavily exploited, for example,
as are tensor computations in low-precision arithmetic.

While these successes are noteworthy, the megatrend of AI poses an important
question to the NLA community, and more broadly to the Numerical Analysis community, namely:
how should its members react to it?  As a new technology AI is proving to be
unusually disruptive in academia.  For example, given the excitement generated by
AI and the availability of high-paying jobs in AI-related fields, many if not
most students in computer science and elsewhere, are interested exclusively in
degrees in AI.  On the educational side, students are now able to use LLMs to
solve most questions given in homeworks and exams and this makes the testing of
students rather challenging.  Meanwhile, in research, AI is promoted by
favorable funding at the expense of traditional fields that have taken decades
to mature.  In these examples, students, educators, and researchers are led to
adapt and react in one way or another.  If we fail to adapt, our work 
may become irrelevant or at least generate little interest.

A reasonable goal in this environment is to continue to seek intellectual
innovation in Linear Algebra while participating actively in topics related to
Deep Learning research.  However, this is not an easy task due to a number of
factors.  
AI as a field has many differences with NLA, in terms of culture, notation, emphasis, etc. The community is huge and diverse not only geographically, but also in terms of its many subareas. It embraces algorithms for neural networks on the one hand, and information theory, or graph methods on the other.
The publication culture is fast-paced and completely different from
what we see in our field.

One of the main goals of this article is essentially to provide a deep look at
one of the most important topics in the field of AI, with an emphasis of
presenting the material from a linear algebra viewpoint. The reader will see that 
linear algebra is at the center of AI models and we will provide a few examples of
important innovations in LLMs that are deeply rooted in basic notions of matrix
/ tensor theory.  As a disclaimer, the authors of this paper are by no means
experts in AI. Our goal is not to provide a complete view of the field in one article
but rather to give an exposition of the main ingredients of LLMs as viewed from
a matrix-theory perspective.
Details on much of the 
material discussed in the paper can be  found in existing books and
articles, see for example
\cite{murphyBook2022,BishopBishop2024,BohnGarckeGriebel2024,LinWangLiuQiu2022,CholletWatson2025,EfficientAttentionSurvey22} among many.

\begin{figure}[htb]
\centerline{
\includegraphics[width=0.6\textwidth]{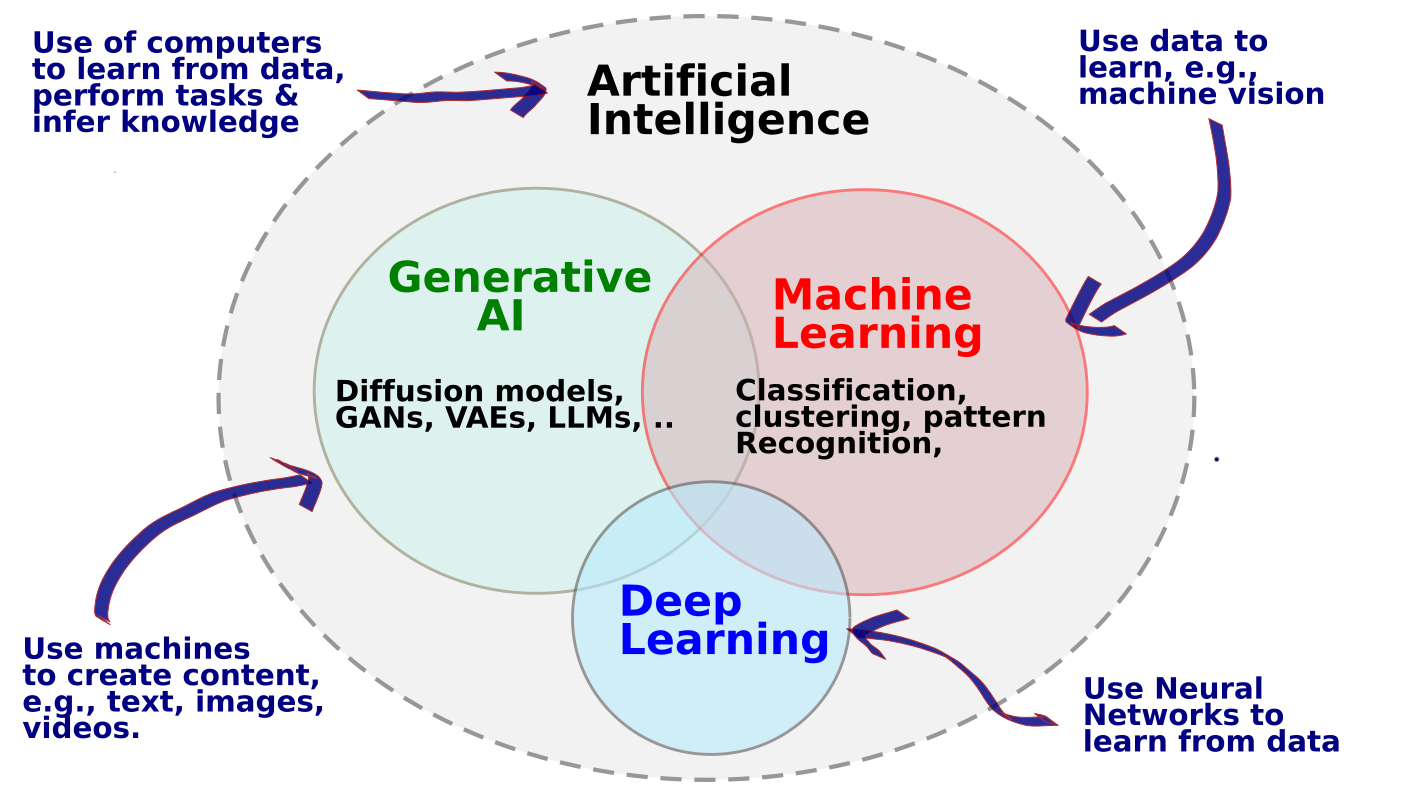}}
\caption{Components of artificial intelligence: the big picture. The schematic situates large language models within the broader AI landscape and highlights the interplay among core ingredients such as data, optimization, numerical linear algebra, hardware, and software frameworks. Its purpose is to emphasize that modern AI systems emerge from the interaction of several mathematical, algorithmic, and computational components rather than from a single isolated idea.}
\label{fig:GenVNGen2}
\end{figure}


\section{Deep Neural Networks}\label{sec:DNN} 
The ideas behind machine learning and more specifically Neural Networks began to
emerge more than six decades ago. Neural Networks were initially viewed as a means
of imitating the function of the brain: some picture is seen (input), and
analyzed to enable a comparison with stored images that have been memorized.  In
a machine learning context, we can view the process from the angle of finding a function
$\phi$ through `training'.
Once the function is found, i.e., once the model is trained, then `inference'
becomes a matter of evaluating the function for some new input and deciding on
an outcome based on the result. This, roughly speaking, is common to most neural
network-based methods.

\subsection{Example of Multi-Layer Perceptrons}\label{sec:MLP} 
Here we consider the case of a simple Multi-Layer Perceptron (MLP) to describe
how the problem of classification can be viewed from the angle of approximating
a function.  Our goal is primarily to establish notation and principles with the
help of the example of classification which is simpler than that of Large
Language Models. MLPs are also key components of Transformers which in turn are at the core
of LLMs.

Suppose we have a set of sample images of digits and
that we know the labels of these samples, a number between 0 and 9.  The problem
is to use the given dataset to build a function which will identify the label of
a new image (not in the sample). This function will take an array of pixels $x$
and produce the label via a function $\phi(x)$, a digit between 0 and 9 or a
vector of length 10, giving the 10 probabilities for $x$ to be equal to 
$0, 1, \cdots, 9$.  The function $\phi$ is determined via a set of parameters.
Thus, \emph{training a neural network can be viewed as a problem of
  approximating a function $\phi$ defined via sets of parameters.  }

\begin{figure}[htb]
  \centerline{\includegraphics[width=0.6\textwidth]{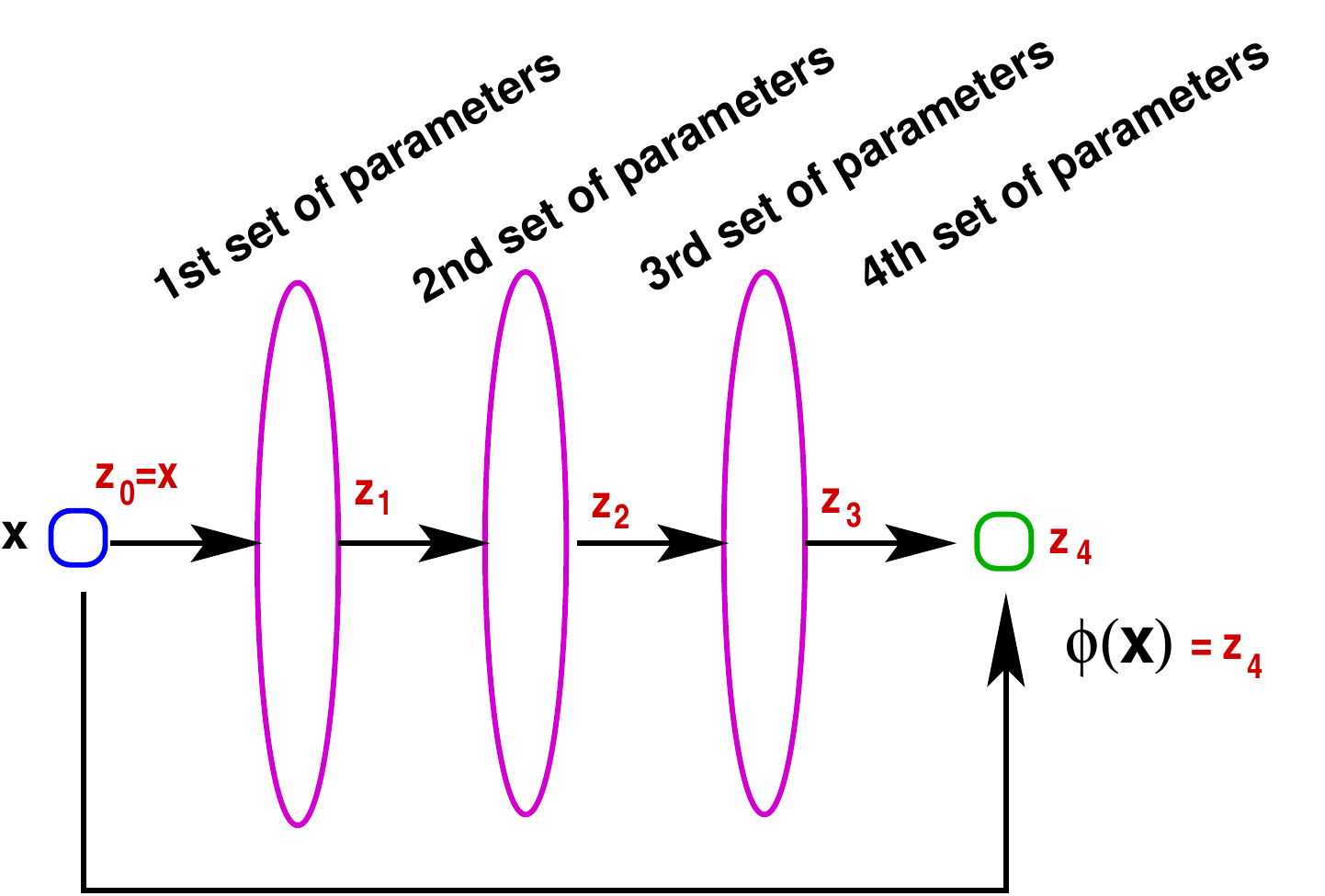}}
  \caption{A simple multilayer perceptron with three hidden layers. The input vector $\bm{x}$ is propagated through successive affine transformations and nonlinear activation functions to produce the final output $\phi ( \bm{x} )$. The figure is intended to illustrate the layered composition that underlies feed-forward neural networks and serves here mainly to establish notation.}
  \label{fig:dnn1in} 
\end{figure}

The unknown function $\phi$ is defined through a number of stages corresponding
to layers in the network. In this section, we consider the simplest case of
Multi-Layer Perceptron, depicted in Figure~\ref{fig:dnn1in}. The network in the
figure  has an input layer (leftmost, where $x$ enters), 3
hidden layers (elongated ellipses) and an output layer (rightmost).  The input on
the left side is typically a vector of length $d$, whose components are often called
features.
An important point about the prevailing notation is that such
feature vectors are often viewed as row-vectors, i.e., $x \in \RR^{1\times d}$, and they form the rows
of a sample matrix that is processed in block through the layers, as is discussed later.
However, for convenience we will assume for now that $x$ is a $d$-dimensional vector
$x \in \ \RR^d$.

If we wish to distinguish between two given sets of input data, we could use a hyperplane:
$\phi(x) = w^T x + \beta$. The two sets can be separated by 
the sign of $\phi(x) $ ---positive sign  for one set and nonnegative for the other.
So we could use the function:
\eq{eq:phiofx}
  \phi(x)  = \sigma ( w^T x + \beta)
\en
where here $\sigma $ is  the sign function. Thus, if we had a number of training  data points
$(x_i, y_i)$ with binary labels (e.g., `spam'--`non-spam',
`malignant' --`non-malignant',...) where $y_i = \pm 1$, we could use this set to
determine an optimal $w$ for which $\phi(x_i) \approx y_i$ for $i=1,\cdots,n$.

The functions used in neural networks are generalizations of the above
function. Instead of a single vector $w$ we will use a $d \times d_{out}$ matrix
$W$ and $\sigma$ is replaced by a continuous function. The result is that
$\phi(x) $ is a $d_{out}$-dimensional vector.
 
Going through
the first layer, $x$ is transformed into $W_1^Tx + b_1$ where
$W_1 \ \in \ \RR^{d \times d_1 }$ is some unknown matrix of weights to be determined
and $b_1 \in \RR^{d_1} $ is a bias, also to be determined.  This first
linear transformation is then compounded with a nonlinear function, called
\emph{activation function} and usually denoted by $\sigma$, so the output of the
first layer, denoted by $z_1 $ is
\eq{eq:z1} z_1 = \sigma(W_1^T x + b_1) .
\en
There are a number of choices for the activation function $\sigma$, but the most
common is the Rectified Linear Unit, or ReLU:
\eq{eq:ReLU} \sigma (t) = \max
\{0, t \} .
\en
Two other common activation functions are the sigmoid
$\sigma(t) = (1+e^{-t})\inv$ and the hyperbolic tangent
$\sigma (t) = \tanh (t) = (e^t - e^{-t}) / (e^t + e^{-t})$.  Note that the value
of ReLU is nonnegative, while those for the sigmoid and tanh lie in $(0,1)$ and
$(-1, 1)$ respectively.\footnote{In fact many other, more effective, activation functions have recently been
  developed.
  The simplest of these is the \emph{smoothed ReLU}, or \emph{Swish} function $\text{Swish}(x) = x  \cdot  \text{sigmoid} (x)$.
A function of the same type is 
$\text{GELU} (x) = x \Phi(x)$ where $\Phi(x) $ is the Gaussian CDF - often approximated by a degree 3 polynomial.
The LLaMA models generally  use a more sophisticated activation known as `SwiGLU', a member of the
Gated Linear Unit (GLU) functions. It has a 'gate' component and a 'value' component and can be written as 
$  \text{SwiGLU} (x) =  x W_v \odot \text{Swish} ( x W_g )$ where $\odot$ denotes the component-wise product.
A difference with the simpler functions just seen is that the weights $W_v, W_g$ are learned during training.}

The second level takes the output $z_1$ and applies to
it a transformation that is similar to the first one:
$z_2 = \sigma(W_2^T z_1 + b_2)$. Generally, going from layer $l-1$ to layer $l$
we have
\eq{eq:zl} z_l = \sigma(W_l^T z_{l-1} + b_l) ,
\en
where $W_l \ \in \ \RR^{d_{l-1} \times d_l}, b_l \ \in \ \RR^{d_l}$, and $\sigma$
is the same activation function as above.  Assuming there are $L$ hidden layers
($L = 3$ in Figure \ref{fig:dnn1in}), then counting the input and output layers,
we have $L+2$ layers altogether. The parameters $W_l, b_l$ define a mapping
from data in layer $l-1$ to data in layer $l$ and we have $L+1$ of these,
starting with $W_1, b_1$ and ending with  $W_{L+1}, b_{L+1}$.
The last vector to be computed is $z_{L+1}$
(e.g., $z_4$ in Figure~\ref{fig:dnn1in}) and $\phi(x)$ is  set to this output.
Thus, for the example in Figure~\ref{fig:dnn1in},
\eq{eq:phiofxF}
\phi(x) = \sigma(W_4^T \sigma(W_3^T \sigma(W_2^T \sigma(W_1^T x + b_1) + b_2)  + b_3) + b_4) . 
\en
This function is better expressed in algorithmic form, as shown in Algorithm~\ref{alg:MLPfwd}.
(Note that the algorithm does not solve a problem yet - it just defines the function $\phi$, given the parameters). 

\begin{algorithm}[htb]
  \centering
  \caption{Forward Propagation}\label{alg:MLPfwd}
  \begin{algorithmic}[1]
\State  \textbf{Input:} $x \in \mathbb{R}^{d}$, \textbf{Output:} 
$y  \in \RR^{ C}$     
\State Set: $z_0 = x$                       
\For{$l=1:$ \texttt{L+1}} 
\State  $z_{l} = \sigma(W_l^T z_{l-1} + b_l)$ 
\EndFor 
\State {Set:} $\phi(x)  := z_{L+1}$ 
\end{algorithmic}
\end{algorithm} 

\begin{figure}[htb]
  \centerline{\includegraphics[width=0.4\textwidth]{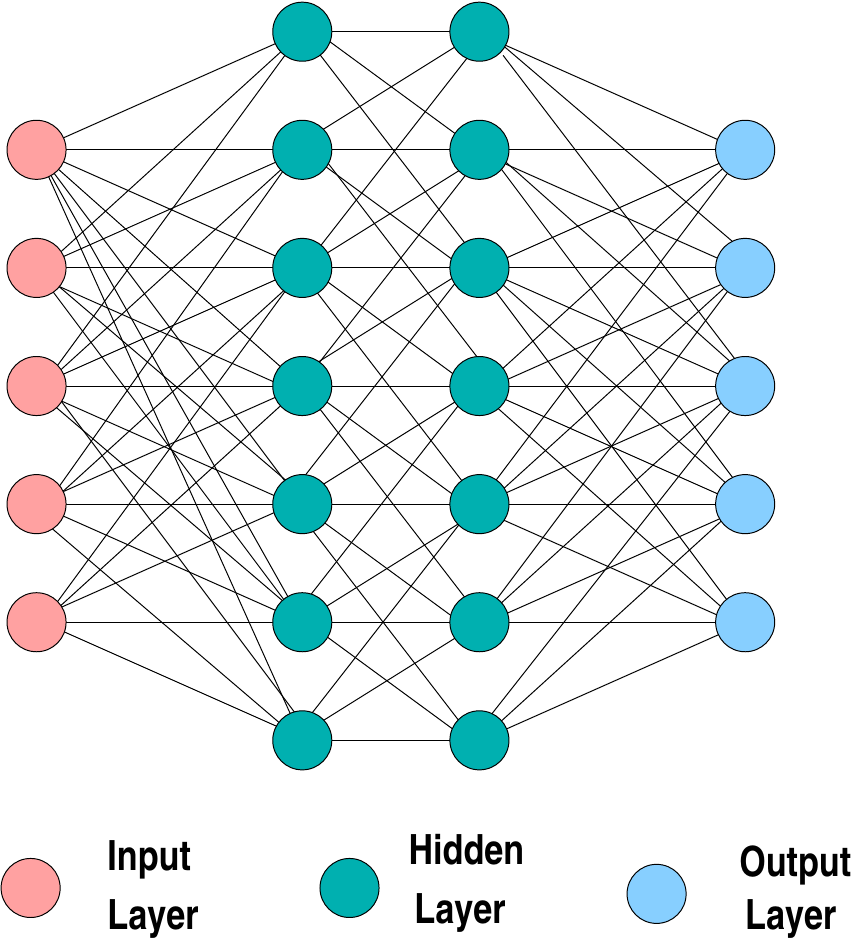}}
  \caption{A simple multilayer perceptron with two hidden layers. The figure illustrates the layer structure used in the classification example and, together with Table~\ref{tab:mlp}, helps explain the dimensions of the weight matrices and the total number of trainable parameters.}
\end{figure}

The problem is to find a function $\phi$ thus defined through these parameters
in such a way that given some data, i.e., a set of $x_i$'s for which the exact
outcome $y_i$ is known (training data), the value of $\phi(x_i)$ is closest to
$y_i$ according to some measure.

In the case of digit recognition, we would have a set $x_1, \cdots, x_n$ of
images of digits for which the exact digit $y_i$ is known. Each image is an
array $m_1 \times m_2$ of pixels which is vectorized into a vector of length
$d_0 \equiv m_1 m_2$. Each of these features $x_i$ will have a value of $y_i$
between 0 and 9.  In many situations, it is convenient to recast the digit $y_i$
into a so-called one-hot vector which is a vector of 10 entries that are zero
except for the one corresponding to the digit $y_i$ which is set to one. Thus,
if the digit is 2, the vector $y_i$ will be the canonical vector $e_3$ in
$\RR^{10}$, which is the 3rd column of the identity matrix of size 10.

\begin{table}[htb]
  \begin{center}
\begin{tabular}{r|c|c|c|} 
                     & Layer 0 (=in) to 1 & Layer 1 to 2 & Layer 2 to 3 (=out) \\ \hline
  $\text{size}(W_l)$ & (400$\times$100)          & (100$\times$100)    &  (100$\times$10)           \\ 
   $\text{size}(b_l)$ & (100$\times$1)           &  (100$\times$1)     &   (10$\times$1)            \\ \hline
 Total Params        &  40,100            &  10,100     &   1,010             \\ 
\end{tabular}
\end{center}
\caption{Dimensions and parameter counts for a simple multilayer perceptron. The table lists the sizes of the weight matrices and bias vectors for each layer in the digit-classification example, together with the corresponding number of trainable parameters. It illustrates how the total parameter count is determined directly by the layer widths. The total number of parameters is $51{,}210$}
\label{tab:mlp}
\end{table}

As an example, suppose that $m_1 = m_2 = 20$ and that we have 1 hidden layer
with $d_1 = 100$.  Then the input data has size $n \times d_0$ where $d_0 = 400$
and the output will be of size $n \times 10$.  The parameters involved in the
training are shown in Table~\ref{tab:mlp}. Note that the number of parameters
does not depend on $n$ the number of input images - which is arbitrary. In the
implementation $n$ disappears from the scene: when coding the training part we
need not worry or specify $n$, but only the number of parameters that define the
move from one layer to the next.  The problem now is to find the function $\phi$
(i.e., matrices $W_l$) such that $\phi(x)\approx y$ for each of the data pairs
$(x,y)$.

\subsection{The loss function for MLPs}\label{sec:trainMLP}


Training the model requires a set of data points $x_i, y_i, i=1:n$. The input
can therefore be set as a matrix $X$ of size $n \times d_0$, in which each row
corresponds to a sample.
The output can be a vector $Y$ of length $n$, whose
entries are the known labels of the samples.  In classification, it is also
common practice to replace the label of item $i$ by a
\emph{one-hot row vector}  $u_i$ of length
$C$, where $C$ is the number of classes, as was described earlier. In this
situation $Y$ is $n\times C$.  Thus, we seek a function $\phi$ such that
$\phi(x_i) \approx y_i$ for $i=1:n$, or, in matrix form, $\phi(X) \approx Y$.
Using matrix notation again, each of the internal variables $z_l$ defined
earlier becomes a matrix $Z_l$ of size $n \times d_l$ and the transformation
\nref{eq:zl} becomes: \eq{eq:zlB} Z_l = \sigma( Z_{l-1} \times W_l + b_l) \en
where $W_l \in \RR^{d_{l-1} \times d_l}, b_l \ \in \ \RR^{1 \times d_l}$, and
$\sigma$ are the same as before.
Note the change of notation where
the samples $x_i$  and internal variables $z_i$ seen in Section~\ref{sec:MLP} are now
row vectors that occupy the rows of the matrix $X$ and $Z_l$ respectively.
Here we also need to shed light on a feature of
notation that is common in this context. The product $ Z_{l-1} \times W_l $ in
\nref{eq:zlB} is a matrix of size $n \times d_l$, and when the row vector $b_l$
is added to it, it is meant that it is added to each of its rows\footnote{This
  is called \emph{broadcasting}, a term borrowed from the Python language.
  Since Python is heavily used in AI, its syntax and terminology has permeated
  the notation used in the field. Incidentally, in R, a
  popular language in computational statistics, the equivalent term to
  broadcasting is \emph{vector recycling} or \emph{recycling rule}.  MATLAB,
  started supporting broadcasting with version R2016b, and it calls 
  this operation \emph{implicit expansion}.}.

For the given training data $X \in \RR^{n\times d_0}$ and corresponding ground
truth $Y \in \RR^{n\times d_{L+1}}$, we need to express an optimization problem
which will consist of finding an optimal set of parameters $W$, such that
$\phi_W (X) \approx Y$. Here by $W$ we mean the set of parameters
$W_1, \cdots, W_{L+1}$ along with the biases $b_1, \cdots, b_{L+1}$.
The final output of the network is the matrix
$Z_{L+1} = \phi_W(X) \in \RR^{n\times C}$.
To simplify notation we will call $Z \equiv Z_{L+1}$ this final output.

We added
 $W$ as a subscript to the function $\phi$ to emphasize its dependence on these
parameters. A better notation might be $\phi(X | W)$, which is to be read as
``the result of $\phi$ for data set $X$, given the parameter set $W$.''
When the input data $X$ is fixed, as is usually the case for training, then we can just write
$\phi (W)$. When $W$ is fixed, as is the case when testing, i.e., during the inference phase,
we evaluate $\phi_W(x)$, which can be written as $\phi(x)$ without ambiguity,
for some data item $x$.

One possible formulation for an objective function to minimize could be: 
\eq{eq:obj}
\min_W { \scl (W) \equiv \| Y - \phi_W (X) \|_F^2 =
  \sum_{i=1}^n \| y_i - \phi_W(x_i)\|_2^2 
}
\en
where $\| . \|_F $ represents the Frobenius norm.
Recall that $Y$ and $ \phi_W (X)$ are of size $n \times d_{L+1}$ where
$d_{L+1} \equiv C$ is the number of classes. This simple formulation works fine but it is seldom employed.
A preferred approach is to exploit the cross-entropy - a notion of  distance based on information theory. 
First, a given output $z = \phi_W(x)$, a row vector of length $C$, is  written as $z = [\zeta_1, \cdots, \zeta_C]$
and transformed as follows to define new output variables:
\eq{eq:softmax0} \hat p_i =
\frac{\exp( \zeta_i) }{\sum_j \exp( \zeta_j) } . 
\en
The above operation is called
\emph{softmax} and it is performed component-wise on a given row (or
column) vector.
Note that each $\hat p_i $ is nonnegative and does not exceed one.
So the choice of symbols is deliberate:  $\hat p_i $ can be viewed as
the probability for $x$ to be in class $i$, for $i=1, \cdots, C$.

Assuming that we have $n$ samples $x_1, x_2, \cdots, x_n$ we will end up with a
final output matrix $Z$ of size $n \times C$. We now apply the softmax operation
to each row of $Z$. Let $z_{i,:} $ denote the $i$th row of $Z$.
With  this notation, the softmax operation will transform each row $z_{i,:}$
into 
\eq{eq:softmax} 
  \hat p_{i,:}  \ = \  \frac{  \exp (z_{i,:} ) }{\sum_{j=1}^C \exp (z_{i,j} )}
\en
where the exponential function is applied componentwise to the row-vector.  The
new output is now the matrix $\hat P$ whose rows are defined above.  The end
result is that each row-sum of $\hat P$ is equal to one and each entry is
nonnegative.  The clear advantage of this simple transformation is that the entries
can now be interpreted as a probability distribution.

Next we set the cross entropy cost for sample $i$ which is 
  \[
    - \langle p_{i,:}, \log (\hat p_{i,:} ) \rangle 
    = -\sum_{j=1}^C p_{i,j} \log \hat p_{i,j} 
  \]
  where we use $\langle . , . \rangle$ to denote the dot product of two
  row (or column) vectors.   Thus, the \emph{cross-entropy} function which we want to minimize is the mean
  of these costs over the $n$ samples
  \eq{eq:CEcost}
\scl (W) = \frac{1}{n} \sum_{i=1}^n  - \langle p_i, \log \hat p_i \rangle \ ,
\en
where each $p_i$ is the one-hot vector  obtained from the ground-truth labels $y_i$.

The cross-entropy loss function is very common in machine learning.
There are compelling statistical justifications for using this measure
of distance between two probabilities rather than the simple Euclidean Norm,
see~\cite{murphyBook2022} for details.

Two ideas from numerical analysis played a major role in DNNs:
the Universal Representation Theorem \cite{Cybenko89}, and the idea of backward
differentiation by Andreas Griewank \cite{griewankBook,Averick-al,Griewank89}
which is known as Backpropagation in machine learning.
These will be covered in the next few sections.

\subsection{The Universal representation theorem} 
\label{subsec:universal_theorem}
Since the goal of a MLP is to approximate an unknown function $f$ by a
composition of functions of the form \nref{eq:phiofx}, this raises the question
as to whether or not such an approximation is always possible. The question was
answered by a theorem in approximation theory proved by George Cybenko in 1989
\cite{Cybenko89}, which provided a major theoretical tool in the study of  neural networks.

We denote the space of continuous functions from $I_n = [0,1]^n$ to $\RR$ by
$C(I_n)$ and consider the set of functions of the form:
\eq{eq:cyb1} G(x) =
\sum_{j=1}^n \alpha_j \sigma (w_j^T x + \theta_j) , \en
where $\sigma$ is a
scalar function, the $\theta_j$'s are scalars and the $w_j$'s are vectors in
$\RR^n$.  The first question addressed by Cybenko's article \cite{Cybenko89} is: Under
what condition(s) is the set of such functions dense in $C(I_n)$?
The set $C(I_n)$ is a normed space and the norm used is the uniform norm, i.e., supremum
or infinity norm.  The set of functions of the form \nref{eq:cyb1} is dense if
given $f \in C(I_n)$, and any $\epsilon>0$ we can find a function $G$ of the
form \nref{eq:cyb1} such that $\| G - f \| \equiv \max_{x\in I_n} | G(x) - f(x)| \le
\epsilon$.  The paper proves that in order for this to be the case, it is
sufficient that $\sigma$ be continuous and \emph{discriminatory}.  A function
$\sigma $ is said to be {discriminatory} when for a given signed Borel measure
$\mu$ \[ \int \sigma(w^T x + \theta) d\mu (x) = 0 \quad \forall w, \forall \
\theta \quad \rightarrow \quad \mu \equiv 0 .  \] In other words if for all $w,
\theta$ the integral of $\sigma(w^T x+ \theta)$ with respect to the measure
$\mu$ is zero then $\mu$ must be zero. If this property does not hold then
$\sigma$ is not
discriminatory.  The proof of the theorem uses functional analysis and
topological arguments.

The paper then establishes that an important class of function used in machine learning
are discriminatory. These are the \emph{sigmoidal} functions that
satisfy the condition
\eq{eq:sigmoidal}
\limit_{t \to +\infty}  \sigma(t) = 1, \qquad
\limit_{t \to -\infty}  \sigma(t) = 0.
\en

\subsection{Stochastic Optimization techniques for DNNs}\label{sec:TrainDNN}
Given some training data consisting of inputs $x_i$ and outputs $y_i$, training the neural network
will consist  of exploiting an optimization technique to minimize an objective function
or `cost function', like \nref{eq:obj} or \nref{eq:CEcost}. However, this goal faces a number of  challenges.
First  the objective functions under consideration are generally far from being  convex.
Second, a solution with a small cost function does not necessarily achieve a small
\emph{`generalization error'}
which is the error on data that is not used in training,  calculated in principle on the
entire population.   When pursued to high accuracy  traditional optimization will often yield
poor generalization and we say that the model is `overfitted' in this situation. 
A third problem is that the number of parameters may be huge. 
A rule of thumb that has gradually emerged in the field is that the larger number of
parameters is necessary to achieve accurate responses from LLMs, see, e.g.,   \cite{kaplan-al20}.
A fundamental approach to address all these issues is to resort to
stochastic optimization.

\subsubsection{Stochastic Gradient Descent (SGD)} \label{sec:SGD} 
To begin with, recall the classical (standard) Gradient Descent (GD) method 
 for minimizing a convex function $\phi(w)$ with
respect to $w$.  If  $\phi$
is differentiable, in addition to being convex, the method consists of
taking the iterates:
\eq{eq:gradMeth} w_{j+1} = w_j - \eta_j \nabla
\phi(w_j),  \en where in the classical approach due to Cauchy
\cite{Cauchy47}, $\eta_j$ is determined by performing a
line search, i.e., by minimizing $\phi(w_j -\eta \nabla \phi(w_j)) $ over $\eta$.

In our situation $w$ is the vector of all parameters of the model,
as represented by the set $W$ of parameters seen earlier, stacked together as a long vector.
For the example shown in Table~\ref{tab:mlp}, the length of $w$ is $51,210$.
Note that the implementation of SGD, or other approaches, does not require that
we explicitly convert the set of parameters $W$ into a single long vector $w$.
 We do it here mainly for clarity.

In the specific context of machine learning $\phi(w)$ is
often the sum or the mean of a large number of other, elementary,  cost functions, i.e., we often have
\eq{eq:fw} \phi(w) = \frac{1}{n} \sum_{i=1}^n \phi_i (w) \qquad
\rightarrow \qquad \nabla \phi(w) = \frac{1}{n}  \sum_{i=1}^n \nabla \phi_i (w) .
\en
Problems that can be expressed in this form are often termed \emph{Finite Sum} problems.
For example, note that both loss functions \nref{eq:obj} and \nref{eq:CEcost}
are in this form.  The Stochastic Gradient Descent (SGD) method is 
designed specifically for the common case where $\phi(w)$ is of the form
\nref{eq:fw}. It is usually expensive to compute $\nabla \phi$ but inexpensive
to compute a component $\nabla \phi_i (w)$. Hence the idea of replacing
the gradient $\nabla \phi(w_j) $ in \nref{eq:gradMeth} by 
$\nabla \phi_{i_j} (w_j)$ where $i_j$ is drawn at random at step $j$. The result
is an iteration of the type:
\eq{eq:SGD} w_{j+1} = w_j -\eta_j \nabla \phi_{i_j}
(w_j) \ .
\en
The step-size $\eta_j$, called the `learning rate' in this context, is often set to a
certain fixed value $\eta$ in advance, or determined adaptively.
Clearly, the lower cost of each step must imply that overall many more iterations will be
required, since the full gradient is roughly approximated. An interesting observation
concerns the case when the learning rate $\eta_j$ is constant,  $\eta_j \equiv \eta/n$.
Then clearly after one sweep of $n$ substeps, we get
\[
  w_{j+n} = w_{j} - \eta \frac{1}{n} \sum_{k=j}^{j+n-1} \nabla \phi_{i_k} (w_k)\]
where each $\phi_{i_k}$ in the sum is drawn at random for  the intermediate point $w_{k}$.
This suggests  that in this case the full gradient is approximated by some sample mean of
gradients drawn at random.

The origin of the SGD algorithm goes back to 1951 with the seminal article of 
Robbins and Munro \cite{RobbinsMunro51}. This paper considered the general
problem of finding the root of the equation $M(w) = \alpha$ where $M(w)$ is not available to
the `experimenter' but can be measured,  or sampled, via a random variable
$H(w)$ that satisfies  $\EE[H(w)] = M(w)$, where $\EE$ stands for expectation.
Under these assumptions the deterministic scheme
$w_{n+1} = w_n -a_n M(w_n) $ is not feasible but can be replaced by an iteration
of the form $w_{n+1} = w_n -a_n g_n $ in which $g_n = H(w_n)$.
The  same article showed convergence results under some assumptions.
In particular the following conditions must be satisfied to establish
convergence in expectation: 
\eq{eq:RobMun}
\sum_{i=1}^\infty a_i = \infty, \quad \sum_{i=1}^\infty a_i^2 < \infty . 
\en
The Robbins and Munro framework has found many uses in deep learning and elsewhere.
Its main premise is that the function $M$ can  only be evaluated  with noise whose mean
is zero. 
Further analysis of the algorithm can be found in
\cite{hardt16,gower2019sgd,ljung1977analysis}.
The article \cite{hardt16} explored a rather interesting observed phenomenon with SGD, namely that
it leads to small (`vanishing') generalization errors, which is a desirable feature.
The more recent article \cite{gower2019sgd}
analyzed the algorithm specifically for the modern machine learning
context.

A straightforward SGD approach that uses a single function $\phi_{i_j}$ at a time is
seldom used in practice because this typically results in a convergence that is
too slow.  Instead, a common alternative is to resort to
\emph{mini-batching}---a middle ground solution between the one-subfunction SGD
and the full Gradient Descent algorithm.  In short the idea consists of
replacing the single function $\phi_{i_j}$ by an average of few such functions -
again drawn at random from the full set.

Mini-batching is often implemented in  \emph{sampling without replacement} fashion:
  once a sample is utilized it is left out for the next sample until all samples in the data set have been utilized.
 Note that  theoretical results for SGD  are often proved for sampling with replacement.
A common way to implement the  procedure is to first partition the set
$\{1,2,\cdots,n\}$ into $n_B$ `mini-batches' ${\cal{B}}_j, j=1, \cdots, n_B$
where
\eq{eq:batches}
\bigcup_{j=1}^{n_B} {\cal{B}}_j =  \{1, 2, \cdots, n\} \quad \mbox{with} \quad
{\cal{B}}_j \bigcap {\cal{B}}_k  = \emptyset \quad \mbox{for} \quad j\ne k     
\en
Here, each ${\cal{B}}_j \subseteq \{1, 2, \cdots, n\}$ is a small set of indices.
Then instead of considering a single function $\phi_i$ we will consider  
 \[ \phi_{ {\cal{B}}_j } (w) \equiv \frac{1}{ | {\cal{B}}_j | } 
   \sum_{k \in {\cal{B}}_j} \phi_k(w) . \]
 Recall that the standard notation $| X |$ represents the cardinality of the
 set $X$.
We will cycle through all mini-batches ${\cal{B}}_j$ of functions, each time
performing a group-gradient step of the form: 
\eq{eq:OneEpoch}
w_{j+1} = w_j - \eta  \nabla \phi_{ {\cal{B}}_j} (w_j)  \quad j=1,2,\cdots, n_B .
\en
If all sets ${\cal{B}}_j $ are small enough then computing the gradient will be
manageable and computationally efficient.  One sweep through the whole partition
of the set of functions in \nref{eq:batches} with the iteration
\nref{eq:OneEpoch} is termed an \emph{`epoch'}.  The number of iterations of SGD
and other optimization learning in Deep Learning is often measured in terms of
epochs.  It is common to select the partition at random each time by reshuffling
the data at each epoch and selecting the batches in sequence on the shuffled
data.  Models are expensive to train: the more complex models may require
tens of thousands of epochs to converge.

Mini-Batch processing in the random fashion described above is advantageous from
a computational point of view since it typically leads to fewer sweeps through
each function to achieve convergence. It is also mandatory if we wish to avoid
reaching local minima and overfitting.  Stochastic Gradient Descent (SGD)
approaches of this type are at the heart of optimization techniques in deep
learning. This basic  method was refined in a number of ways.


\subsubsection{Adagrad and RMSprop} \label{sec:adagrad}
In the standard gradient descent algorithm, all parameters are updated using the same learning
rate.  This can cause some difficulties when dealing with sparse data or when different
features have different scales.
The Adaptive Gradient Algorithm (AdaGrad or Adagrad) deals with this problem by adaptively
adjusting the learning rate for each parameter individually in a model by
considering the history of past gradients \cite{adagrad-paper}.

By its nature, the algorithm must be expressed component-wise and so
we will make use of component-wise vector operations. The weights are  updated by the 
following recurrence:
\eq{eq:ADAg}
w_{t+1} = w_{t} -\eta  \frac{ g_t }{\sqrt{\tilde g_t^2  +\epsilon}} 
\en
where $\eta$ is the starting learning rate,  $g_{t} $  is the  gradient of the loss function, 
$\tilde g_t^2 $ is the sum of the squares of     the past gradients, i.e.,  
\eq{eq:ADAg1} \tilde g_t^2 = \sum_{k=0}^t (g_{k})^2 , \en 
    and $\epsilon$ is a small constant used to avoid divisions by zero or small numbers.
    The vector operations are to be understood as component-wise operations. Thus,
    $g_k^2$ in \nref{eq:ADAg1} is the vector of the squares of the components of $g_k$
    and the vector division in \nref{eq:ADAg} is a component-wise division. 
    Observe that the algorithm is essentially equivalent to adjusting the learning rate $\eta$
    by dividing it by a scalar that varies with the component:
    \eq{eq:ADAg2}
      \eta \to  
      \frac{ \eta }{\sqrt{\tilde g_t^2  +\epsilon} } . 
    \en
    
    A known drawback of the scheme just described is what is called
    \emph{learning rate exhaustion}: as the algorithm progresses, the modified learning
    rate \nref{eq:ADAg2} will tend to vanish  because of the tendency of the denominator to
    become large when many steps are taken.

RMSprop \cite{RMSprop-paper} addresses this issue by keeping a 
running average $v_t$ of the squared gradients instead of their sums.
The algorithm is as follows: 
\begin{align}
  v_{t} &= \beta   v_{t-1} + (1- \beta) g_{t}^2 
                \label{eq:RMSprop1} \\ 
  w_{t+1} &= w_{t} - \eta \frac{ g_{t} }{\sqrt{v_{t} +\epsilon}}  . 
              \label{eq:RMSprop2}
\end{align}
Here $\eta$ is the starting learning rate, $\beta <1$ is a parameter, and the
notation for $g_{t} $ and $\epsilon$ is the same as for Adagrad.  Note that the
vector operations in equations \nref{eq:RMSprop1} and \nref{eq:RMSprop2} are
again carried out component-wise as in Adagrad.


The vector $v_t$ defined in \nref{eq:RMSprop1} is an Exponential Moving Average
(EMA) of the squared gradients.  Indeed, $v_t$ is a linear combination of the
previous squared gradients, where the coefficients decay exponentially as we
proceed backward.  This is to be contrasted with Adagrad where this EMA is
replaced by a sum of previous gradients.
However, the article \cite{ReddiAl19} indicates that this strategy is not
without problems, and that this has to do primarily with the use of mini-batches.

\subsubsection{Momentum} Momentum techniques are inspired by second order acceleration
methods, \`a la Aitken~\cite{Aitken}.
Each iterate of the basic scheme \nref{eq:gradMeth} is modified by
adding a small multiple of the previous increment $w_j - w_{j-1}$:
\eq{eq:mmtm} w_{j+1} =
\left(w_j -\eta \nabla \phi_{\calB_j} (w_j) \right)  +  \mu (w_j - w_{j-1}) . 
\en
When $\mu = 0$ the scheme falls back to the original SGD approach.

\centerline{
\includegraphics[width=0.4\textwidth]{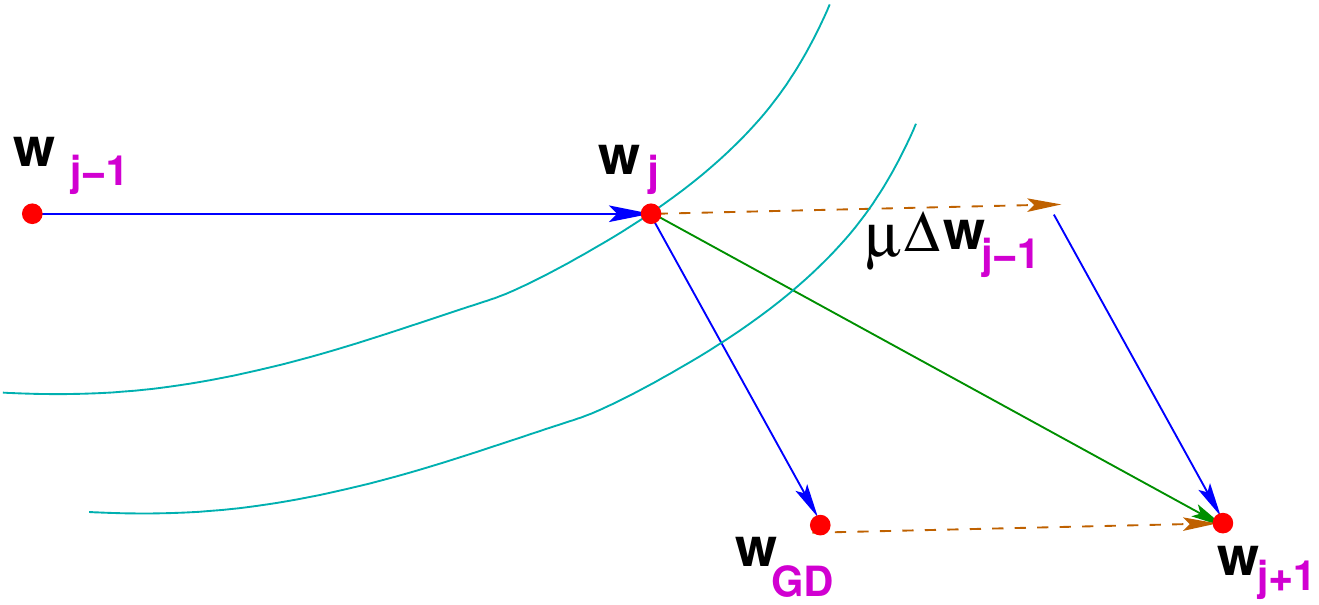}
}

The numerical linear algebra specialist may recall that the Chebyshev iteration
for solving linear systems also termed ``2nd Order semi-iterative method''
 \cite{GolubVarga1961}, takes the form of momentum with
scalars $\eta$ and $\mu$ that vary at each step.
The iteration \nref{eq:mmtm} is often rewritten in the following form:
\eq{eq:mmt1} 
      v_{j+1} = \mu v_j - \eta \nabla \phi_{\calB_j}(w_j) ; \qquad
      w_{j+1} = w_j + v_{j+1}  . 
      \en
Here  the term $v_j$ is  termed `velocity'. 
There are many variations around this idea which lead to some of the most
successful schemes used in deep learning.

\subsubsection{Adam and AdamW}

The Adaptive Moment Estimation (Adam) algorithm combines the ideas of momentum with
those of RMSprop \cite{Adam-paper}. The main difference with RMSprop is that Adam has two
momentum terms, one for the gradient  and the other for variance. Both of these
 are attenuated  exponentially. Below is a
 sketch of the algorithm, where $g_t$ is the gradient at time step $t$,and
$\beta_1$ and $\beta_2$, are two exponential decay rates.
\begin{align}
m_{t} &= \beta_1 m_{t-1} +  (1-\beta_1) g_{t} \ ,  \qquad
& \hat m_t &= m_t / (1 - \beta_1^t) &   \label{eq:adam1} \\
v_{t} &= \beta_2 v_{t-1} +  (1-\beta_2) (g_{t})^2 \ ,  \qquad
& \hat v_{t} &= v_{t} / (1 - \beta_2^t) & \label{eq:adam2} \\
  w_{t+1} &= w_{t} -  \frac{ \eta \hat m_{t} }
              {  \sqrt{\hat v_{t}} + \epsilon } .  
        & & \label{eq:adam3} 
\end{align}
As in Adagrad and RMSprop,  the algorithm uses  component-wise vector operations.
The generally recommended parameters are 
$\beta_1 = 0.9, \ \beta_2 = 0.999, \ \epsilon = 10^{-8}$.
The algorithm is fairly efficient in training neural networks and is one of the
preferred techniques used in this context. 
 

Adam is often implemented with a form of regularization known as \emph{weight decay} whose goal is
to penalize large weights \cite{hanson1988comparing,AdamW-paper}. It is best to first
introduce the method
in the context of SGD, where it can be interpreted as a regularization technique.
It was observed that \cite{hanson1988comparing} improved networks can result if 
``Weights  decay differentially allowing large weights to persist and small weights to
decrease to zero sooner.''
In the case of SGD, the iteration \eqref{eq:SGD}, where the learning rate $\eta_j$ is now constant,
is replaced by
\eq{eq:SGDdec}
w_{j+1} = (1-\lambda) w_j -\eta \nabla \phi_{i_j} ,
 \en
 where $\lambda$ defines the rate of weight-decay.  As can be readily seen
  the above scheme is mathematically equivalent to
 applying the standard SGD to the following regularized problem at step $j$:
 \eq{eq:regSGD}
 f^{reg}_{i_j} (w)  = f_{i_j}(w) + \frac{\mu}{2} \| w \|_2^2 \quad \mbox{with}\quad
 \mu = \frac{\lambda }{\eta} . 
 \en

 When introduced into the Adam framework, the above idea was first extended by replacing 
 \eqref{eq:adam3} by: 
\eq{eq:adam4} 
  w_{t+1} = w_{t} -  \frac{ \eta}{  \sqrt{\hat v_{t}} + \epsilon } . ( \hat m_{t}  + \lambda w_t) . 
\en
However, it was observed in \cite{AdamW-paper} that we  can no longer interpret this modification as
a form of regularization when the gradients are \emph{adaptive}, i.e., when they are scaled by their
historic magnitude as is the case in Adam. In this case, the term in the parentheses of
expression \eqref{eq:adam4} indicates that modification $\lambda w_t$ is added to the
\emph{unscaled} gradient weighted average of the gradients, i.e., $\hat m_t$. It was argued that it is best to
modify the scaled version of the average $\hat m_t$. This will have the effect of
regularizing weights with large magnitude more than with standard regularization. In effect this just
amounts to moving the left parenthesis back so the update becomes:
\eq{eq:adam5} 
  w_{t+1} = w_{t} -  \eta \left(  \frac{ \hat m_t }{  \sqrt{\hat v_t} +\epsilon } + \lambda w_t \right) . 
\en
AdamW has been shown to yield much better generalization than standard Adam and it is often used as the 
default optimizer in Large Language Models.

\subsection{Challenges of Deep Learning and the issue of generalization}
\label{sec:gener} 

Methods like Adagrad, RMSprop, and Adam are termed \emph{adaptive} and are known to perform generally
better than the non-adaptive techniques such as SGD. What this means is that they
will result in faster convergence when considering the learning rate.
However, it has been discovered that better convergence, i.e., lower values of the
cost function, does not mean that the method will perform well on new test data.
This is referred to the problem of \emph{generalization}: How will the model perform
on data that is  not in the training set?
Even though adaptive methods may reach a lower loss during the training process, the model
performance on test data is often worse than the  simpler non-adaptive  methods.
In \cite{wilsonAl17} it was shown  that when we 
apply both types of methods on the Cifar10 dataset, adaptive methods seem to converge faster
but they lead to poorer generalization than their non-adaptive counterparts.
Also, the authors of \cite{chenAlClosingGap18} introduce a variation of Adam termed
\emph{Partially adaptive momentum method (PAdam)}, which adapts the learning parameter
partially according to a parameter $p$. They suggest that the issue with adaptive methods
is `over-adaptation' and their experiments 
indicate that adaptive methods do a poor job at  traversing through the parameter space in
the middle to late stage of the training process.


Here we should ask what role do stochastic methods play.  We could consider using the full
gradient which amounts to taking the full batch, i.e., the whole data set, at once at 
each step.  However, it is often argued that in deep learning an exact
minimization of the objective function using the full data-set at once is not
only difficult but also counter productive.  Indeed, mini-batching serves other
purposes than just better scalability. For example, it helps prevent
`overfitting': Using all the data samples at once is similar to interpolating a
function in the presence of noise at all the data points.  Randomization also
helps the process escape from bad local minima.

This brings us to the main  problem namely the lack of convexity of the
objective functions invoked in deep learning.
The lack of convexity and the fact that the problem is heavily
over-parameterized means that there are many solutions to which the algorithms
can converge. Which one of these is better? 
If we consider only the objective function as the sole criterion,
one may think that the answer is clear: the lower the better. However,
practitioners in this field are more interested in `generalization' or the
property to obtain good classification results on data that is not among the
training data set.

In the past few years quite a few articles have appeared that were devoted to
understanding the nature of deep learning.  The problem of generalization in
particular has been the subject of numerous studies, see, e.g.,
\cite{zhang21_understanding,lossLandscape18,WeinanELandscape,WeinanE20} among
many others.  A puzzling character of neural networks is that they tend to do
quite well at classifiying items that do not belong to the training set.
However, the paper \cite{zhang21_understanding} shows by means of experiments
that looking at DL from the angle of minimizing the loss function fails to
explain these nice generalization properties. The authors show that they can
achieve a perfect loss of zero in training models on well-known datasets (MNIST,
CIFAR10) that have been modified by randomly changing all labels. In other words
one can obtain parameters whose loss function is minimum but with the worst
possible generalization since the resulting classification would be akin to
assigning a random label to each item.  A number of other papers explore this
issue further \cite{WeinanE20,InductiveBias14,lossLandscape18,WeinanELandscape}
by attempting to explain generalization with the help of the `loss landscape',
the geometry of the loss function in high dimensional space. What can be
understood from these works is that the problem is far more complex than just
minimizing a function.
Thus, the random nature of the optimization plays a central role.  There are many
minima and some are better than others. A local minimum that has a smaller loss
function will not necessarily lead to better inference accuracy. It is the random
character of the learning algorithms that  helps to achieve good generalization.

Another challenge  is that advanced optimization methods tend to be memory intensive,
requiring to store possibly tens of additional vectors to be effective.  In deep learning
this is not an affordable option. For example, a model like GPT3 has 175B
paramaters while  Llama3 involves 405B parameters.  This is the
primary reason why simple methods like SGD or Adam \cite{Adam-paper} are favored in
this context.
We will revisit optimization methods in Section~\ref{sec:opt}.

\subsection{Computational graphs and back-propagation}\label{sec:BProp} 
The roots of back-propagation emerged from the desire to develop
efficient software for optimization by `automatic differentiation'.
The idea itself emerged from different corners, see \cite{annotatedHist22} for
a detailed history. 
It may be argued that the impact of back-propagation is just as important
as that of high-performance hardware in the success of deep learning. 
In scientific computing, the idea of automatic differentiation, another name for 
back propagation,  was popularized by the work of  Andreas Griewank
\cite{Griewank89,griewankBook,Averick-al} among others. 

We start with the idea of computational graphs. These are directed  graphs, where 
vertices   represent  tasks that must be executed in the order
dictated by the directed edges of the graph:
An evaluation of a node will depend on other (incoming) nodes.
For example we may have a node that evaluates

\eq{eq:xplbp}
f(x,y,z) = g(a(x,y,z),b(x,y,z), c(x,y,x)) . 
\en
In this case, node $(f)$ will be
evaluated once nodes $(a)$, $(b)$  and $(c)$ have been calculated. 
As a trivial  example we can write the expression:
$ f(x,y,z) = (x+y-2)*(y+1) +2*z$ as
$ f(x,y,z) = g(a,b,c) $
where $g(a,b,c) \equiv(a*b +c)$,
and: $a(x,y,z) = x+y-2$; \ 
$b(x,y,z) = y+1$; \  $c(x,y,z) = 2*z $.

If  the graph terminates at a function $f$ at the root, we often have to 
(i) Evaluate the nodes  and 
(ii) the  derivatives of the root function $f$ with respect to the
primary variables $x,y, z$, for some set  values of $x,y, z$. 

\begin{figure}[htb]
\centerline{  \includegraphics[width=0.35\textwidth]{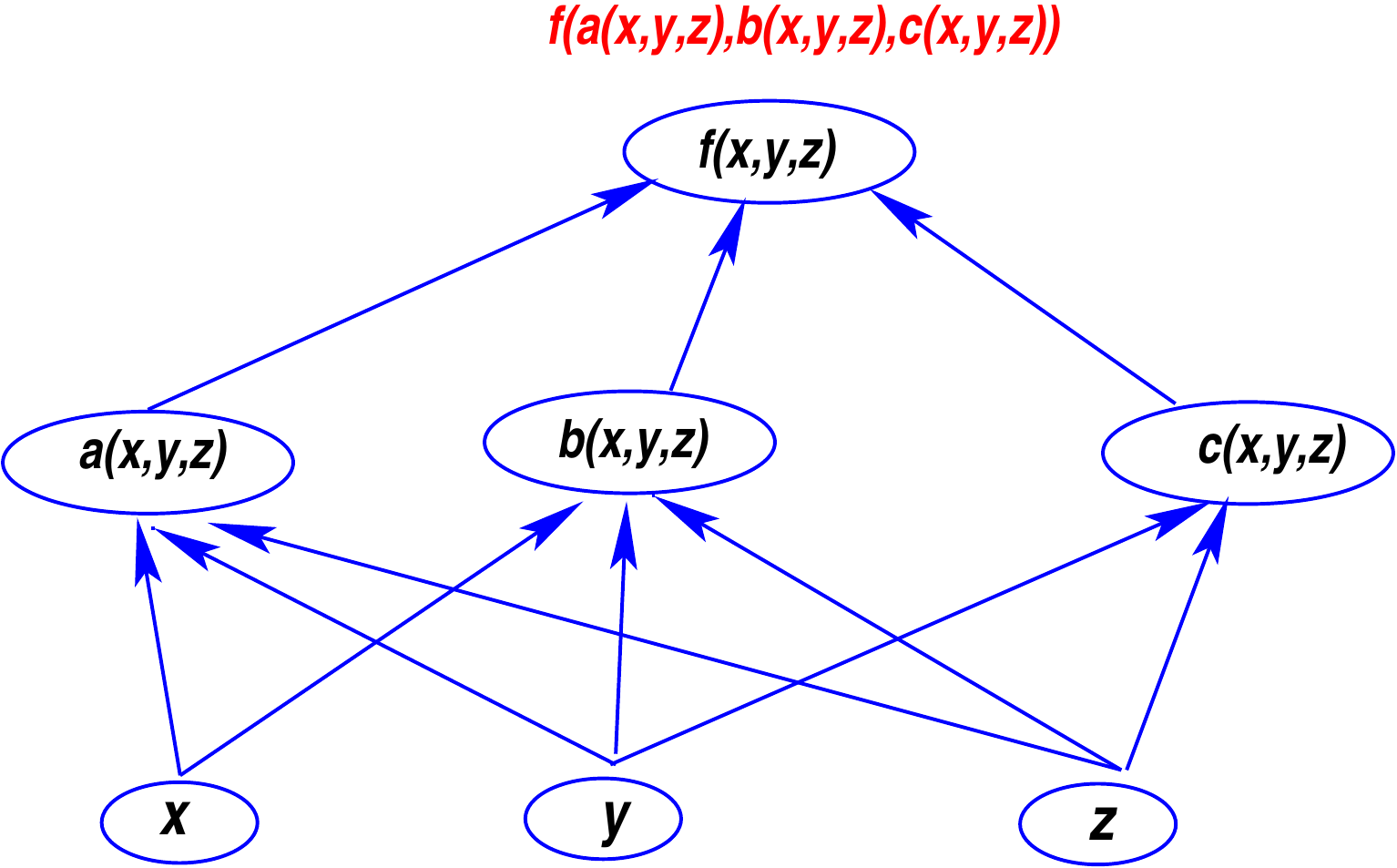} }
\caption{A simple computational graph for composite-function evaluation. The leaf nodes $x, y$, and $z$ are primary variables; the intermediate nodes $a(x, y, z), b(x, y, z)$ and $c(x, y, z)$ are derived quantities; and the root node is the target function. The figure illustrates how forward evaluation computes intermediate quantities first, while reverse-mode differentiation (back-propagation) propagates derivatives backward from $f$ to the leaves via the chain rule.}
\label{fig:compg0}
\end{figure}

For  part (i) we will just have to follow the 
graph up - starting from the input nodes. The order in which
the execution is performed for  a given graph is called the topological order.
For (ii) we will need to  use the chain rule.
For example assume that we have the expression \nref{eq:xplbp} where
$a, b, c$ are direct functions of the primary variables $x,y, z$  as shown in
Figure~\ref{fig:compg0}. Then the derivatives of $f$  with respect to
 $x, y, z$ will be: 
\begin{align} 
  \frac{\del f}{\del x}
  = & \frac{\del f}{\del a} \frac{\del a}{\del x} + \frac{\del f}{\del b} \frac{\del b}{\del x} +
      \frac{\del f}{\del c} \frac{\del c}{\del x} \label{eq:dfdx}      \\
  \frac{\del f}{\del y}
  = & \frac{\del f}{\del a} \frac{\del a}{\del y} + \frac{\del f}{\del b} \frac{\del b}{\del y} +
      \frac{\del f}{\del c} \frac{\del c}{\del y} \label{eq:dfdy}      \\
  \frac{\del f}{\del z}
  = & \frac{\del f}{\del a} \frac{\del a}{\del z} + \frac{\del f}{\del b} \frac{\del b}{\del z} +
      \frac{\del f}{\del c} \frac{\del c}{\del z} \label{eq:dfdz}    
\end{align}
With the above simple example as an illustration, the idea of the `reverse-mode'
differentiation results from the observation that it is computationally
convenient to compute the values of $a, b, c$ first (in a forward propagation)
before obtaining
$ \frac{\del f}{\del a},\frac{\del f}{\del b}, \frac{\del f}{\del c} $ and
finally $ \frac{\del f}{\del x}, \ \frac{\del f}{\del y} $ and
$ \frac{\del f}{\del z} $ with the help of (\ref{eq:dfdx}--\ref{eq:dfdz}),
by proceeding  backward from the root in a  second stage.
Note that the primary variables ($x,y,z$ in the example) are at the
leaves of the tree obtained by changing the directions of the edges,
see Figure~\ref{fig:compg0}.

In a general computational graph, functions like
$a, b, c$ are in turn functions of other functions located at
intermediate nodes in the graph. 
In this case, the back propagation phase would proceed backward to 
compute the partial derivatives of $f$ with respect to  
the intermediate nodes recursively until reaching the leaves. 

\begin{figure}[htb]
        \centerline{\includegraphics[width=0.25\textwidth]{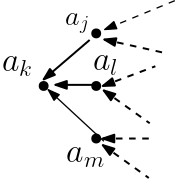} }
        \caption{Local back-propagation step at an intermediate node in a computational graph. Assuming the partial derivatives of the objective $f$ with respect to the downstream nodes $a_j, a_\ell$, and $a_m$ are already known, the derivative with respect to the current node $a_k$ is obtained by applying the chain rule and summing the contributions along all outgoing paths. The figure illustrates the recursive local structure that underlies reverse-mode automatic differentiation.}
        \label{fig:bp4}  
        \end{figure}

Consider now a generic  situation like the one shown in Figure~\ref{fig:bp4} illustrating a
back-propagation computation where we assume that the partial derivatives of $f$
with respect to the incident nodes $a_j, a_l, a_m$ have already been
computed. We now need to compute $\del f / \del a_k$ and this can be done with
the chain rule:

        \eq{eq:bpcr}
        \displaystyle \frac{\del f}{\del a_k} =
          \frac{\del f}{\del a_j} \frac{\del a_j}{\del a_k}  +
          \frac{\del f}{\del a_l} \frac{\del a_l}{\del a_k} +
          \frac{\del f}{\del a_m} \frac{\del a_m}{\del a_k}
          \en 
          Here, the nodes $a_i, i=1:n$ are tasks in a computational graph with the last node $a_n \equiv f$
          the target function.  The leaf nodes  $a_i, i=1,\cdots,e$ are the variables
          (e.g., $a_1 =x, a_2 = y, a_3 = z$ in the above example). The goal is to compute
          $\del f / \del a_1, \del f / \del a_2, \cdots, \del f / \del a_e$.
          
          The notation often used is to 
          let $\delta_k = \frac{\del f}{\del a_k} $ (called `errors').  Then
          back-propagation amounts to successively evaluating the $\delta_k$'s,
          following the graph back from the root (function $f$):
        \eq{eq:bpForm}  \displaystyle \delta_k =  \delta_j \frac{\del a_j}{\del a_k}  +
        \delta_l \frac{\del a_l}{\del a_k} +  \delta_m \frac{\del a_m}{\del a_k} .
        \en
        The nodes  $\delta_j,  \delta_l, \delta_m$ have been evaluated in earlier steps of
        back-propapation and the terms $\del a_i/ \del a_k$ are readily computable. 
        Note that the initial `error' corresponding to $f$ is $\delta_n = \del f / \del f  \equiv 1$.
        The leaves in the back-propagation graph correspond to the desired partial derivaties.

      
        It should be added  that in a general computational graph,
        there is an order in which to proceed in the back-propagation. Back-propagation is a textbook example of
        \emph{topological sorting}, an ordering of the nodes 
        in a directed acyclic graph (DAG).  Since a task cannot be started before
        its parent nodes in the graph have been processed we need to traverse the graph in
        topological order.

Finally, we note that the computation involved in (\ref{eq:dfdx}--\ref{eq:dfdz}) is a matrix-vector product
and this can also be understood from the propagation equation \nref{eq:bpForm}.  
If we write the gradient $\nabla f_{x,y,z} = [ \frac{\del f }{\del x}, \frac{ \del f }{\del y},
  \frac{\del f}{\del z}]$
  as a row-vector then the simple case of (\ref{eq:dfdx}--\ref{eq:dfdz}) can be written in matrix form as:
\[
  \begin{bmatrix}
    \frac{\del f}{\del x } &  \frac{\del f}{\del y} & \frac{\del f}{\del z}
  \end{bmatrix} = 
    \begin{bmatrix}
      \frac{\del f}{\del a } &  \frac{\del f}{\del b} & \frac{\del f}{\del c}
\end{bmatrix}
\times \begin{bmatrix}
  \frac{\del a}{\del x }&  \frac{\del a}{\del y} & \frac{\del a}{\del z} \\
  \frac{\del b}{\del x }&  \frac{\del b}{\del y} & \frac{\del b}{\del z} \\
  \frac{\del c}{\del x }&  \frac{\del c}{\del y} & \frac{\del c}{\del z}
\end{bmatrix} \ .
  \] 
   Denoting by $J_F$ the Jacobian of the mapping $(x,y,z) \to F = [a, b, c]^T $
  and by $ \nabla f_{a,b,c}$  the gradient of $f$ with respect to
its components $a,b,c$, then the above expresses the equality: 
  \eq{eq:BasicChain}
    \nabla f_{x,y,z}  = \nabla f_{a,b,c} \times J_F . 
\en

  The above discussion was for a standard case in which some function is
  defined recursively from other functions in a computational graph, in this
  case a tree where the primary variables $x,y,z$ are at the leaves of the final
  function $f$ is at the root. This scenario is common in optimization.
  Deep learning involves special computational graphs but the core idea utilized is the same
  as the one just described.

    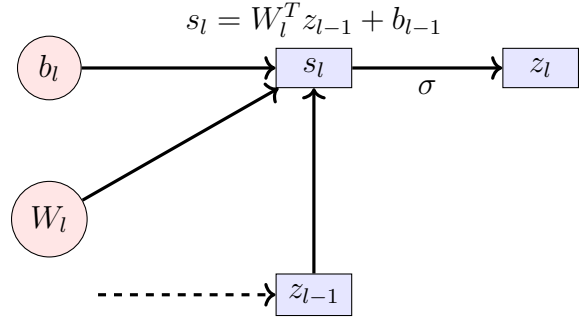
\begin{figure}[htb]
  \begin{center}
      \begin{tikzpicture}
    \draw (-1.5,1.0) node[circle,minimum width=1cm,draw,fill=red!10] (leaf1) {$W_l$}; 
    \draw (2.0,0.0)  node[minimum width=1cm,fill=blue!10,draw] (zl) {$z_{l-1}$};
    \draw (-1.5,3) node [circle, fill=red!10,draw] (leaf2) {$b_l$};
    \draw (2,3)  node[minimum width=1cm,fill=blue!10,
    label={$s_l=W_l^T z_{l-1} + b_{l-1}$},draw] (a1) {$s_l$};
    \draw (5.0,3)  node[minimum width=1cm,fill=blue!10,draw] (zl1) {$z_{l}$};
    \draw (-1,0) node [minimum width=0cm] (aa) {};  
    node (aa) at (-1.5,0) { } ; 
    \draw[->,very thick] (leaf1) -- (a1);
    \draw[->,very thick] (leaf2) -- (a1);
    \draw[->,very thick] (zl) -- (a1);
    \draw[->,very thick] (a1) --  (zl1)  node[midway,below] {$\sigma$ };
    \draw[->,dashed,very thick] (aa) --  (zl);  
  \end{tikzpicture}
\end{center}
  \caption{A local computational-graph view of one neural-network layer. The figure shows how the pre-activation $s_\ell = W_\ell^T z_{\ell-1} + b_\ell$ is formed from the previous-layer representation $z_{\ell-1}$, the weight matrix $W_\ell$, and the bias $b_\ell$ and is then mapped by the activation function $\sigma$ to produce $z_\ell$. It illustrates how each layer of a neural network can be interpreted as a small computational subgraph whose variables participate in forward evaluation and back-propagation. The circled nodes represent input or parameter variables, whereas the rectangular nodes represent computational operations or derived quantities.}
  \label{fig:compnode}
\end{figure}
 
In neural networks, the variables are the matrices $W_l$ that link the different
layers in the transformation represented by \nref{eq:zl} and the biases $b_l$. These
appear at each layer.  Suppose we
have just two hidden layers and that the cost function is the simple regression:
\eq{eq:simpl}
f((x,y),\Theta) = \eta \equiv \half \| \hat y - z_3 \|_2^2  . 
\quad \text{with} \quad z_k = \sigma( W_{k}^T z_{k-1} + b_k) \quad k=3, 2, 1
\en
in which we recall that $z_0=x$ the input. The set of all parameters is often denoted by
$\Theta $. With two hidden layers 
we will have three sets of parameters $\theta_1 = \{W_1, b_1\}$ (layer 0 to 1),
$\theta_2 = \{W_2, b_2\}$ (layer 1 to 2) and $\theta_3=\{W_3, b_3\}$ (layer 2 to 3).
We will have the intermediate functions
\begin{align}
  s_1 =& s_1(z_0, \theta_1) = W_1^T z_0 + b_1   \label{eq:chain1} \\
  s_2 =& s_2(z_1, \theta_2) = W_2^T \sigma(s_1)  + b_2 \label{eq:chain2} \\
  s_3 =& s_3(z_2, \theta_3) = W_3^T \sigma(s_2)  + b_3 \label{eq:chain3} \\ 
  \eta =& \half \| \sigma(s_3) -  \hat y \|_2^2 \label{eq:chain4}
\end{align}

Then the back-propagation procedure will compute the partial derivatives of $\eta$ at all the nodes,
namely, $s_3,s_2,s_1, W_3, W_2, W_1, b_3,b_2,b_1,$ as follows:
\begin{align*}
  \text{Compute:} & &  \frac{\del \eta }{\del s_3} & & \text{from \nref{eq:chain4}}
& & \text{then} &&
    \frac{\del \eta }{\del W_3} = \frac{\del \eta }{\del s_3} \frac{\del s_3 }{\del W_3}
    \ , \quad 
    \frac{\del \eta }{\del b_3} = \frac{\del \eta }{\del s_3} \frac{\del s_3 }{\del b_3}
 & & \text{from \nref{eq:chain3}} \\  
  \text{Compute:} & &  \frac{\del \eta }{\del s_2} =  \frac{\del \eta }{\del s_3}
 \frac{\del s_3}{\del s_2}  & & \text{from \nref{eq:chain3}}
& & \text{then} && 
    \frac{\del \eta }{\del W_2} = \frac{\del \eta }{\del s_2} \frac{\del s_2 }{\del W_2}
    \ , \quad
    \frac{\del \eta }{\del b_2} = \frac{\del \eta }{\del s_2} \frac{\del s_2 }{\del b_2}
  & & \text{from \nref{eq:chain2}} \\
  \text{Compute:} & &  \frac{\del \eta }{\del s_1} =  \frac{\del \eta }{\del s_2}
 \frac{\del s_2}{\del s_1}  & & \text{from \nref{eq:chain2}}
& & \text{then} &&
    \frac{\del \eta }{\del W_1} = \frac{\del \eta }{\del s_1} \frac{\del s_1 }{\del W_1}
    \ , \quad
    \frac{\del \eta }{\del b_1} = \frac{\del \eta }{\del s_1} \frac{\del s_1 }{\del b_1}
  & & \text{from \nref{eq:chain1}}    
\end{align*}

In Algorithm~\ref{alg:BP} we assume that the function $\eta$ is of the form
$\eta(y,\hat y)$ and that we have $L$ layers. At each layer we have an update of
the form \eq{eq:sip1} s_{i+1} = W_i^T \sigma(s_i) + b_i, \quad i=0, \cdots, L
\en with $s_0 = x$. There are three types of derivatives:
$\del \eta (y,\hat y) / \del W_l $, $\del \eta (y,\hat y) / \del b_l $ and
$\delta_l = \del \eta (y,\hat y) / \del s_l $ (referred to as `errors' earlier).

\begin{algorithm}[htb]
  \centering
  \caption{Backward Propagation}\label{alg:BP}
  \begin{algorithmic}[1]
    \State Assume: Forward propagation carried out
    \State Compute gradient $\delta_L \gets \frac{\del \eta(y,\hat y)}{\del s_{L}} $
    \For{$l=L: 1 $}   
    \State
    Compute $ \frac{\del \eta(y,\hat y)}{\del W_l}  = 
    \frac{\del \eta(y,\hat y)}{\del s_l} \times   \frac{\del s_l}{\del W_l} = \delta_l   \frac{\del s_l}{\del W_l}$
    \Comment{Gradient w.r.t. $W_l$ in level $l$ }
    \State
    Compute $ \frac{\del \eta(y,\hat y)}{\del b_l}  = 
    \frac{\del \eta(y,\hat y)}{\del s_l} \times   \frac{\del s_l}{\del b_l} = \delta_l   \frac{\del s_l}{\del b_l}$
    \Comment{Gradient w.r.t. $ b_l$ in level $l$ }
    \State $\delta_{l-1} =  
     \frac{\del \eta(y,\hat y)}{\del s_{l}}  \frac{\del s_{l}}{\del s_{l-1}}  = \delta_{l}\frac{\del s_{l}}{\del s_{l-1}}$
    \Comment{Back-propagate `error' to layer $l-1$ }
    \EndFor 
\end{algorithmic}
\end{algorithm}

         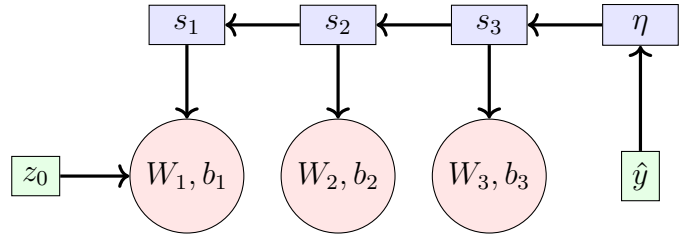
\begin{figure}[tb]
  \begin{center}
      \begin{tikzpicture}
        \draw (1.0,0) node[circle,minimum width=.5cm,draw,fill=red!10] (th1) {$W_1,b_1$};
        \draw (3.0,0) node[circle,minimum width=.5cm,draw,fill=red!10] (th2) {$W_2,b_2$};
        \draw (5.0,0) node[circle,minimum width=.5cm,draw,fill=red!10] (th3) {$W_3,b_3$};
        \draw (-1.0,0.0) node[minimum width=.5cm,draw,fill=green!10] (z0) {$z_0$};
        \draw (1.0,2.0)  node[minimum width=1cm,fill=blue!10,draw] (a1) {$s_1$};
        \draw (3.0,2.0)  node[minimum width=1cm,fill=blue!10,draw] (a2) {$s_2$};
        \draw (5.0,2.0)  node[minimum width=1cm,fill=blue!10,draw] (a3) {$s_3$};
        \draw (7.0,0.0) node[minimum width=.5cm,draw,fill=green!10] (yh) {$\hat y$};
        \draw (7.0,2.0) node[minimum width=1cm,draw,fill=blue!10] (eta) {$\eta$}; 
        \draw[->,very thick] (z0) -- (th1);
        \draw[->,very thick] (yh) -- (eta);
        \draw[<-,very thick] (a2) -- (a3);
        \draw[<-,very thick] (a1) -- (a2);
        \draw[<-,very thick] (a3) -- (eta);
        \draw[<-,very thick] (th1) -- (a1);
        \draw[<-,very thick] (th2) -- (a2);
        \draw[<-,very thick] (th3) -- (a3);        
  \end{tikzpicture}
\end{center}
\caption{Backward propagation in a multilayer perceptron. The figure schematically illustrates the reverse flow of derivatives through the layered computational graph associated with equations \eqref{eq:chain1}--\eqref{eq:chain4}. Starting from the loss $\eta$, sensitivities are propagated backward through the intermediate states $s_3, s_2, s_1$ in order to compute gradients with respect to the parameters $(W_3, b_3), (W_2, b_2)$, and $(W_1, b_1)$.}
\label{fig:bpe}
\end{figure}

We wish to make a point  about notation. The partial derivatives
like $\del \eta / \del s_l$ are the gradients of
$\eta$ with respect to components of the vector $s_l$.  As before we may
think of these as row vectors.
Similarly for the terms  $\del \eta /  \del b_l$.
Each term $\del s_l / \del b_l$ is the Jacobian of the function $s_l$ when
viewed as a function of the components of $b_l$. Thus, the expressions
$ \del \eta / \del b_l = (\del \eta / \del s_l )\times (\del \eta / \del s_l ) $
correspond to Equation \nref{eq:BasicChain} seen earlier.

Consider now  the terms in the equation
$\del \eta / \del W_l = (\del \eta / \del s_l ) \times (\del s_l/  \del W_l)$
and recall that $W_l \in \ \RR^{d_{l-1}\times d_l}$.
We start with  the case where $d_l = 1$ so that  $W_l$ is a vector in
$\RR^{d_{l-1}}$.
In this situation  $\del \eta / \del W_l$ is the desired gradient
with respect to the components of $W_l$  while the term $\del \eta / \del s_l$ is
the (previously computed) gradient of $\eta $ with respect to  $s_l$
and  $\del s_l / \del W_l = \sigma(s_{l-1})^T$ per \nref{eq:chain2}.
As can be seen the situation is rather similar to that of $b_l$.
When $d_l>1$, $W_l$ is a matrix in $\RR^{d_{l-1}\times d_l}$ and 
we can deal with this general case by
 thinking of the matrix $W_l$ as a vector of size 
$d_l \cdot  d_{l-1}$ consisting of the stacked columns of $W_l$.
Note that the partial derivative of $s_l $ with respect to each column of
$W_l$ is the same and it is equal to $\sigma(s_{l-1})^T$.

\subsection{Tensor Computations in DNNs}

Computations in Neural Networks rely heavily on matrix and tensor
computations. This was understood early on and has led in particular to the
development and exploitation of adapted hardware to speed-up training.  Packages
such as TensorFlow, Pytorch, and Jax, utilize tensor computations internally.
By default an object in these packages is a tensor in the same way that
variables in Matlab are viewed as  matrices by default.

Tensor computations can start from the beginning when the input data itself
is in the form of tensors. For example, when dealing with pictures we may have
3-D tensors ($\texttt{height}$, $\texttt{width}$ coordinates of pixels,
along with color chanels).
In addition the data is often batched, so in this case each batch is a 4-Tensor
(batch size, height of image in pixels, width of image in pixels, color channels).
In convolution neural networks (CNNs), a sliding `filter' is applied to pieces of
a picture that are small 3-D tensors. This leads to several 
opportunities for exploiting tensor computations during training.

In many cases, there is no loss in flattening tensors and it is often convenient to do
so in order to end up with matrix computations. It may also be required by the
nature of the network.  For example, fully connected layers (see section on MLP)
expect 2-D tensors with the shape $\text{batch\_size} \times d_l$ where $d_l $
is the dimension of the features in layer $l$. In this case, tensors are
flattened accordingly.  Thus, a batch of images of $32 \times 32$ pixels of RGB
channels will have size $ (batch\_size, 3, 32, 32)$. In CNN, applying 128
$3 \times 3 $ filters with a stride of one will result in a shape of
$ (batch\_size, 128, 32, 32)$. After pooling we may get a size of
$ (batch\_size, 128, 16, 16)$. Repeating this twice will convert the original
tensor into a $(batch\_size, 128, 4, 4)$ tensor which is finally flattened into
a tensor of size $(batch\_size, 128\times 4 \times 4)$ = $(batch\_size, 2048)$
because the last layer of CNN is a dense (fully connected) layer. Note that such
dense layers don't know how to handle spatial structure since they just connect
every input to every output.  Most of the work of CNNs is done prior to
flattening to exploit spatial structure.


In other cases tensors are kept in their original shape or at least as tensors
with more than 2 dimensions. In the example just seen,
CNNs act on  4D tensors  in all but the last (dense) layer.
Convolutions and pooling operations expect these spatial dimensions.
Similarly in Language models (RNN, LSTM, Transformers)
the sequence data has the shape: $(batch\_size, sequence\_length, embedding\_dim)$
and no  flattening is performed. This is because positional or temporal structure
must be preserved.

\subsection{NA  thinking vs. ML/AI thinking: Attention}

The notion of attention plays a key role in transformers as might be
inferred  from the intentionally provocative  title
\emph{``Attention is all you need''} of the original
article \cite{vaswani2017attention}.
In this section we will illustrate the notion of attention with a simple
example that contrasts the Numerical Analysis way vs the machine learning
way of solving a simple problem: approximating a function.

The example is as follows. We are  given very noisy `training points'
$x_i, y_i$ which are x-coordinates and y-coordinates of some  unknown
function $f$, at specific points. The goal is to  `recover' $f$ in some
form. The simplest \emph{numerical analysis} approach to the problem
is to use the data points to
interpolate the function in  the Least-Squares sense.
This involves  selecting the type of interpolating type, for example
cubic polynomial. Often, but not always,  we do know the form of the function
in advance. 

\begin{figure}
\centerline{
\includegraphics[width=0.6\textwidth]{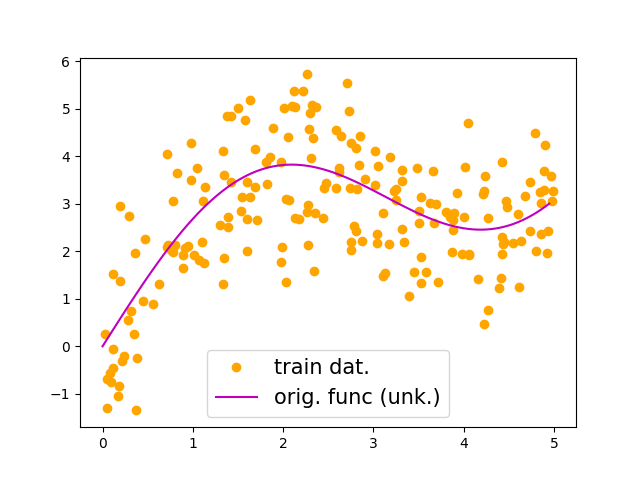} 
}
 \caption{Recovering an underlying function from noisy data. The figure illustrates the basic problem used to motivate attention: given a cloud of noisy training points $(x_i, y_i)$, estimate the value of an unknown function at a new query point. In the numerical-analysis viewpoint this suggests interpolation or least-squares approximation, whereas in the machine-learning viewpoint it motivates weighted averaging based on relevance, i.e., attention.}
 \label{fig:myplot1} 
\end{figure}

The \emph{Machine Learning} solution to the problem relies
entirely on the given data points. One approach is to use
an advanced  form of averaging with `attention'.
The idea of attention comes from databases.
We are given known key and value pairs 
$\{ k_i, v_i\}$ and would like to guess a value for a certain query
$q$ which is of the same type as  the keys $k_i$  in the database. 
In the example given above, the key-value pairs are the pairs
($x^{train}_i, y^{train}_i \}$) of the given data.
In the same example, $q$ is an arbitray $x$-coordinate where we want to 
calculate an  approximate value of $f$. A naive first solution  would be to take the closest
$k_l$ to $q$ and declare the value of $f$ at $q$ to be $v_l$.
A better solution is to take some weighted average of all values $v_i$,
with weights defined so as to give more importance (attention) to more
relevant training points.

\begin{figure}
\centerline{
  \includegraphics[width=0.60\textwidth]{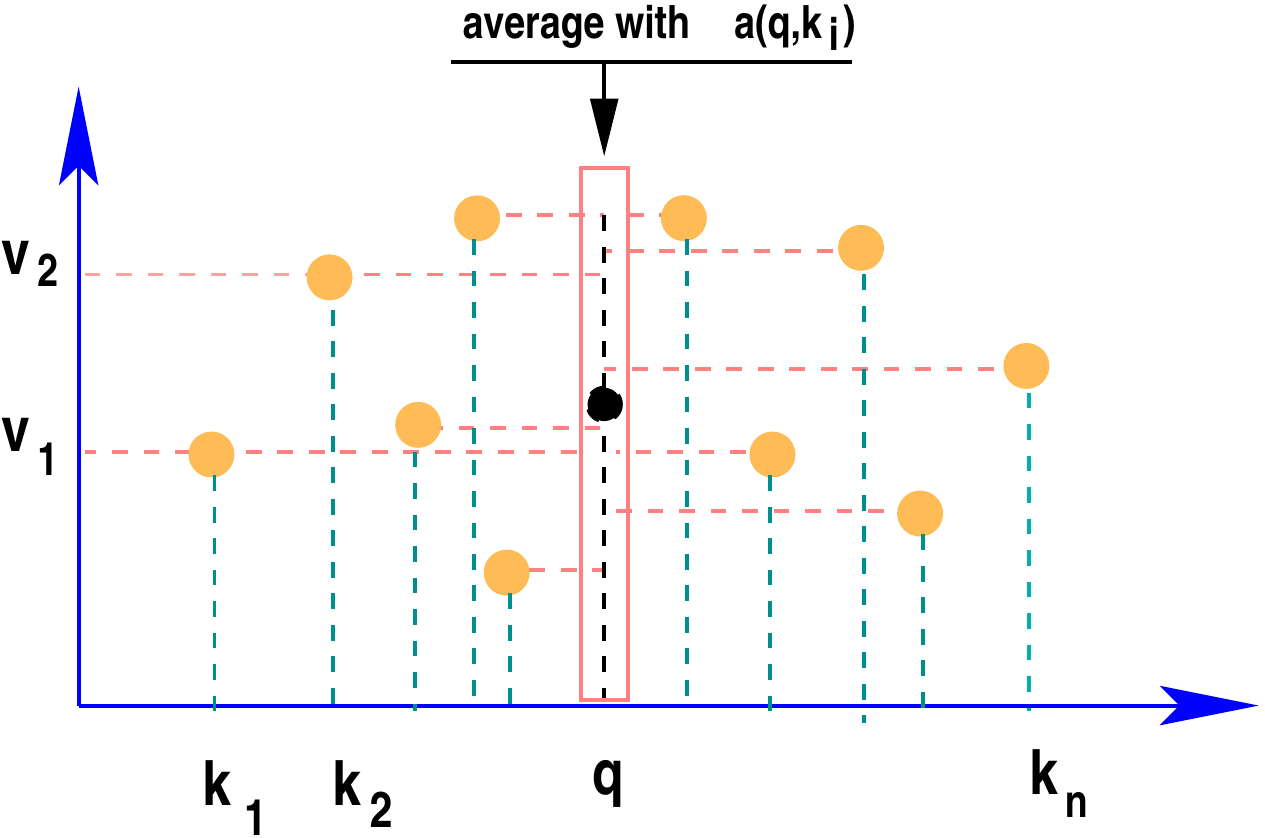}}
\caption{Illustration of Nadaraya–Watson kernel regression. Given a query point $\bm{q}$, the prediction is formed as a weighted average of the observed values ${\bm{v}}_i$, with weights that depend on the similarity between $\bm{q}$ and the keys ${\bm{k}}_i$. This provides a useful bridge to attention, where output representations are likewise obtained by similarity-based weighted averaging.}
\label{fig:NW} 
\end{figure}

\begin{figure}
\centerline{
  \includegraphics[width=0.60\textwidth]{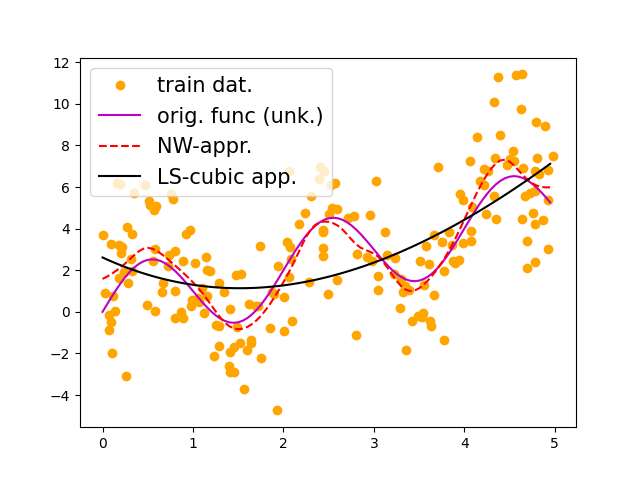}}
\caption{Nadaraya–Watson kernel regression on periodic data. The figure compares an attention-based kernel-regression estimate with a classical low-degree approximation for an oscillatory target function. In this setting, the similarity-based weighted average captures the periodic behavior more accurately than a simple cubic approximation, illustrating how attention can recover useful structure directly from data when the functional form is not known in advance.}
\label{fig:NW2} 
\end{figure}

We can define an \emph{``attention''} mechanism that 
averages the values by giving more importance to points located  near $q$ in
a number of ways by exploiting  a  Kernel $a(q,k)$.
There are a number of options for the kernel $a$, one of which is
the  Gaussian kernel  expressed below for the
more general case where the $k_i$'s, and $ q $ are in $\RR^d$:
\eq{eq:attGauss}
a(q,k_i) = \frac{\exp (- \half \|q - k_i\|^2/\sigma^2 )}
{\sum_{l=1}^n \exp (- \half \|q - k_l\|^2/\sigma^2 )} 
\en
Observe that the values of $a(q,k_i)$ are positive and scaled so that they
add-up to unity and so they play the role of probabilities.  The inferred value
$v$ for $q$ is then given by:
\[
  \sum_{i=1}^n a(q,k_i) v_i . 
\]
This process is a form of Kernel Regression \cite{zhang2023dive,murphyBook2022}
known as Nadaraya-Watson
attention. It is illustrated in Figure~\ref{fig:NW}.
An example is given in Figure~\ref{fig:NW2} for a function
that is oscillatory. If we did not know the nature of the function we would approximate it with
a cubic. As can be seen this results in a poor approximation in this case.
Here, the attention-based approximation does a much better job.
The main point of this illustration  is that with a lot of data, we can
succeed in obtaining a good approximation to an unknown function by using
attention. 

\section{Large Language Models and Transformer Architecture}
Large language models are built on the machinery of deep neural networks reviewed in the previous section, but they introduce architectural and conceptual features that deserve a separate treatment. 
Transformers changed the field. In particular, self-attention and large-scale autoregressive training moved representation, sequence modeling, scaling, and optimization to the center of modern machine learning. In this section, we identify the mathematical ingredients that make large language models work. We keep the focus narrow: linear algebra. We use that lens to describe how Transformers are designed, trained, and analyzed.

We adopt the convention that scalars are denoted by italic letters, vectors by bold lowercase letters, and matrices by bold uppercase letters. 
At layer $\ell$, the sequence representation is the matrix ${\bm{X}}_{\ell} \in {\mathbb{R}}^{n\times d}$, whose $t$-th row ${\bm{x}}_t^{\ell} \in {\mathbb{R}}^d$ is the representation of token position $t$. 
Projection matrices are denoted by $\bm{W}$ with suitable subscripts.
We use $t$ for token position, $\ell$ for layer index, and $h$ for attention-head index.
\subsection{Language modeling and the rise of LLMs}
A language model assigns probabilities to token sequences and is the basic probabilistic object underlying many natural-language-processing tasks.
Given a sequence of tokens $( u_1,\dots,u_{t-1} )$, a language model gives the probability of the next token $u_t$ conditioned on the preceding context, i.e.,

\begin{equation}
p_{\mathrm{LM}}(u_t \mid u_1,\dots,u_{t-1}).
\label{equa2-1}
\end{equation}
%

%
Language modeling involves two closely related tasks: assigning probability to token sequences and generating continuations from those probabilities. 
Associated with the model is a vocabulary ${\mathcal{V}}$ of size $|{\mathcal{V}}| = n_{\text{vocab}}$, whose elements are tokens such as words or sub-words.
Given a context $u_{<t}=(u_1,\dots,u_{t-1})$, the model outputs a probability distribution over the next token $u_t$.

The probability of the sequence $(u_1,\dots,u_t)$ can be defined autoregressively using the chain rule
\begin{equation}
p_{\mathrm{LM}}(u_1,\dots,u_t)
=
\prod_{i=1}^{t} p_{\mathrm{LM}}(u_i \mid u_1,\dots,u_{i-1}),
\label{equa2-2}
\end{equation}
or alternatively in logarithmic form
\begin{equation}
\log p_{\mathrm{LM}}(u_1,\dots,u_t)
=
\sum_{i=1}^{t} \log p_{\mathrm{LM}}(u_i \mid u_1,\dots,u_{i-1}).
\label{equa2-3}
\end{equation}
Thus, autoregressive language modeling reduces sequence modeling to the estimation of a sequence of conditional distributions.

\begin{figure}[h]
\begin{center}
\includegraphics[scale=0.4]{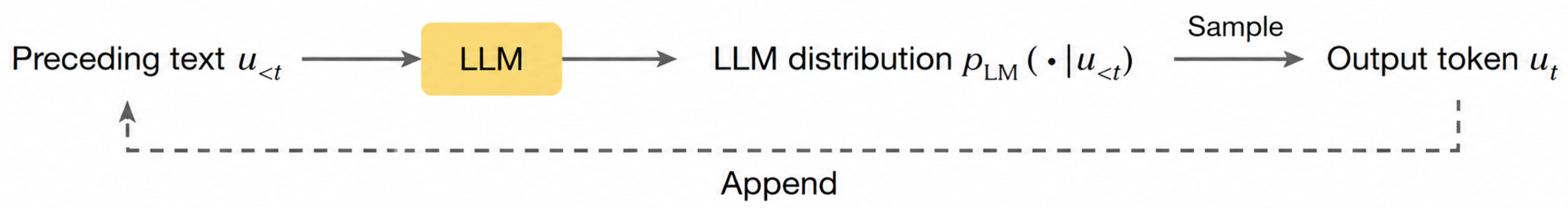}
\end{center}
\caption{High-level view of autoregressive language modeling. Given a context $u_{\,<\,t} = (u_1, \dots, u_{t-1})$, the model assigns a probability distribution to the next token $u_t$, and text generation proceeds by repeating this prediction step sequentially.}
\label{fig:text_generation}
\end{figure}

In practice, the conditional probabilities in \eqref{equa2-1}--\eqref{equa2-3} are estimated by a deep neural network. 
At inference time, one may either select the most likely next token or sample from the conditional distribution. 
The first viewpoint leads to a deterministic prediction rule,
\begin{equation}
\widehat{u}_t
=
\arg\max_{u \,\in\, {\mathcal{V}}}
p_{\mathrm{LM}}(u \mid u_1,\dots,u_{t-1}) ,
\label{equa2-4}
\end{equation}
whereas the second leads to a stochastic generation procedure,
\begin{equation}
u_t \sim p_{\mathrm{LM}}(\,\cdot \mid u_1,\dots,u_{t-1}) ,
\label{equa2-5}
\end{equation}
which is then iterated autoregressively by appending each predicted or sampled token to the context before predicting the next one. 
Thus a language model is both a probabilistic model of sequences and the basis of a generation procedure.
A schematic of this autoregressive prediction-and-generation loop is shown in Figure~\ref{fig:text_generation}.
This distinction between assigning probability to sequences and generating new sequences from those probabilities is central to modern large language models.
%
%
%

%
The training objective for language modeling is the cross-entropy loss, equivalently the negative log-likelihood of the training sequence:
\begin{equation}
\mathcal{L}_\Theta
=
\frac{1}{n}
\sum_{t=1}^{n}
-\log p_{\mathrm{LM}}(u_t \mid u_1,\dots,u_{t-1}).
\label{equa2-6}
\end{equation}
This objective is the standard next-token prediction loss used in autoregressive Transformer language models and remains the core pretraining objective of modern decoder-only large language models.

Large language models (LLMs) are language models trained at a scale that makes them useful across a wide range of tasks. 
The Transformer architecture has become the backbone of virtually all state-of-the-art large language models, owing to its efficient, highly parallelizable training and its ability to handle long-range dependencies. In this sense, an LLM is not defined by a different probabilistic objective from that of a classical language model, but by the scale of the model, the training data, and the computational resources used to optimize it.
The key architectural point is that modern LLMs are overwhelmingly based on decoder-only Transformers trained autoregressively. They take a prompt $u_1,\dots,u_{t-1}$, form contextual hidden representations through many stacked Transformer blocks, and then predict a distribution over the next token $u_t$. This same mechanism is iterated during generation to produce coherent continuations over long contexts. 
Thus the rise of LLMs is best understood as the rise of large-scale autoregressive Transformer language models \cite{vaswani2017attention, brown2020gpt3}.

A central distinction is that between training and inference.
During training, the entire token sequence is known, and the model can evaluate the loss in \eqref{equa2-6} at all positions in parallel, subject to the causal masking that prevents each token from using future information. During inference, by contrast, future tokens are not yet available, so generation is inherently sequential: each newly predicted token must be appended to the context before the next token can be produced. This distinction is one of the central computational ideas behind Transformer-based language modeling: the probabilistic model is autoregressive, but training is still highly parallelizable. 

This also clarifies what historically changed with the rise of LLMs. The language-modeling objective itself did not fundamentally change; what changed was the architectural and computational regime in which that objective could be optimized. The Transformer made it possible to train very large autoregressive models efficiently, and when combined with scale in data and parameters this led to the modern large-language-model era, see e.g., \cite{brown2020gpt3, kaplan-al20}.

\subsection{Tokens, embeddings, and positional structure}
\label{sec:tokens}
Associated with any language model is a vocabulary ${\mathcal{V}}$, a finite set of tokens of size $| {\mathcal{V}} | = n_{\mathrm{vocab}}$. 
A token may correspond to a word, a sub-word, or another discrete textual unit, depending on the tokenization scheme.

Consider a token sequence $( u_1,u_2,\dots,u_n )$, where $u_i \in {\mathcal{V}}$ for $i=1, \cdots, n$.
A standard representation associates to each token $u_i$ a one-hot vector in $\mathbb{R}^{|{\mathcal{V}}|}$, and stacks these vectors row-wise into a matrix
\begin{equation}
{\bm{U}} \in \mathbb{R}^{n\times |{\mathcal{V}}|} ,
\qquad
{\bm{U}}_{i, \colon} = {\bm{e}}_{u_i}^\top ,
\qquad i = 1, \cdots, n ,
\label{equa3.2}
\end{equation}
where ${\bm{e}}_{u_i} \in {\mathbb{R}}^{|\mathcal{V}|}$ is the canonical basis vector associated with token $u_i$. 
Hence, each row of $\bm{U}$ specifies one token in the sequence relative to the vocabulary $\mathcal{V}$.

At this stage, the representation is purely symbolic and contains no notion of semantic similarity. Two tokens are distinct basis vectors in \(\mathbb{R}^{|{\mathcal{V}}|}\), even if their meanings are closely related. 
The role of the embedding map is to replace this sparse symbolic representation by a dense learned representation in a much lower-dimensional vector space.
%
\begin{figure}[htb]
\centerline{
\begin{tikzpicture}[
    >=Stealth,
    blkg/.style={fill=green!30!gray!30,align=center,minimum width=15mm},
    blky/.style={fill=yellow!50,align=center,minimum width=15mm},
    add/.style={font=\Large, inner sep=0},
 ]
\definecolor{blue2}{rgb}{0,0,0.69}
\definecolor{cyan3}{rgb}{0,0.82,0.82}
\definecolor{red3}{rgb}{0.82,0,0}
\definecolor{mrn}{rgb}{0.867,0.667,0.533}
\pgfsetlinewidth{2pt}
\node[fill=red!20](W0)   at (0, 0) {$W_0$ \textbf{(Frozen)}};

\pgfsetstrokecolor{blue}
\pgfsetfillcolor{mrn}

\Large
\boldmath 

\filldraw[fill=green!30!gray!20,very thick] (-2,-1) -- (-2,2) -- (2,2) -- (2,-1) -- (-2,-1);

\node (w0) at (0.12,1.25) {$W_0 \ \textcolor{red}{(d \times d_v)}$};
\node (fr) at (0.15,0.25) {\textbf{(Frozen)}};
\node [rectangle,draw,fill=gray!10,very thick] (c) at (2.75,4.5)  {$h = x \times W_0 + x \times A B$};

\draw[very thick, black][->]  (0.0, 2.0) -- (2.5,3.5) ;
\draw[very thick, black][->]  (5.7, 2.4) -- (3.0,3.5) ;

\filldraw[fill=mrn!30,very thick] (3.5,-1) -- (3.5,2) -- (3.8,2) -- (3.8,-1) -- (3.5,-1);
\filldraw[fill=mrn!30,very thick] (3.9,2) -- (3.9,2.3) -- (7.5,2.3) -- (7.5,2) -- (3.9,2);
\node (rr) at (6.0,0.5) {\textbf{(Trained)}};
\node (aa) at (5.4,-0.8) {$A \ \textcolor{red}{ (d \times r)}$};
\node (bb1) at (7.8,2.2) {$B$};
\node (bb2) at (5.8,1.6) {$\textcolor{red}{(r \times d_v)}$};
\node (op) at (2.75,3.6) {$\oplus$};


\filldraw[fill=blue2!60,thick] (0.75,-2.5) -- (.75,-2.3) -- (4.75,-2.3) -- (4.75,-2.5) -- (.75,-2.5);
\node (xx) at (6.55,-2.405) {$x \quad \textcolor{red}{ (1 \times d)}$};

\draw[very thick, black][->]  (2.73, -2.2) -- (0.0,-1.05) ;
\draw[very thick, black][->]  (2.77, -2.2) -- (3.65,-1.05) ;

\draw[dashed,very thick, red]  (2.73, -1.5) -- (2.73,3.0) ;

\end{tikzpicture}}
\caption{Schematic pipeline of a decoder-only large language model. Tokens are mapped to embeddings with positional structure, propagated through repeated Transformer blocks, and finally projected to logits for next-token prediction.}
\label{fig:fg1}
\end{figure}
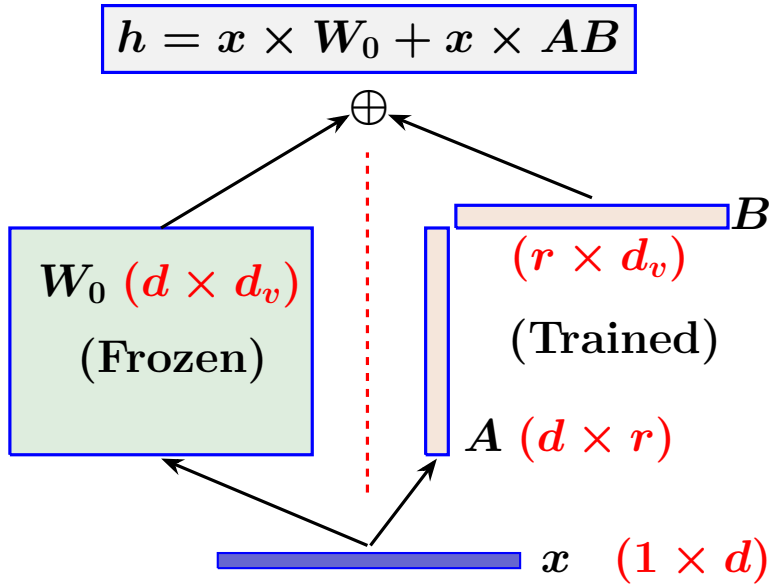

The first step in a Transformer is to associate each token with a learned vector of dimension \(d_{\mathrm{model}}\), called its embedding. At this stage, the embedding is context-independent: each token is mapped through the same learned lookup table regardless of the surrounding sequence.

Let
\begin{equation}
{\bm{W}}_E \in \mathbb{R}^{|{\mathcal{V}}|\times d_{\mathrm{model}}}
\label{equa3.3}
\end{equation}
denote the token embedding matrix. Then, under the row-vector convention, the embedding of token $u_i$ is
\begin{equation}
{\bm{U}}_{i,\colon} {\bm{W}}_E  \;=\; {\bm{e}}_{u_i}^\top {\bm{W}}_E    \;\;\in \mathbb{R}^{d_{\mathrm{model}}} ,
\label{equa3.4}
\end{equation}
and equivalently, stacking all tokens together yields the matrix of token embeddings
\begin{equation}
{\bm{U}} {\bm{W}}_E \in \mathbb{R}^{n\times d_{\mathrm{model}}}.
\label{equa3.5}
\end{equation}

This matrix multiplication is the first important linear-algebra step in the architecture. The one-hot token matrix $\bm{U}$ selects rows of ${\bm{W}}_E$, and the result is a dense matrix whose rows are learned token vectors. Thus, token embedding is a learned linear projection from the vocabulary basis into the model’s representation space.
This overall input-to-output pipeline is summarized schematically in Figure~\ref{fig:fg1}.

A naive self-attention mechanism is insensitive to token order: if the input rows are permuted, the output rows are permuted in the same way. In language, however, word order carries meaning, whereas the token embedding matrix ${\bm{U}} {\bm{W}}_E$ encodes token identity but not token order. Positional structure is therefore essential: it is what distinguishes an ordered sequence from an unordered set of tokens.

\begin{figure}[t]
\begin{center}
\includegraphics[scale=0.35]{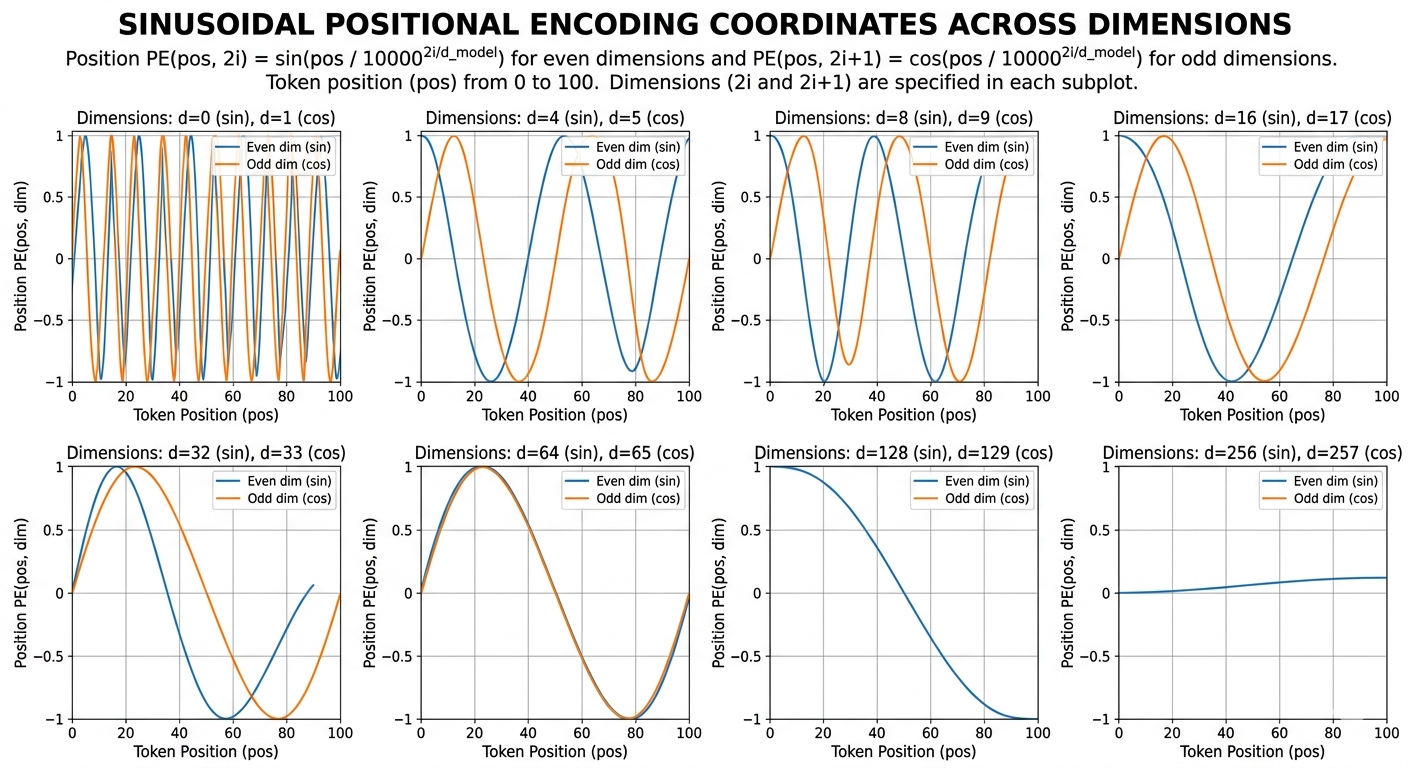}
\end{center}
\caption{Sinusoidal positional encoding coordinates across dimensions. Each subplot shows the two coordinates ${\bm{P}} (i,2j) = \sin \left( i / 10000^{2j/d_{\text{model}}} \right)$ and ${\bm{P}} (i,2j+1) = \cos \left( i / 10000^{2j/d_{\text{model}}} \right)$ as functions of token position $i$ for a representative pair of dimensions. Low-index dimensions vary rapidly with position, while higher-index dimensions vary more slowly, so the encoding provides a multiscale representation of sequence order.}
\label{fig:sinusoidal_position}
\end{figure}

In the original Transformer \cite{vaswani2017attention}, positional information is incorporated through additive absolute positional encodings. Let
\begin{equation}
{\bm{P}} \in \mathbb{R}^{n\times d_{\mathrm{model}}}
\label{equa3.6}
\end{equation}
denote the positional encoding matrix. Then the initial input to the Transformer is
\begin{equation}
{\bm{X}}_0 = {\bm{U}} {\bm{W}}_E + {\bm{P}} .
\label{equa3.7}
\end{equation}
Thus, positional information enters the model at the level of the initial representation matrix, before any attention computation takes place.
The sinusoidal positional encoding used in the original Transformer is defined by
\begin{align}
{\bm{P}} (i,2j)
&=
\sin\!\left(
\frac{i}{10000^{2j/d_{\mathrm{model}}}}
\right),
\label{equa3.8}
\\
{\bm{P}} (i,2j+1)
&=
\cos\!\left(
\frac{i}{10000^{2j/d_{\mathrm{model}}}}
\right),
\label{equa3.9}
\end{align}
for $j = 0,\dots,d_{\mathrm{model}}/2 - 1$. Each row ${\bm{P}} (i,:)$ is therefore a vector representation of the $i$-th position in the sequence.
The matrix $\bm{P}$ modifies the token representation matrix before the first attention layer is applied. As a result, 
the parameters of the attention layers will depend on both token identity and token position. 
In this way, sequence order enters the model through the geometry of the vectors on which attention is built.
The multiscale oscillatory structure of these sinusoidal coordinates across dimensions is illustrated
in Figure~\ref{fig:sinusoidal_position}.

The sinusoidal construction is only one option to encode position. Often, we need the model to capture more than absolute position. It must also track how positions relate to one another: relative displacement, directional order, and the left-to-right structure of the sequence. This motivates positional mechanisms beyond additive sinusoidal embeddings, including relative positional biases and rotary position embeddings.

In relative positional mechanisms, we adjust the score between two token positions using a term that depends on their offset $i-j$ \citep[][]{shaw-etal-2018-self}.
%
In rotary position embeddings, we apply position-dependent rotations to the query and key vectors before forming their dot product \cite{su2024roformer}.
These mechanisms differ in implementation. The goal is the same. They break permutation symmetry and make the attention operator depend on sequence order.

We can view positional structure as an algebraic change to the geometry on which attention acts. Whether position is added to the token vectors, incorporated directly into the score function, or encoded through rotations in projected subspaces, we modify the pairwise relations among token representations so that they reflect sequential order.

\medskip
Putting these ingredients together, we write the Transformer input as
\begin{equation*}
{\bm{X}}_0 = {\bm{U}} {\bm{W}}_E + {\bm{P}} \;\in \mathbb{R}^{n\times d_{\mathrm{model}}} ,
\label{equa3.10}
\end{equation*}
whose $i$-th row is the initial representation of the $i$-th token in the sequence. 
It combines token identity and positional information in a shared $d_{\mathrm{model}}$-dimensional space.
This matrix ${\bm{X}}_0$ is the input state of the Transformer. 
Each layer then maps a matrix of token representations to another matrix in the same ambient space. Thus, we can describe the Transformer as propagating a sequence representation through depth rather than evolving a single hidden state through time. The survey article
\cite{DufterSchmittSchuetze2022} provides a wealth of  information on positional encoding.

\subsection{The Transformer architecture}
\label{sec:transformer_architecture}
We build the Transformer by stacking blocks. Each block acts on a matrix of token representations while preserving the ambient space ${\mathbb{R}}^{n\times d_{\text{model}}}$.
The pattern is repeated. At each depth, we apply the same structural ingredients: self-attention, feed-forward transformations, normalization, and residual addition.
In this section, we develop the block structure, write the Pre-LayerNorm formulation explicitly, and explain how hidden states are mapped to vocabulary scores.

\subsubsection{Transformer models as stacked sequence operators}
Transformers marked a disruptive shift in sequence modeling by replacing recurrence with stacked blocks that combine self-attention, feed-forward transformations, normalization, and residual connections \cite{vaswani2017attention}.
Modern large language models, including the GPT, Claude, Gemini, and Llama families, are built from many such blocks composed in depth \citep[][]{brown2020gpt3, geminiteam2024gemini, dubey2024llama3}. 
For the purposes of this review, the essential point is that a Transformer acts on a matrix of token representations and repeatedly transforms that matrix through depth.
%
%
Mathematically, each transformer block \(\ell\) is a parameterized map
\begin{equation}
{\mathcal{T}}_{\ell}(\,\cdot\,;\Theta_\ell) : \mathbb{R}^{n\times d_{\mathrm{model}}} \to \mathbb{R}^{n\times d_{\mathrm{model}}},
\label{equa4-1}
\end{equation}
which maps a sequence-representation matrix in $\mathbb{R}^{n\times d_{\mathrm{model}}}$ to another matrix in the same space, i.e.,
\begin{equation}
{\bm{X}}_\ell = {\mathcal{T}}_{\ell}({\bm{X}}_{\ell-1};\Theta_\ell),
\qquad
\ell = 1,\dots,n_{\mathrm{blocks}},
\label{equa4-2}
\end{equation}
where
\begin{equation}
{\bm{X}}_\ell \in \mathbb{R}^{n\times d_{\mathrm{model}}} ,
\label{equa4-3}
\end{equation}
where \(d_{\mathrm{model}}\) is the hidden dimension of the model. 

Equations \eqref{equa4-1}--\eqref{equa4-3} are the basic architectural equations of the Transformer.
Unlike a recurrent network, which updates a single hidden state through time, the Transformer evolves an entire matrix of token representations through a stack of layer maps.

At the input layer, \(\ell = 0\), tokenization and positional encoding produce the initial representation matrix ${\bm{X}}_0 \in \mathbb{R}^{n\times d_{\mathrm{model}}}$. 
As explained in Section~\ref{sec:tokens}, the initial hidden-state matrix is
\begin{equation}
{\bm{X}}_0 = {\bm{U}} {\bm{W}}_E + {\bm{P}} ,
\label{equa4.4}
\end{equation}
where ${\bm{U}} \in \mathbb{R}^{n\times |{\mathcal{V}}|}$ is the token matrix, ${\bm{W}}_E \in \mathbb{R}^{|{\mathcal{V}}|\times d_{\mathrm{model}}}$ is the token embedding matrix, and 
${\bm{P}} \in \mathbb{R}^{n\times d_{\mathrm{model}}} $ is the positional representation.
The matrix ${\bm{X}}_0$ is the initial state of the model. Each subsequent block maps ${\bm{X}}_{\ell-1}$ to ${\bm{X}}_{\ell}$ in the same ambient space ${\mathbb{R}}^{ n\times d_{\mathrm{model}} }$, so both sequence length and feature dimension are preserved across depth.
%
%

Applying these block transformations sequentially across depth yields the overall Transformer mapping
\begin{equation}
{\bm{X}}_{n_{\mathrm{blocks}}}
=
{\mathcal{T}} ({\bm{X}}_0)
=
\bigl(
{\mathcal{T}}_{n_{\mathrm{blocks}}}
\circ
{\mathcal{T}}_{n_{\mathrm{blocks}}-1}
\circ \cdots \circ
{\mathcal{T}}_1
\bigr)({\bm{X}}_0).
\label{equa4.5}
\end{equation}
Through this repeated composition, the model progressively transforms the initial token embeddings into contextualized representations that reflect interactions across the sequence.
%

Equation \eqref{equa4.5} shows that depth in a Transformer arises from repeated composition of layer maps on the same ambient space \(\mathbb{R}^{n\times d_{\mathrm{model}}}\). 
This is a fundamental architectural distinction from recurrent models, in which the main state variable evolves through time rather than through depth.

Each transformer block consists of two distinct stages or sublayers, with layer normalization in each stage and residual connections, as in the standard Pre-LayerNorm formulation. 
The first stage operates across the sequence of tokens, e.g., how much a word in a sequence at position $i$ depends on other words at position $i'$. This is the multi-headed self-attention sublayer (MultiHead or ATT for short), 
which will be defined in detail in Section~\ref{sec:multi_head}. The self-attention mechanism is the heart of the Transformer and can effectively capture long-term contextual information.
The second stage operates across the features of each token. This is the multilayer perceptron (MLP), which contains a nonlinear activation function to make the transformer block more expressive.
Details on the MLP sublayer will be provided  in Section~\ref{sec:MLP_layer}.

This division of labor is central to the architecture. The attention sublayer mixes information across token positions, whereas the MLP sublayer mixes information across feature coordinates within each token. The Transformer block is therefore built from an alternation of token mixing and channel mixing.

\subsubsection{Pre-LayerNorm Transformer block}
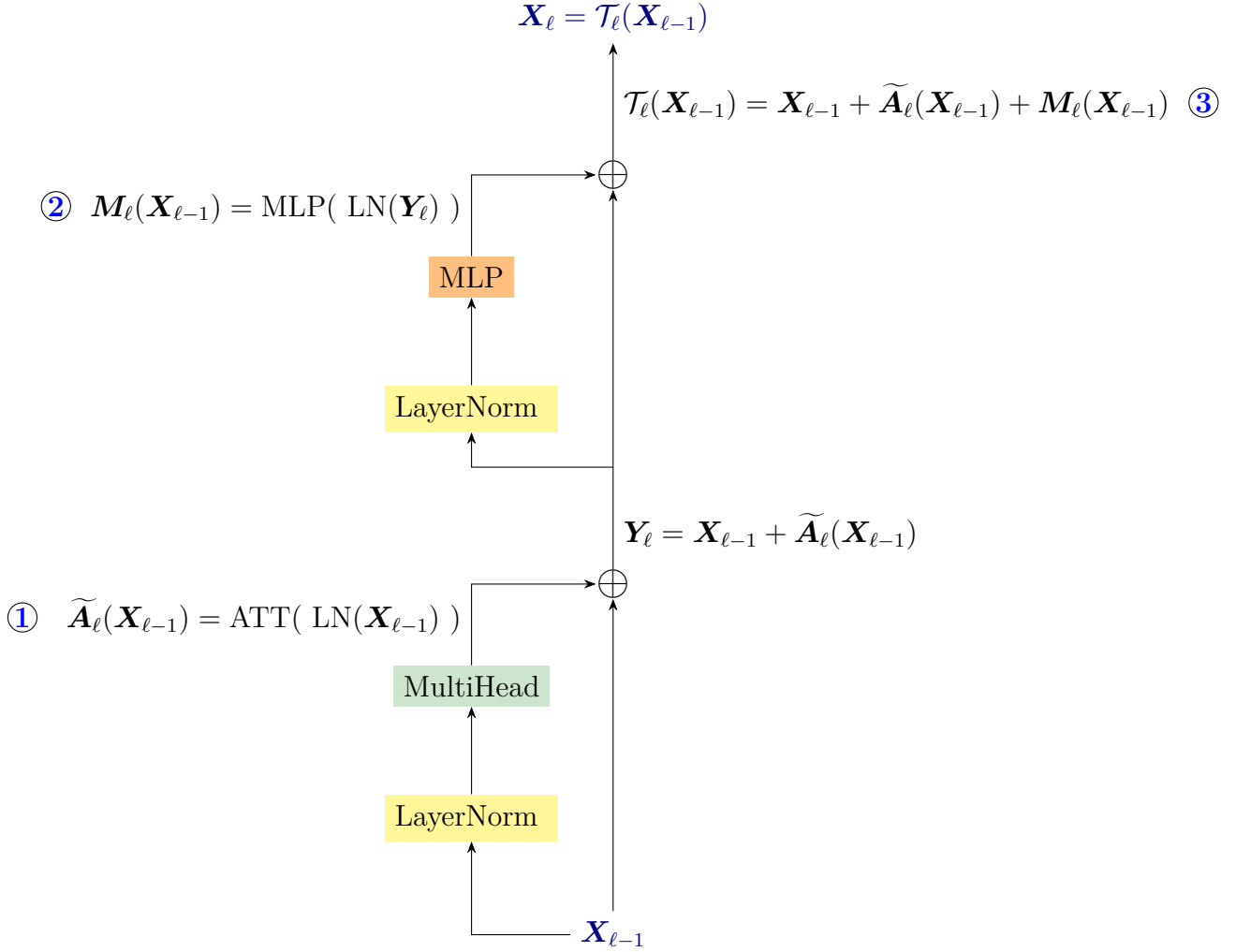
\begin{figure}[ht]
 \begin{tikzpicture}[
    >=Stealth,
    blkg/.style={fill=green!30!gray!30,align=center,minimum width=15mm},
    blky/.style={fill=yellow!50,align=center,minimum width=15mm},
    add/.style={font=\Large, inner sep=0},
 ]
    \node[blue!50!black]       (in1)   at (0, 0) {${\bm{X}}_{\ell-1}$};
    \node[blky] (l2) at (-2,4em) { $\operatorname{LayerNorm}$ };
    \node[blkg, above=3em of l2] (l1)  {$\operatorname{MultiHead}$};
    \node[add]  (add) at (0,12em) {$\oplus$};

    \node[blky] (out) at (-2,18em) { $\operatorname{LayerNorm}$ };
    \node[fill=orange!50, above=3em of out] (l3) { $\operatorname{MLP}$ };
    \node[add]  (add2) at (0,26em) {$\oplus$};
    
    \node[blue!50!black, above=4em of add2]  (out2) {${\bm{X}}_{\ell} = {\mathcal{T}}_{\ell} ({\bm{X}}_{\ell-1} )$};


    \draw[->] (in1) -| (l2) {};
    \draw[->] (l2) -- (l1) {};
    \draw[->] (l1) |- (add) node[left, pos=0.3] 
    {\Circled[inner color=blue]{\textbf{1}} ~ ${\bm{\widetilde A}}_{\ell} ({\bm{X}}_{\ell-1})=\operatorname{ATT} (~\operatorname{LN}({\bm{X}}_{\ell-1})~) $};
    \draw[->] (in1) -- (add) {};
    \draw[->] (out) -- (l3) node[left, pos=0.5] {};
    \draw[->] (l3) |- (add2) node[left, pos=0.3] {\Circled[inner color=blue]{\textbf{2}} ~${\bm{M}}_{\ell} ({\bm{X}}_{\ell-1}) = \operatorname{MLP} (~\operatorname{LN}( {\bm{Y}}_{\ell} )~)$};
    \draw[->] (add) -- (add2) node[right, pos=0.1] 
    { ${\bm{Y}}_{\ell} = {\bm{X}}_{\ell-1} + {\bm{\widetilde A}}_{\ell} ({\bm{X}}_{\ell-1})$ };
    \draw[->] (add2) -- (out2) node[right, pos=0.5] {${\mathcal{T}}_{\ell} ({\bm{X}}_{\ell-1} ) = {\bm{X}}_{\ell-1} +  {\bm{\widetilde A}}_{\ell} ({\bm{X}}_{\ell-1}) +  {\bm{M}}_{\ell} ({\bm{X}}_{\ell-1})$~~\Circled[inner color=blue]{\textbf{3}}};
    \draw[->] (0,16em) -| (out) {};
 \end{tikzpicture}
\caption{$\operatorname{Pre-LayerNorm}$ Transformer block at layer $\ell$. The block applies layer normalization before the attention and MLP sublayers, and writes both outputs additively into the same residual stream. This makes the block an identity-plus-correction map composed of token mixing through attention and channel mixing through the MLP.}
\label{fig1:pre_norm_transformer}
\end{figure}

Mathematically, the layer map ${\mathcal{T}}_{\ell}$ for transformer block $\ell$, shown schematically in Figure~\ref{fig1:pre_norm_transformer} and summarized in Algorithm~\ref{alg:postLayerNorm}, is given by the equations
\footnote{The tilde used in Equation~\eqref{equa4.6} is necessary in order to
  distinguish this operator from the attention operator $\bm{A}_\ell$
  defined in Section~\ref{sec:single_head} which
  is single headed and does  not include normalization.}:
\begin{align}
{\bm{\widetilde A}}_\ell({\bm{X}}_{\ell-1})
&=
\operatorname{ATT} \bigl(
\operatorname{LN}( {\bm{X}}_{\ell-1} )
\bigr),
\label{equa4.6}
\\
{\bm{M}}_\ell({\bm{X}}_{\ell-1})
&=
\operatorname{MLP}\Bigl(
\operatorname{LN}\bigl(
{\bm{X}}_{\ell-1} + {\bm{A}}_\ell({\bm{X}}_{\ell-1})
\bigr)
\Bigr),
\label{equa4.7}
\\
{\bm{X}}_\ell
&=
{\bm{X}}_{\ell-1}
+
{\bm{A}}_\ell({\bm{X}}_{\ell-1})
+
{\bm{M}}_\ell({\bm{X}}_{\ell-1}),
\label{equa4.8}
\end{align}
where the ATT, LN, and MLP are understood to be the operators associated with layer \(\ell\).

Equation \eqref{equa4.6} defines the attention contribution, \eqref{equa4.7}
defines the feed-forward contribution, and \eqref{equa4.8} shows that the full
block writes both contributions additively into the same state matrix. This
additive structure is the beginning of the residual-stream viewpoint that will
become central later in this section.
%
%
\begin{algorithm}[htb]
\centering
\caption{Pre-LayerNorm Transformer}
\label{alg:postLayerNorm}
\begin{algorithmic}[1]
\Input ${\bm{X}} \in {\mathbb{R}}^{n \times d_\texttt{model}}$: initial representation of a given sequence.
\Ensure ${\bm{X}}_{n_\texttt{blocks}} \in {\mathbb{R}}^{n \times d_\texttt{model}}$: the final representation.
\Function{$\operatorname{PreLayerNormTransformer}$}{${\bm{X}}$}
\State ${\bm{X}}_0 \gets {\bm{X}}$
\For{$l=1, \cdots, n_\texttt{blocks}$} 
\Comment{There are $n_\texttt{blocks}$ transformer blocks}
\State ${\bm{\widetilde A}}_{\ell} = \operatorname{ATT} (~\operatorname{LayerNorm} ({\bm{X}}_{\ell-1})~)$ 
\Comment{$\operatorname{MultiHead}$ self-attention stage of block $\ell$}
\State ${\bm{M}}_{\ell} = \operatorname{MLP}(~\operatorname{LayerNorm} ({\bm{X}}_{\ell-1} + {\bm{\widetilde  A}}_{\ell})~)$
\Comment{$\operatorname{MLP}$ stage of transformer block $\ell$}
\State ${\bm{X}}_{\ell} = {\bm{X}}_{\ell-1} + {\bm{\widetilde  A}}_{\ell} + {\bm{M}}_{\ell}$
\Comment{Residual connection}
\EndFor
\State \textbf{return:} ${\bm{X}}_{n_\texttt{blocks}} \gets \operatorname{LayerNorm} ({\bm{X}}_{n_\texttt{blocks}})$

\EndFunction
\end{algorithmic}
\end{algorithm} 

%
\subsubsection{Final normalization and logits}
In the final representation, an additional layer normalization is applied to ${\bm{X}}_{n_{\mathrm{blocks}}}$, i.e.,
\begin{equation}
{\bm{X}}_{n_{\mathrm{blocks}}}
\leftarrow
\operatorname{LN}({\bm{X}}_{n_{\mathrm{blocks}}}),
\label{equa4.9}
\end{equation}
where the resulting hidden states are then projected into vocabulary space through the output un-embedding matrix ${\bm{W}}_U \in \mathbb{R}^{d_{\mathrm{model}}\times |{\mathcal{V}}|}$.
This yields the matrix
%
\begin{equation}
{\bm{Z}} = {\bm{X}}_{n_{\mathrm{blocks}}} {\bm{W}}_U
\;\;\in \mathbb{R}^{n\times |{\mathcal{V}}|} ,
\label{equa4.10}
\end{equation}
whose $i$-th row is the logits vector at token position $i$, that is, the vector of unnormalized scores assigned to all vocabulary items. 
A row-wise softmax then converts each row of ${\bm{Z}}$ into a probability distribution over the vocabulary,
see Equation~\ref{eq:softmax}.

\subsubsection{Encoder-only, decoder-only, and encoder-decoder variants}
\begin{figure}[h]
\centerline{
\includegraphics[width=0.6\textwidth]{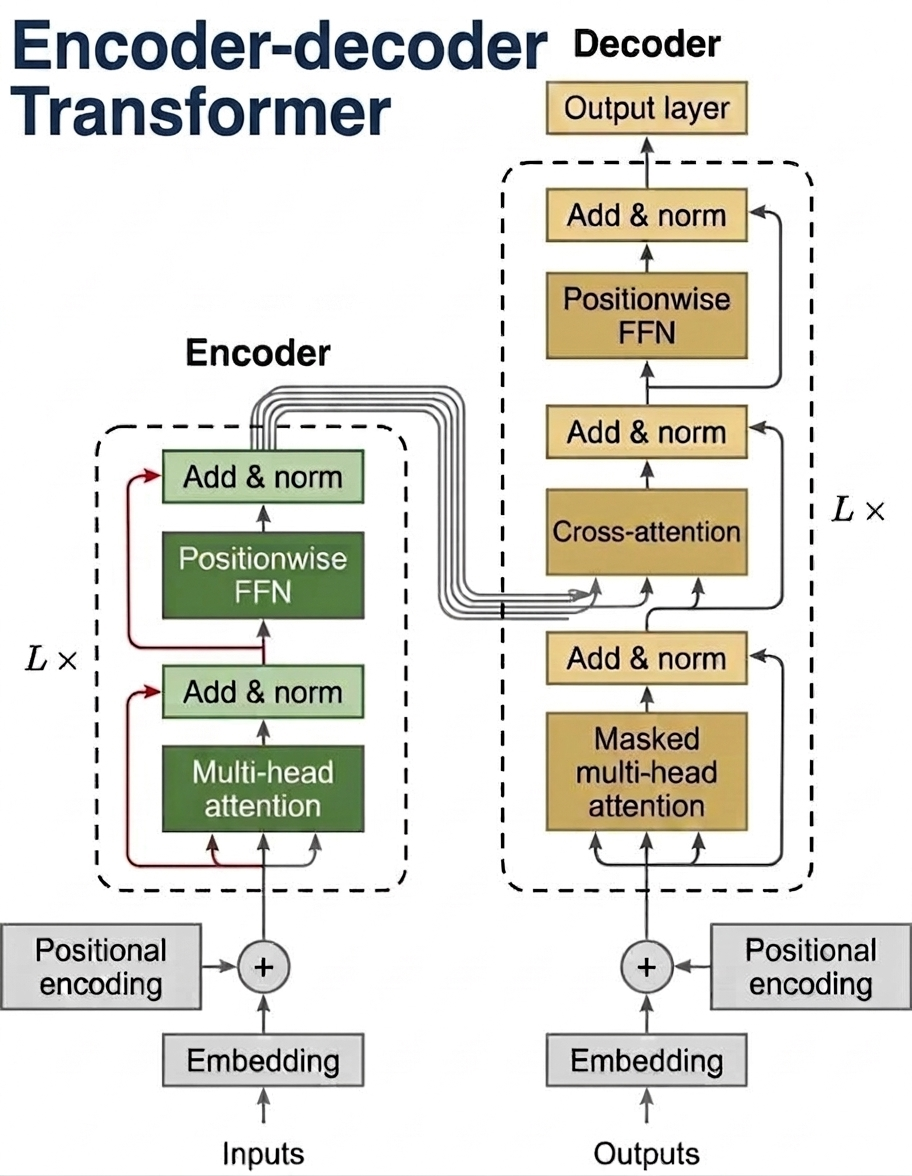}}
\caption{Encoder-decoder Transformer architecture. The encoder computes source-side contextual representations, while the decoder combines masked self-attention and encoder-decoder attention to generate target-side outputs.}
\label{fig:transformer}
\end{figure}

\begin{figure}[h]
\centerline{
\includegraphics[width=0.95\textwidth]{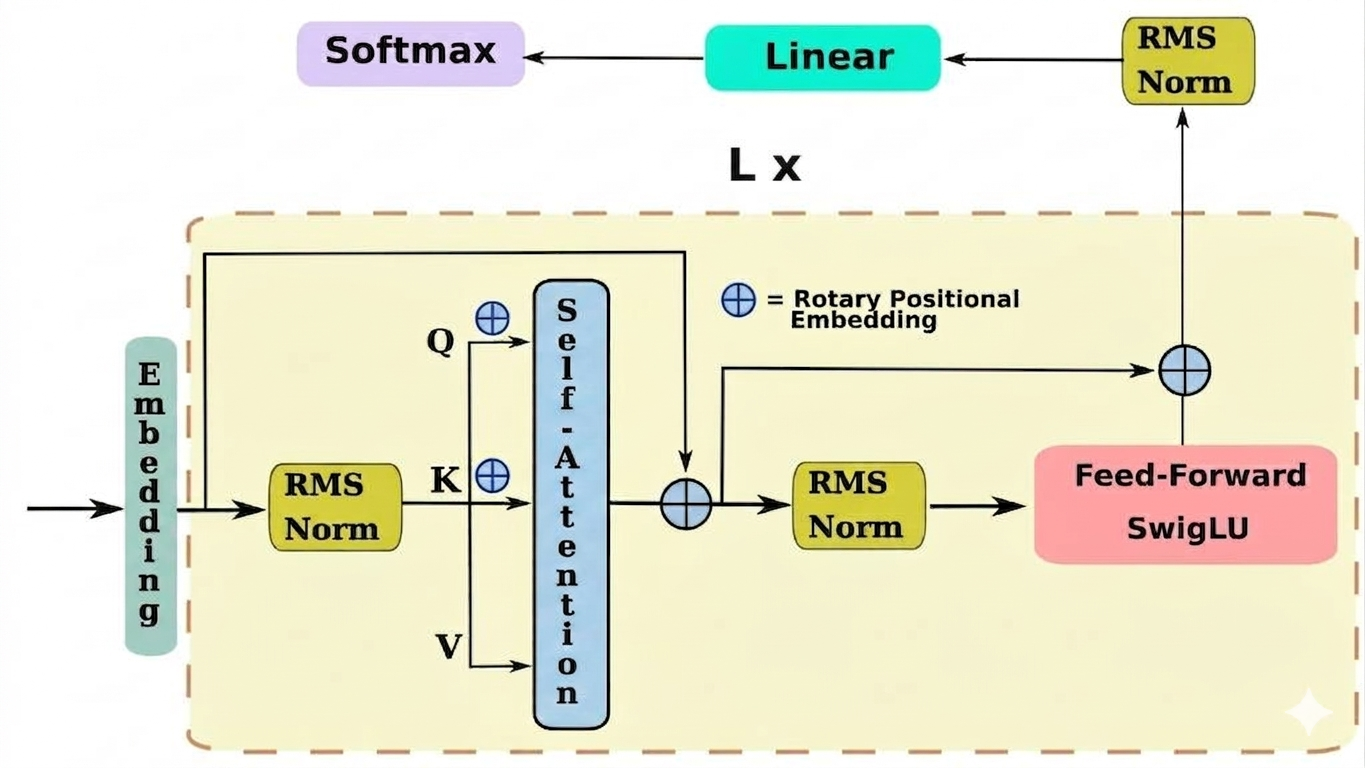}}
\caption{Decoder-only Transformer architecture. Repeated masked self-attention and feed-forward updates act along a shared residual stream, producing contextual representations used for autoregressive prediction.}
\label{fig:llama2}
\end{figure}

The equations above describe the basic Transformer block independently of the architecture family in which it is used.
At a higher level, Transformers appear in three principal forms: encoder-only, decoder-only, and encoder-decoder models. Encoder-only models apply bidirectional self-attention to build contextual representations of an input sequence. Encoder-decoder models combine a source-side encoder stack with a target-side decoder stack. Decoder-only models use masked self-attention so that position $i$ can depend only on positions $1,\dots,i$ \citep[][]{vaswani2017attention, devlin-etal-2019-bert, raffel2020t5}.
For large language models, the decoder-only form is the most important. 
The encoder-decoder and decoder-only organizations are illustrated in Figures~\ref{fig:transformer} and \ref{fig:llama2}, respectively.
Modern LLMs are overwhelmingly implemented as stacks of masked decoder-style Transformer blocks trained for autoregressive next-token prediction. Thus the generic block equations \eqref{equa4.6}--\eqref{equa4.10} remain central, but in the decoder-only case the attention operator must be causally masked. This will be made explicit in Section~\ref{sec:masking}.
%
%
A Transformer is therefore a composition of structured maps acting on the same state space ${\mathbb{R}}^{n\times d_{\text{model}}}$,
\begin{equation}
{\bm{X}}_0
\mapsto
{\bm{X}}_1
\mapsto
\cdots
\mapsto
{\bm{X}}_{n_{\mathrm{blocks}}},
\label{equa4.11}
\end{equation}
with
\begin{equation}
{\bm{X}}_\ell = {\mathcal{T}}_{\ell}({\bm{X}}_{\ell-1}),
\qquad
\ell=1,\dots,n_{\mathrm{blocks}}.
\label{equa4.12}
\end{equation}
The later sections examine the main ingredients of the block ${\mathcal{T}}_{\ell}$: attention, multi-head structure, positional mechanisms, normalization, residual connections, and the feed-forward sublayer.

\subsection{Single-head self-attention}
\label{sec:single_head}
Single-head self-attention is the basic token-mixing operation in the Transformer. Starting from a matrix of token representations, it forms query, key, and value projections, constructs a score matrix from pairwise similarities, and uses the resulting attention weights to combine information across the sequence. This section develops that construction in detail, including the masked form used in autoregressive language modeling.
\subsubsection{Query, key, and value projections}
Among the learned parameters of a Transformer block \(\ell\) are three matrices
\begin{equation}
{\bm{W}}_Q^{(\ell)}, {\bm{W}}_K^{(\ell)} \in \mathbb{R}^{d_{\mathrm{model}}\times d_k},
\qquad
{\bm{W}}_V^{(\ell)} \in \mathbb{R}^{d_{\mathrm{model}}\times d_v}.
\label{equa5.1}
\end{equation}
They map an input vector ${\bm{x}}_i^{\ell-1} \in \mathbb{R}^{d_{\mathrm{model}}}$ to three vectors ${\bm{q}}_i^\ell, {\bm{k}}_i^\ell \in \mathbb{R}^{d_k}$ and ${\bm{v}}_i^\ell \in \mathbb{R}^{d_v}$, called the query, key, and value vectors, respectively:
\begin{equation}
{\bm{q}}_i^\ell = {\bm{x}}_i^{\ell-1} {\bm{W}}_Q^{(\ell)},
\qquad
{\bm{k}}_i^\ell = {\bm{x}}_i^{\ell-1} {\bm{W}}_K^{(\ell)},
\qquad
{\bm{v}}_i^\ell = {\bm{x}}_i^{\ell-1} {\bm{W}}_V^{(\ell)}.
\label{equa5.2}
\end{equation}

Stacking these row-wise over all token positions gives the matrix form
\begin{equation}
{\bm{Q}}_\ell = {\bm{X}}_{\ell-1} {\bm{W}}_Q^{(\ell)},
\qquad
{\bm{K}}_\ell = {\bm{X}}_{\ell-1} {\bm{W}}_K^{(\ell)},
\qquad
{\bm{V}}_\ell = {\bm{X}}_{\ell-1} {\bm{W}}_V^{(\ell)},
\label{equa5.3}
\end{equation}
where ${\bm{Q}}_\ell, {\bm{K}}_\ell \in \mathbb{R}^{n\times d_k}$ and ${\bm{V}}_\ell \in \mathbb{R}^{n\times d_v}$. Usually one takes $d_v=d_k$. 
The multi-head setting will be introduced in Section~\ref{sec:multi_head}.

The matrices ${\bm{Q}}_\ell$, ${\bm{K}}_\ell$, and ${\bm{V}}_\ell$ are obtained by projecting the input data matrix \({\bm{X}}_{\ell-1}\) using different learned projections \({\bm{W}}_Q^{(\ell)}\), \({\bm{W}}_K^{(\ell)}\), and \({\bm{W}}_V^{(\ell)}\). 
This separation is important because it decouples the representation used to query information from the representation used to offer information, while the value projection determines what feature content is transported once the interaction weights have been computed.
%

\subsubsection{Scaled dot-product attention}
The attention mechanism relies on the three matrices, ${\bm{Q}}_\ell$, ${\bm{K}}_\ell$, and ${\bm{V}}_\ell$, which are linear transformations of the input feature matrix ${\bm{X}}_{\ell-1}$. The score matrix is generated from the input sequence itself and measures similarity between tokens at different positions. 
Define 
\begin{equation}
{\bm{S}}_{\ell} = \frac{ {\bm{Q}}_{\ell} {\bm{K}}_{\ell}^\top }{ \sqrt{d_k} } \;\; \in {\mathbb{R}}^{n\times n} .
\label{eq:Sl} 
\end{equation}
Then the attention matrix, which we will often denote by $\bm{P}_\ell$ is
\begin{equation}
\bm{P}_\ell \equiv 
  \operatorname{Attention}({\bm{Q}}_\ell, {\bm{K}}_\ell)
=
\operatorname{softmax} \left( \bm{S}_\ell \right), 
\label{eq:AttDef}
\end{equation}
where $\operatorname{Attention}({\bm{Q}}_\ell, {\bm{K}}_\ell) \in \mathbb{R}^{n\times n}$ relies on the scaled dot-product attention introduced in \cite{vaswani2017attention}. 

Here attention is implemented through scaled dot products between queries and keys. Larger dot products produce larger attention scores, and the factor $1/\sqrt{d_k}$ is introduced to stabilize their magnitude before the row-wise softmax is applied.

Note that the softmax in Equation \eqref{eq:AttDef} is not a matrix function in the spectral sense. Rather, it is shorthand for applying the softmax  function defined earlier in \eqref{eq:softmax} row-wise to ${\bm{Q}}_\ell {\bm{K}}_\ell^\top$. 
If
\begin{equation}
{\bm{s}} = [s_1,s_2,\dots,s_n] \in \mathbb{R}^{1\times n},
\label{equa5.5}
\end{equation}
then
\begin{equation}
\operatorname{softmax}({\bm{s}})
=
\frac{\exp (\bm{s})} {\sum_{j=1}^{n}\exp(s_j)},
\label{equa5.6}
\end{equation}
where the exponential is applied component-wise to the row vector $\bm{s}$.
Thus each row of the attention matrix is a probability distribution over token positions.

\subsubsection{Attention output as weighted averaging}
The output of the self-attention stage in transformer block \(\ell\) is a weighted average of the value vectors, i.e.,
using the notation of Equation~\eqref{eq:AttDef}
\begin{equation}
{\bm{A}}_\ell ( {\bm{X}}_{\ell-1} )
= \bm{P}_\ell \bm{V}_\ell  \equiv 
\operatorname{Attention}( {\bm{Q}}_\ell, {\bm{K}}_\ell ) {\bm{V}}_\ell.
\label{eq:AlDef}
\end{equation}
Equivalently, if $a_{ij}^\ell$ denotes the $(i,j)$-entry of $\operatorname{Attention}({\bm{Q}}_\ell, {\bm{K}}_\ell)$, then the $i$-th output row is
\begin{equation}
\bigl( {\bm{A}}_\ell({\bm{X}}_{\ell-1})\bigr)_{i,:}
=
\sum_{j=1}^{n} a_{ij}^\ell\, {\bm{v}}_j^\ell.
\label{equa5.8}
\end{equation}
This shows explicitly that self-attention replaces each token representation by a weighted average of value vectors from across the sequence.

From a linear algebra point of view, once the attention matrix has been formed, the map
\begin{equation}
{\bm{V}}_\ell \mapsto \operatorname{Attention}({\bm{Q}}_\ell, {\bm{K}}_\ell) {\bm{V}}_\ell
\label{equa5.9}
\end{equation}
is linear in ${\bm{V}}_\ell$.
The nonlinearity comes from  the coefficient matrix  itself which 
  applies the softmax function to inner products of inputs
  defined through the projected queries and keys.

\subsubsection{Masked self-attention}\label{sec:masking}
In a language model, the representation at position $i$ must not depend on future tokens. Therefore the upper-triangular part of the score matrix is masked by setting forbidden entries to $- \infty$. Since $\exp(-\infty) = 0$, these entries do not contribute to the row-wise softmax.
A convenient way to write this is to introduce a mask matrix ${\bm{M}} \in \mathbb{R}^{n\times n}$ carrying the causal structure. 
The masked self-attention matrix used in the decoder, which we still denote by $\bm{P}_\ell$, is then 
\begin{equation}
{\bm{P}}_\ell = 
\operatorname{softmax} \left(\bm{S}_\ell + \bm{M} \right)
\label{eq:MaskedAtt} 
\end{equation}
and as before the transformation is
$\bm{A}_\ell \bm{P}_\ell ( \bm{X}_{\ell-1}) = \bm{P}_\ell \bm{V}_\ell   $.  
Equivalently, one may think of the mask as setting forbidden entries of the score matrix to $-\infty$ before the row-wise softmax is applied. 
The result is a lower-triangular support pattern in the attention matrix, ensuring that token $i$ depends only on tokens $1,\dots,i$.
The attention operator remains a token-mixing matrix, but its support is now restricted by causality, which is what makes decoder-only Transformers suitable for autoregressive language modeling \cite{brown2020gpt3}.

It is also useful to write the attention operator in a form where the row normalization is explicit.
An alternative way of representing attention is
\begin{align}
{\bm{A}}_\ell({\bm{X}}_{\ell-1})
&=
{\bm{D}}^{-1}
\exp \left(
\frac{({\bm{X}}_{\ell-1}{\bm{W}}_Q^{(\ell)})({\bm{X}}_{\ell-1}{\bm{W}}_K^{(\ell)})^\top}{\sqrt{d_k}}
\right)
({\bm{X}}_{\ell-1}{\bm{W}}_V^{(\ell)}),
\label{equa5.17}
\\
\intertext{with}
{\bm{D}}
&=
\operatorname{Diag} \left(
\exp \left(
\frac{({\bm{X}}_{\ell-1}{\bm{W}}_Q^{(\ell)})({\bm{X}}_{\ell-1}{\bm{W}}_K^{(\ell)})^\top}{\sqrt{d_k}}
\right) \mathbf{1}_n
\right) .
\label{equa5.18}
\end{align}
Here the exponential is applied elementwise, and ${\bm{D}}$ is the diagonal matrix of row sums needed to normalize the score matrix into a row-stochastic attention matrix.

This form is useful because it makes the normalization explicit: the score matrix is exponentiated entrywise and then normalized by a diagonal matrix of row sums. It therefore highlights attention as a normalized similarity operator.

Algorithm~\ref{alg:singleHead1} summarizes the matrix-form computation of a single attention head.
This computation is the building block of multi-head attention, where the same operation is carried out in parallel using different learned projections.
%
%
\begin{algorithm}[htb]
\centering
\caption{$\operatorname{SingleHead}$ Self-Attention}
\label{alg:singleHead1}
\begin{algorithmic}[1] 
\Input ${\bm{X}} \in {\mathbb{R}}^{n \times d_\texttt{model}}$ 
\Comment{Representation of a given sequence of $n$ tokens} 
\Require ${\bm{W}}_Q, {\bm{W}}_K, {\bm{W}}_V$
\Comment{Learned weight matrices}
\Ensure ${\bm{A}} \in {\mathbb{R}}^{n \times d_v}$ 
\Comment{Updated representation of tokens in ${\bm{X}}$}
\Function{${\bm{A}} = \operatorname{SingleHead}$}{${\bm{X}} \mid {\bm{W}}_Q, {\bm{W}}_K, {\bm{W}}_V$}
\Comment{Computes a single self-attention head}
\State ${\bm{Q}}, \; {\bm{K}}, \; {\bm{V}} = {\bm{X}} {\bm{W}}_Q, \; {\bm{X}} {\bm{W}}_K, \; {\bm{X}} {\bm{W}}_V $
\Comment{Query, Key, Value matrices}
\State $d_k=\texttt{size}({\bm{Q}},2)$
\State ${\bm{S}} = {\bm{Q}} \, {\bm{K}}^\top / \sqrt{d_k}$
\Comment{Scaled dot-product, ${\bm{S}}$ is a matrix of size $n\times n$}
\State ${\bm{P}} = \texttt{softmax} ( {\bm{S}} )$
\Comment{Attention matrix of probabilities}
\Return ${\bm{A}} = {\bm{P}} \times {\bm{V}}$
\EndFunction
\end{algorithmic}
\end{algorithm} 

\subsubsection{Rank and computational bottleneck}
The self-attention matrix is generated from the product ${\bm{Q}}_\ell {\bm{K}}_\ell^\top$, whose rank satisfies
\begin{equation}
\operatorname{rank} \bigl( {\bm{Q}}_\ell {\bm{K}}_\ell^\top \bigr)
\le
\min \{ \operatorname{rank}( {\bm{Q}}_\ell ), \operatorname{rank}( {\bm{K}}_\ell ) \}
\le d_k.
\label{equa5.19}
\end{equation}
Thus the score matrix before softmax is rank-constrained by the projection dimension $d_k$.
At the same time, it is of size $n\times n$, so computing and storing it incurs $ {\mathcal{O}} (n^2)$ arithmetic and memory costs in the sequence length. 
This quadratic dependency is one of the central computational bottlenecks of standard Transformer attention and motivates many later efficient-attention variants.

\medskip
%

\subsection{Multi-head attention}
\label{sec:multi_head}
Multi-head attention addresses a limitation of single-head attention: one attention map may not be flexible enough when task-relevant information is distributed across different parts of the input. A single head imposes one token-mixing geometry on the whole layer, whereas many problems require several kinds of relations to be represented simultaneously, such as local and long-range dependencies, lexical and semantic associations, or multiple contextual patterns.

%
To address this limitation, the input representation is projected $n_{\mathrm{heads}}$ times into distinct query, key, and value subspaces \cite{vaswani2017attention}.
An independent attention computation is then applied in each projected subspace, and the resulting head outputs are concatenated and projected back into the model space. 
In this way, multi-head attention learns several token-mixing operators in parallel rather than relying on a single one.

\subsubsection{Headwise projections}
For each head \(h = 1,\dots,n_{\mathrm{heads}}\), introduce learned projection matrices
\begin{equation}
{\bm{W}}_{Q,h}^{(\ell)},\;\; {\bm{W}}_{K,h}^{(\ell)} \in \mathbb{R}^{d_{\mathrm{model}}\times d_{\mathrm{head}}},
\qquad
{\bm{W}}_{V,h}^{(\ell)} \in \mathbb{R}^{d_{\mathrm{model}}\times d_{\mathrm{head}}},
\label{equa6.1}
\end{equation}
where one further partitions the projected dimension across heads and writes
\begin{equation}
d_{\mathrm{head}} = \frac{d_k}{n_{\mathrm{heads}}},
\label{equa6.2}
\end{equation}
and in many standard implementations one has \(d_v=d_k=d_{\mathrm{model}}\) at the full-layer level, with each head operating on a lower-dimensional slice.

The transformer’s multi-head scaled dot-product attention is given by
\begin{align}
{\bm{A}}_\ell({\bm{X}}_{\ell-1})
&=
\operatorname{MultiHead} ({\bm{Q}}_\ell, {\bm{K}}_\ell, {\bm{V}}_\ell)
=
\operatorname{Concat} 
\big[ {\bm{H}}_1, \cdots, {\bm{H}}_{n_{\mathrm{heads}}} \big]
{\bm{W}}_O^{(\ell)},
\label{equa6.3}
\\
\intertext{where}
%
{\bm{H}}_h
&=
\operatorname{Attention} \bigl(
{\bm{X}}_{\ell-1} {\bm{W}}_{Q,h}^{(\ell)},
{\bm{X}}_{\ell-1} {\bm{W}}_{K,h}^{(\ell)}
\bigr)
\bigl(
{\bm{X}}_{\ell-1} {\bm{W}}_{V,h}^{(\ell)}
\bigr),
\label{equa6.4}
\\
\intertext{that is,}
%
{\bm{H}}_h
&=
\operatorname{softmax} \left(
\frac{
({\bm{X}}_{\ell-1} {\bm{W}}_{Q,h}^{(\ell)})
({\bm{X}}_{\ell-1} {\bm{W}}_{K,h}^{(\ell)})^\top
}{
\sqrt{d_{\mathrm{head}}}
}
\right)
({\bm{X}}_{\ell-1} {\bm{W}}_{V,h}^{(\ell)}) .
\label{equa6.5}
\end{align}
Here ${\bm{W}}_O^{(\ell)}$ is the output projection matrix that maps the concatenated heads back to the shared model space.
Equations \eqref{equa6.3}--\eqref{equa6.5} show that each head computes its own score matrix, its own attention matrix, and its own value transport. The heads therefore define parallel but distinct operator pathways.

\subsubsection{Concatenation and output projection}
If each head produces an output in \(\mathbb{R}^{n\times d_{\mathrm{head}}}\), then concatenation yields a matrix in \(\mathbb{R}^{n\times (n_{\mathrm{heads}} d_{\mathrm{head}})}\). The output projection
\begin{equation}
{\bm{W}}_O^{(\ell)} \;\;\in \mathbb{R}^{(n_{\mathrm{heads}} d_{\mathrm{head}})\times d_{\mathrm{model}}}
\label{equa6.6}
\end{equation}
then maps this concatenated representation back to the shared residual-stream space $\mathbb{R}^{n\times d_{\mathrm{model}}}$.

Let us decompose the output projection matrix ${\bm{W}}_O^{(\ell)}$ into $n_{\mathrm{heads}}$ block matrices 
%
\begin{equation}
{\bm{W}}_{O,h}^{(\ell)} \in \mathbb{R}^{d_{\mathrm{head}}\times d_{\mathrm{model}}},
\qquad
h=1,2,\dots,n_{\mathrm{heads}}.
\label{equa6.7}
\end{equation}
Therefore,
\begin{align}
{\bm{A}}_\ell({\bm{X}}_{\ell-1})
&=
\operatorname{MultiHead} ({\bm{Q}}_\ell, {\bm{K}}_\ell, {\bm{V}}_\ell)
=
\operatorname{Concat} \bigl[
{\bm{H}}_1, \cdots, {\bm{H}}_{n_{\mathrm{heads}}}
\bigr] {\bm{W}}_O^{(\ell)} ,
\label{equa6.8}
\\
&=
\sum_{h=1}^{n_{\mathrm{heads}}}
{\bm{H}}_h
{\bm{W}}_{O,h}^{(\ell)},
\label{equa6.9}
\\
&=
\sum_{h=1}^{n_{\mathrm{heads}}}
\operatorname{softmax} \left(
\frac{
{\bm{X}}_{\ell-1} {\bm{W}}_{Q,h}^{(\ell)} ({\bm{W}}_{K,h}^{(\ell)})^\top {\bm{X}}_{\ell-1}^\top
}{
\sqrt{d_{\mathrm{head}}}
}
\right)
{\bm{X}}_{\ell-1} {\bm{W}}_{V,h}^{(\ell)} \, {\bm{W}}_{O,h}^{(\ell)}.
\label{equa6.10}
\end{align}
Thus multi-head attention may be interpreted as computing self-attention heads independently, projecting each head’s output into the shared model space, and then summing the results.

\subsubsection{Multi-head attention as a block-structured operator}
Equation \eqref{equa6.10} is especially useful from a linear-algebra point of view. It shows that multi-head attention is not a single attention operator, but a sum of headwise operators, each with its own projected query, key, and value spaces and its own output map.

More concretely, for each head \(h\), define the headwise score matrix
\begin{equation}
{\bm{S}}_h^\ell
=
\frac{
({\bm{X}}_{\ell-1} {\bm{W}}_{Q,h}^{(\ell)})
({\bm{X}}_{\ell-1} {\bm{W}}_{K,h}^{(\ell)})^\top
}{
\sqrt{d_{\mathrm{head}}}
}
\in \mathbb{R}^{n\times n},
\label{equa6.11}
\end{equation}
the headwise attention matrix
\begin{equation}
{\bm{P}}_h^\ell
=
\operatorname{softmax}( {\bm{S}}_h^\ell),
\label{equa6.12}
\end{equation}
and the headwise value matrix
\begin{equation}
{\bm{V}}_h^\ell
=
{\bm{X}}_{\ell-1} {\bm{W}}_{V,h}^{(\ell)}.
\label{equa6.13}
\end{equation}
Then the full multi-head output can be written schematically as
\begin{equation}
{\bm{A}}_{\ell,\text{multihead}}   ({\bm{X}}_{\ell-1})
=
\sum_{h=1}^{n_{\mathrm{heads}}}
{\bm{P}}_h^\ell\, {\bm{V}}_h^\ell\, {\bm{W}}_{O,h}^{(\ell)}.
\label{equa6.14}
\end{equation}
This expression makes the block structure explicit. Each head contributes its own token-mixing operator ${\bm{P}}_h^\ell$. It acts on that head's projected value space, and we then map the result back into the common residual stream.

\subsubsection{Subspace structure, expressivity, and specialization}
Because the projections are separate, we let each head operate in its own learned subspace of the model representation space. Multi-head attention is then not just a larger attention module. We can view it as splitting the token-interaction problem across several lower-dimensional projected spaces.

This linear-algebra view is useful. It says more than an architectural description alone. Instead of forcing one attention matrix to capture all relevant token relations, we let the model learn several interaction geometries at the same time. Different heads may emphasize different patterns of similarity, and the output projection recombines their contributions into a single update of the residual stream. Why does this help? We do not yet have a complete answer, but the architecture points to two plausible effects.

First, multi-head attention can increase expressivity because it lets the model represent different token-token relations at once. Second, it can add redundancy. That matters. Redundancy may make optimization easier and may allow several heads to share or overlap in function. Empirically, we see both behaviors: some heads show recognizable specialization, whereas others appear partly redundant.
%
Mathematically, we should not treat multi-head attention as a single token-mixing operator. It gives a block-structured family of such operators. This matters later. It makes questions about rank, subspace overlap, redundancy, and interpretability arise naturally.

\medskip
The multi-head computation may be summarized in three stages: headwise query, key, and value projections; parallel computation of the head outputs; and concatenation followed by projection back into the shared model space. 
More explicitly, for each head $h$, one computes
    \begin{equation}
    {\bm{Q}}_h = {\bm{X}} {\bm{W}}_{Q,h},\qquad {\bm{K}}_h = {\bm{X}} {\bm{W}}_{K,h},\qquad {\bm{V}}_h = {\bm{X}} {\bm{W}}_{V,h},
    \label{equa6.15}
    \end{equation}
forms
    \begin{equation}
    {\bm{H}}_h
    =
    \operatorname{softmax} \left(
    \frac{ {\bm{Q}}_h {\bm{K}}_h^\top}{\sqrt{d_{\mathrm{head}}}}
    \right) {\bm{V}}_h,
    \label{equa6.16}
    \end{equation}
and then combines the head outputs through concatenation and output projection, i.e.,
    \begin{equation}
    {\bm{A}}
    =
    \operatorname{Concat} \big[
    {\bm{H}}_1, \cdots, {\bm{H}}_{n_{\mathrm{heads}}}
    \big] {\bm{W}}_O.
    \label{equa6.17}
    \end{equation}
Thus multi-head attention is a parallel composition of several single-head attention computations followed by a linear recombination.



\subsection{Positional structure and sequence order}
Positional structure is what turns self-attention from a permutation-equivariant operation into a sequence model. Because attention is built from projected similarities among token representations, positional information enters the model by modifying the geometry on which those similarities are computed. 
The section first explains why order is invisible to self-attention alone, then examines the main positional mechanisms used in Transformers, and finally discusses their role in decoder-only language models.
\subsubsection{Why self-attention alone ignores order}
\label{sec:self_attention}
A naive self-attention model sees only an unordered set of tokens. In language, however, sequence order carries information, so positional structure must be injected into the model for token order to influence the computation.

The self-attention mechanism introduced in Sections~\ref{sec:single_head} and \ref{sec:multi_head} acts on the rows of the representation matrix \({\bm{X}}_{\ell-1}\) through learned projections and pairwise dot products. If the rows of \({\bm{X}}_{\ell-1}\) are permuted, then the corresponding queries, keys, and values are permuted in the same way, and the resulting attention output is permuted accordingly. Thus, by itself, self-attention is permutation-equivariant rather than sequence-aware.

In other words, token identity alone does not determine sequence structure. The model must be told, in some algebraic form, where each token lies in the sequence or how token positions relate to one another.

\subsubsection{Absolute and relative positional mechanisms}
The original Transformer introduces positional information through additive absolute positional encodings. Let \({\bm{P}}(i,:)\) denote the positional encoding of token \(i\). Then the initial representation is
\begin{equation}
{\bm{X}}_0 = {\bm{U}} {\bm{W}}_E + {\bm{P}} ,
\label{equa7.1}
\end{equation}
where ${\bm{U}} {\bm{W}}_E$ is the token embedding matrix and ${\bm{P}} \in \mathbb{R}^{n\times d_{\mathrm{model}}}$ is the matrix of positional encodings.

The idea is to represent position in such a way that tokens with similar relative positions have related positional encodings. In the original Transformer, this is done through frequency-based sinusoidal representations:
\begin{equation}
{\bm{P}}(i,2j)
=
\sin\!\left(
\frac{i}{10000^{2j/d_{\mathrm{model}}}}
\right),
\label{equa7.2}
\end{equation}
\begin{equation}
{\bm{P}}(i,2j+1)
=
\cos\!\left(
\frac{i}{10000^{2j/d_{\mathrm{model}}}}
\right),
\label{equa7.3}
\end{equation}
for \(j = 0,\dots,d_{\mathrm{model}}/2-1\). 
Each row ${\bm{P}}(i,\colon)$ is therefore a $d_{\mathrm{model}}$-dimensional representation of the $i$-th token position.

From a linear algebra point of view, absolute positional encoding perturbs the token embedding matrix before any attention operator acts. Thus position enters the model by modifying the vectors whose projected similarities determine the score matrix. The attention mechanism itself is unchanged in form, but the geometry of the underlying token representations is no longer permutation-invariant.

The addition in \eqref{equa7.1} is structurally important. It means that positional information is introduced not by a separate symbolic channel, but by modifying the vectors on which all subsequent linear projections operate. If
\begin{equation}
{\bm{Q}}_\ell = {\bm{X}}_{\ell-1} {\bm{W}}_Q^{(\ell)},
\qquad
{\bm{K}}_\ell = {\bm{X}}_{\ell-1} {\bm{W}}_K^{(\ell)},
\label{equa7.4}
\end{equation}
then already in the first layer,
\begin{equation}
{\bm{Q}}_1 = ({\bm{U}} {\bm{W}}_E + {\bm{P}}) {\bm{W}}_Q^{(1)},
\qquad
{\bm{K}}_1 = ({\bm{U}} {\bm{W}}_E + {\bm{P}}) {\bm{W}}_K^{(1)}.
\label{equa7.5}
\end{equation}
Thus both token identity and token position are mixed into the projected query and key spaces before the score matrix is formed.

This is why positional encoding is best viewed as a perturbation of token geometry. The self-attention operator still depends on projected dot products, but those dot products now reflect not only lexical content but also token position within the sequence.

The sinusoidal scheme is historically important, but it is not the only way to encode sequence order. Modern Transformer architectures often use other positional mechanisms, especially when long contexts or stronger inductive biases are desired.

In relative positional mechanisms, the score between two token positions is adjusted by a term that depends on their offset $i-j$. 
Algebraically, this means that the score matrix is no longer determined solely by a bilinear form in the token representations, but also by offset-dependent corrections.

Rotary position embedding (RoPE) gives another influential family. In RoPE, we rotate the queries and keys in a position-dependent way before taking their dot product. The basic matrix form of attention remains in place, but the rotations make the attention mechanism produce similarities that are sensitive to relative displacement.

The implementations differ. The purpose does not. Each scheme breaks permutation symmetry and makes the token-mixing operator aware of sequence order.

At a conceptual level, position may enter the Transformer in three broad ways: by adding an absolute positional vector to each token representation, as in Equations \eqref{equa7.1}--\eqref{equa7.3}; by modifying the pairwise score function directly through relative offsets; or by transforming the projected query and key vectors through position-dependent linear operators, as in rotary schemes.
They act at different stages of the attention pipeline. Still, they all change the geometry of the score matrix
${\bm{S}}_\ell$ defined in Equation~\eqref{eq:Sl}
Absolute positional encoding changes the input matrix before projection. Relative-position methods modify the score function itself. Rotary schemes modify the projected coordinates in which we measure similarity.

%
Position changes the matrices on which attention depends, and hence changes the stochastic token-mixing operators produced by self-attention.


\subsubsection{Sequence order in decoder-only LLMs}
\label{sec:decoder1}
In modern decoder-only language models, positional structure works together with causal masking. The causal mask ensures that token \(i\) cannot attend to positions \(j>i\), while the positional mechanism determines how the allowed positions \(j\le i\) are represented geometrically relative to token i. Sequence order therefore enters through two distinct mechanisms: causality, which restricts the support of attention, and positional structure, which shapes the similarities among the positions that remain visible. Causal masking alone tells the model which positions are forbidden, but not how earlier visible positions are arranged relative to one another. Positional structure provides that second ingredient.

\medskip
Thus decoder-only language models require both causal masking and a positional mechanism. The first restricts which token positions may interact, while the second determines how the visible positions are represented geometrically relative to one another. Whether position is introduced through absolute encodings, relative score modifications, or rotary transformations of the projected coordinates, the effect is to make the token-mixing operator sequence-aware.

\subsection{The MLP / feed-forward sublayer}\label{sec:MLP_layer}
Within each Transformer block, the MLP provides the nonlinear per-token transformation that complements the token-mixing role of attention.
While self-attention allows each token to incorporate information from other positions in the sequence, the MLP acts separately on each token representation and enriches it through learned feature transformations. In this way, attention and the MLP play distinct but complementary roles within the block, and it is therefore natural to examine the feed-forward sublayer in more detail.
\subsubsection{The feed-forward stage: expansion, compression, and channel mixing}
The MLP is the second main operator in the Transformer block. 
It is a feed-forward map built from affine transformations and nonlinear activations.
In its simplest form, it consists of an up-projection parameterized by $({\bm{W}}_{\mathrm{up}}^\ell,{\bm{b}}_{\mathrm{up}}^\ell)$, a nonlinearity, and a down-projection parameterized by $({\bm{W}}_{\mathrm{down}}^\ell,{\bm{b}}_{\mathrm{down}}^\ell)$. 
The output of the multi-head self-attention sublayer in transformer block $\ell$,
\begin{equation}
{\bm{Y}}_\ell = {\bm{X}}_{\ell-1} + \operatorname{ATT} \bigl( \operatorname{LN} ( {\bm{X}}_{\ell-1} ) \bigr),
\label{equa8.1}
\end{equation}
is then passed to a multilayer perceptron (MLP), typically consisting of two affine layers separated by a nonlinearity.

The MLP first expands the representation from dimension $d_{\mathrm{model}}$ to a higher dimension $d_{\mathrm{mlp}}$, applies a nonlinearity, and then projects the result back to $d_{\mathrm{model}}$.
Thus,
\begin{equation}
{\bm{M}}_\ell({\bm{X}}_{\ell-1})
=
\operatorname{MLP} ( {\bm{Y}}_\ell )
= \bigl(\operatorname{ReLU} ( { \operatorname{LN}}({\bm{Y}}_\ell) {\bm{W}}_{\mathrm{up}}^\ell + {\bm{b}}_{\mathrm{up}}^\ell) \bigr) {\bm{W}}_{\mathrm{down}}^\ell + {\bm{b}}_{\mathrm{down}}^\ell.
\label{equa8.2}
\end{equation}
Here the learned parameter matrices ${\bm{W}}_{\mathrm{up}}^\ell$ and ${\bm{W}}_{\mathrm{down}}^\ell$ are of size
\begin{equation}
{\bm{W}}_{\mathrm{up}}^\ell \in \mathbb{R}^{d_{\mathrm{model}}\times d_{\mathrm{mlp}}},
\qquad
{\bm{W}}_{\mathrm{down}}^\ell \in \mathbb{R}^{d_{\mathrm{mlp}}\times d_{\mathrm{model}}} .
\label{equa8.3}
\end{equation}
%

In contrast with attention, which mixes information across token positions through an $n\times n$ attention matrix, the MLP acts independently on each row of the representation matrix. Thus the MLP is a channel-mixing operator rather than a token-mixing operator.

This expansion-compression structure is a key architectural ingredient of the Transformer block. In many implementations, one takes $d_{\mathrm{mlp}} \approx 4d_{\mathrm{model}}$.
A single linear transformation would remain affine and could not provide the input-dependent nonlinear feature transformation performed by the MLP.

At the level of a single token representation, using the row-vector convention adopted in this section, one may write
\begin{equation}
\operatorname{MLP}(\bm{x})
=
\sigma \bigl(
{\bm{x}} {\bm{W}}_{\mathrm{up}}^{\ell} + {\bm{b}}_{\mathrm{up}}^\ell
\bigr) {\bm{W}}_{\mathrm{down}}^{\ell}
+
{\bm{b}}_{\mathrm{down}}^\ell,
\label{equa8.4}
\end{equation}
where \(\sigma(\cdot)\) denotes a nonlinearity such as ReLU, GELU, or a gated activation.

The intermediate expansion to \(d_{\mathrm{mlp}}\) introduces a large number of latent coordinates. After the nonlinearity, the subsequent compression to dimension \(d_{\mathrm{model}}\) recombines these activated channels into a compact representation.
From a linear-algebra perspective, the MLP therefore performs a lift into a higher-dimensional feature space followed by projection back into the residual stream, i.e., 
\begin{equation}
\mathbb{R}^{d_{\mathrm{model}}}
\;\xrightarrow{\;{\bm{W}}_{\mathrm{up}}^\ell\;}\;
\mathbb{R}^{d_{\mathrm{mlp}}}
\;\xrightarrow{\;\sigma\;}\;
\mathbb{R}^{d_{\mathrm{mlp}}}
\;\xrightarrow{\;{\bm{W}}_{\mathrm{down}}^\ell\;}\;
\mathbb{R}^{d_{\mathrm{model}}} .
\label{equa8.5}
\end{equation}
%


\subsubsection{Nonlinearity and input-dependent Jacobians}

The Jacobian
\begin{equation}
J(\bm{x})
=
{\bm{W}}_{\mathrm{up}}^{\ell}
\operatorname{Diag} \Bigl(
\sigma' \bigl(
{\bm{x}} {\bm{W}}_{\mathrm{up}}^{\ell} + {\bm{b}}_{\mathrm{up}}^\ell
\bigr)
\Bigr)
{\bm{W}}_{\mathrm{down}}^{\ell}
\label{equa8.6}
\end{equation}
depends on the input ${\bm{x}} \in \mathbb{R}^{d_{\mathrm{model}}}$, and therefore defines an input-dependent effective linear map. 
This is one way to view the MLP as a mechanism for dynamic feature selection. 
A single affine transformation would have a constant Jacobian and could not produce this behavior.

The intermediate activation
\begin{equation}
{\bm{h}} = {\bm{x}} {\bm{W}}_{\mathrm{up}}^{\ell} + {\bm{b}}_{\mathrm{up}}^\ell
\label{equa8.7}
\end{equation}
lies in ${\mathbb{R}}^{ d_{\mathrm{mlp}} }$ and is then passed through the nonlinearity $\sigma( \cdot )$.
For ReLU or GELU activations, each coordinate of $\bm{h}$ is either suppressed or attenuated according to its sign or magnitude. 
This selective effect may be interpreted as a learned gating mechanism.

For the ReLU activation, \(\sigma(z)=\max(0,z)\), define the diagonal matrix
\begin{equation}
{\bm{D}} ({\bm{x}})
=
\operatorname{Diag} \Bigl(
\mathbf{1}_{ \{\, {( {\bm{x}} {\bm{W}}_{\mathrm{up}}^{\ell} +{\bm{b}}_{\mathrm{up}}^\ell)}_k \,>\, 0 \, \} }
\Bigr)
\in \mathbb{R}^{d_{\mathrm{mlp}}\times d_{\mathrm{mlp}}} ,
\label{equa8.8}
\end{equation}
Then
\begin{equation}
\operatorname{MLP}({\bm{x}})
=
\bigl(
{\bm{x}} {\bm{W}}_{\mathrm{up}}^{\ell} + {\bm{b}}_{\mathrm{up}}^\ell
\bigr)
{\bm{D}} ({\bm{x}}) {\bm{W}}_{\mathrm{down}}^{\ell}
+
{\bm{b}}_{\mathrm{down}}^\ell.
\label{equa8.9}
\end{equation}
Because the binary mask \(D({\bm{x}})\) depends on the input, the map \( {\bm{x}} \mapsto \operatorname{MLP}({\bm{x}})\) is piecewise affine, i.e.,
\begin{equation}
\operatorname{MLP}({\bm{x}})
=
{\bm{x}}
\underbrace{
{\bm{W}}_{\mathrm{up}}^{\ell} {\bm{D}} ({\bm{x}}){\bm{W}}_{\mathrm{down}}^{\ell}
}_{ {\bm{A}}_{ {\bm{D}} ({\bm{x}})} }
\;+\;
\underbrace{
{\bm{b}}_{\mathrm{up}}^\ell
{\bm{D}} ({\bm{x}}) {\bm{W}}_{\mathrm{down}}^{\ell} + {\bm{b}}_{\mathrm{down}}^\ell
}_{ {\bm{c}}_{ {\bm{D}} ({\bm{x}}) } }.
\label{equa8.10}
\end{equation}
Each distinct activation pattern corresponds to a different affine region in \(\mathbb{R}^{d_{\mathrm{model}}}\).

Thus the MLP may be viewed as an input-dependent family of affine maps selected by the activation pattern of the expanded hidden units. This is one of the main sources of nonlinear feature transformation in the Transformer block.

\subsubsection{Gated variants: SwiGLU and related designs}
In modern Llama-style architectures, the MLP stage uses three weight matrices rather than the two used in the original Transformer. Hence
\begin{equation}
\operatorname{MLP}({\bm{Y}}_\ell)
=
\operatorname{SwiGLU} \Bigl(
({\bm{Y}}_\ell {\bm{W}}_{\mathrm{up}}^\ell + {\bm{b}}_{\mathrm{up}}^\ell)
\odot
({\bm{Y}}_\ell {\bm{W}}_{\mathrm{gate}}^\ell + {\bm{b}}_{\mathrm{gate}}^\ell)
\Bigr)
{\bm{W}}_{\mathrm{down}}^\ell
+
{\bm{b}}_{\mathrm{down}}^\ell,
\label{equa8.11}
\end{equation}
where the activation function SwiGLU is defined through a gated swish-type nonlinearity \cite{shazeer2020glu_variants}. The main structural point is that the MLP now includes an elementwise product of two projected branches, introducing multiplicative gating into the feature transformation.
This makes the channel-mixing role of the MLP even more explicit. Instead of a single expanded hidden representation being activated and projected back, the model computes two expanded branches and combines them coordinatewise before the down projection.

\subsubsection{Why the MLP dominates parameter count and compute}
Because the MLP contains two large dense matrices of sizes $(d_{\mathrm{model}}\times d_{\mathrm{mlp}})$ and $(d_{\mathrm{mlp}}\times d_{\mathrm{model}})$, it contributes approximately $2d_{\mathrm{model}}d_{\mathrm{mlp}}$
parameters per dense Transformer block, up to lower-order bias terms.
In GPT-3 175B, the parameters in the MLP stages account for roughly two-thirds of the total model parameters \cite{brown2020gpt3}. In Llama-3 405B, the fraction is even larger because of the gated three-matrix design \cite{dubey2024llama3}. Thus, in large dense Transformer models, the MLP layers account for the majority of parameters and a large share of the computational cost.
They also account for a substantial share of the computation during training.
This observation is important for the later discussion of Mixture-of-Experts models. Sparse expert designs typically replace the dense MLP stage rather than the attention stage, precisely because the MLP is where most of the parameters and much of the computation reside.

\medskip
Beyond parameter count, the MLP also plays a distinct architectural role. Its expand-compress structure provides the Transformer with a mechanism for per-token nonlinear modeling that complements the cross-token interactions of attention. Attention mixes information across token positions through an $n\times n$ stochastic operator, whereas the MLP acts independently on each token representation through a nonlinear map on ${\mathbb{R}}^{ d_{\mathrm{model}} }$.
The alternation between token mixing and channel mixing is one of the key structural features of the Transformer block.


\subsection{Layer normalization and residual connections}

Deep Transformer stacks are difficult to optimize without architectural mechanisms that stabilize signal propagation through depth. Two such mechanisms are residual connections and layer normalization. In the Transformer, the residual branch carries an identity pathway through the network, while the nonlinear update is supplied by self-attention in one sublayer and by the feed-forward map in the other. Layer normalization controls the scale and centering of the hidden representations flowing through this structure, thereby improving optimization stability\cite{ba2016layernorm_article}.

In the present section, these two ingredients should be understood together. Residual connections provide an identity pathway through depth, while normalization regulates the scale and centering of the vectors flowing through that pathway. Without these mechanisms, repeated compositions of attention and MLP blocks would be much more difficult to train stably.

Layer normalization acts across the feature dimension of each token representation. In the original Transformer, it is applied around each sublayer in both the encoder and decoder stacks. At the level of a single token vector, it subtracts the mean, rescales by the standard deviation, and then applies learned affine parameters. For the purposes of this review, the essential point is that LayerNorm acts row-wise on the sequence representation matrix and helps stabilize repeated composition of deep operator blocks.

Let ${\bm{x}} \in \mathbb{R}^{d_{\mathrm{model}}}$
denote an intermediate hidden representation, for example a single row of the sequence representation matrix. The corresponding normalized, scaled, and shifted vector ${\bm{y}}$ is
\begin{equation}
{\bm{y}}
=
\operatorname{LayerNorm}( \bm{x} )
=
\frac{ {\bm{x}}- \mu( \bm{x} ) }{ \sqrt{ \sigma^2 ( \bm{x} )  + \varepsilon } }
\odot \boldsymbol{ \gamma }
+
\boldsymbol{ \beta } ,
\label{equa7.2a}
\end{equation}
where $\boldsymbol{ \gamma }, \boldsymbol{ \beta } \in \mathbb{R}^{d_{\mathrm{model}}}$ are scaling and shifting vector parameters learned during training, and $\varepsilon$ is used for numerical stability. The mean $\mu$ and variance $\sigma^2$ of the elements in $\bm{x}$ are defined as
\begin{align}
\mu(\bm{x})
&=
\frac{1}{d_{\mathrm{model}}}\sum_{i=1}^{d_{\mathrm{model}}} x_i,   \\
\sigma^2(\bm{x})
&=
\frac{1}{d_{\mathrm{model}}}
\sum_{i=1}^{d_{\mathrm{model}}}
\bigl(x_i-\mu( \bm{x} )\bigr)^2.
\label{equa7.4_1}
\end{align}

Equations (\ref{equa7.2a}--\ref{equa7.4_1}) are enough for the main architectural discussion. They show that LayerNorm acts row-wise on the representation matrix, centering and rescaling each token representation independently across its feature coordinates.

Written in matrix form, LayerNorm is applied row-wise to the sequence representation matrix ${\bm{X}} \in \mathbb{R}^{n\times d_{\mathrm{model}}}$.
At a schematic level, one may think of it as the row-wise transformation
\begin{equation}
{\bm{X}}
\mapsto
\operatorname{LN}({\bm{X}}),
\end{equation}
where the \(i\)-th row of \(\operatorname{LN}({\bm{X}})\) is obtained by applying \eqref{equa7.2} to the \(i\)-th row of \({\bm{X}}\). 
Thus LayerNorm does not mix token positions. Like the MLP, it acts independently on each row, but unlike the MLP it performs normalization rather than learned feature expansion and compression.

From a linear-algebra viewpoint, LayerNorm is not globally linear because its centering and scaling depend on the current input vector. Still, it is built from familiar operations: subtraction of the mean component, rescaling by a row-dependent norm, and coordinatewise affine transformation through $\boldsymbol{ \gamma }$ and $\boldsymbol{ \beta }$.

\subsubsection{RMSNorm as a modern variant}
$\operatorname{RMSNorm}$
is a popular variant of layer normalization used to train Llama series of models. Rather than centering and normalizing by the standard deviation,
RMSNorm rescales by the root-mean-square magnitude of the input\cite{zhang2019rmsnorm}.
Specifically, Equations (\ref{equa7.2a}--\ref{equa7.4_1}) are replaced by
\begin{equation*}
\operatorname{RMSNorm}( \bm{x} )
=
\frac{ \bm{x} }{ \operatorname{RMS} (\bm{x}) }
\odot \boldsymbol{ \gamma }
+
\boldsymbol{ \beta } , \quad \text{where}\quad
\operatorname{RMS} (\bm{x}) = \left[
  \frac{1}{d_{\mathrm{model}}}
\sum_{i=1}^{d_{\mathrm{model}}}
x_i^2 \right]^{1/2} . 
\end{equation*}  

At a high level, RMSNorm removes the mean-subtraction part of LayerNorm and keeps only scale normalization together with learned gain parameters. This simplification has proved effective in modern LLMs, especially in the Llama family.
For the purposes of the present section, the key point is that RMSNorm plays the same architectural role as LayerNorm: it stabilizes the repeated composition of deep operator blocks while preserving the row-wise structure of the sequence representation.
%

Normalization and residual connections should therefore be understood together: normalization controls the scale of the representations being propagated, while residual pathways preserve an identity channel through depth.
The distinction between $\operatorname{LayerNorm}, \operatorname{RMSNorm}, \operatorname{Pre-LayerNorm}$, and $\operatorname{Post-LayerNorm}$ becomes most important when one examines how normalization interacts with residual addition across depth. That interaction is the focus of the next section.

\subsection{Pre-LayerNorm, Post-LayerNorm, and the residual stream}
\label{sec:residual_1}
In early versions of the Transformer, layer normalization is placed after the element-wise residual addition. This yields the Post-LayerNorm formulation. In more recent implementations, LayerNorm is often applied before each sublayer, leading to the Pre-LayerNorm formulation~\cite{xiong2020layer}.

For the attention stage,
\begin{align}
{\bm{Y}}_\ell
&=
\operatorname{LayerNorm} \bigl(
{\bm{X}}_{\ell-1} + \operatorname{ATT}({\bm{X}}_{\ell-1})
\bigr)
\qquad
\text{(Post-LayerNorm)},
\label{equa7.7}
\intertext{whereas}
{\bm{Y}}_\ell
&=
{\bm{X}}_{\ell-1} + \operatorname{ATT} \bigl(
\operatorname{LayerNorm}({\bm{X}}_{\ell-1})
\bigr)
\qquad
\text{(Pre-LayerNorm)}.
\label{equa7.8}
\end{align}

For the MLP stage,
\begin{align}
{\bm{X}}_\ell
&=
\operatorname{LayerNorm} \bigl(
{\bm{Y}}_\ell + \operatorname{MLP}({\bm{Y}}_\ell)
\bigr)
\qquad
\text{(Post-LayerNorm)},
\label{equa7.9}
\intertext{whereas}
{\bm{X}}_\ell
&=
{\bm{Y}}_\ell + \operatorname{MLP} \bigl(
\operatorname{LayerNorm}({\bm{Y}}_\ell)
\bigr)
\qquad
\text{(Pre-LayerNorm)}.
\label{equa7.10}
\end{align}

This distinction is architecturally important. In the Pre-LayerNorm form, the residual branch remains an exact identity path, while normalization is applied only to the input of the sublayer. In the Post-LayerNorm form, normalization acts after the residual addition. Modern large language models typically prefer the Pre-LayerNorm arrangement because it improves optimization stability at depth.


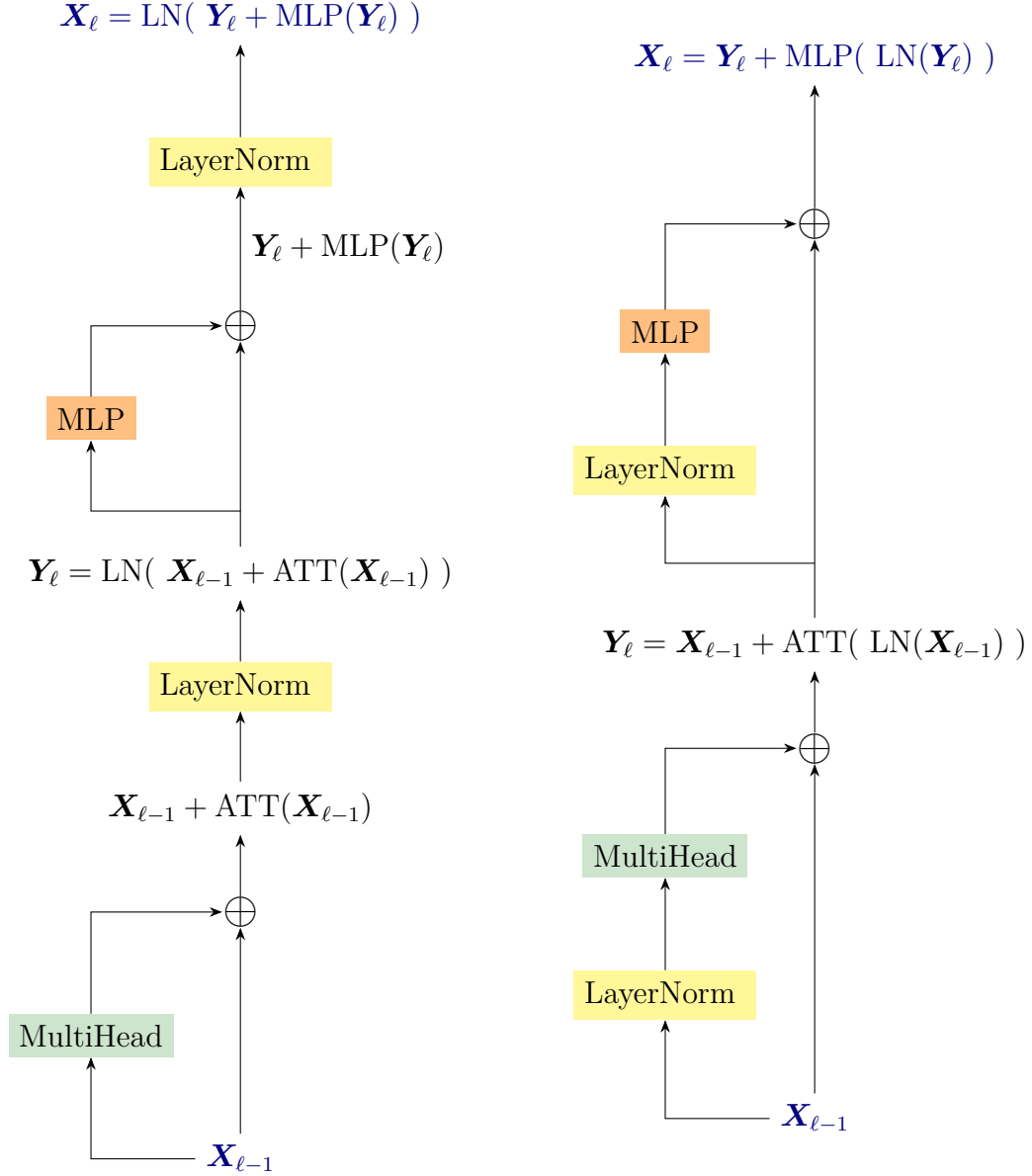
\begin{figure}
\label{figure:pre_post_LayerNorm3}
\hfill
\begin{minipage}{0.3\textwidth} 
 \begin{tikzpicture}[
    >=Stealth,
    blkg/.style={fill=green!30!gray!30,align=center,minimum width=15mm},
    blky/.style={fill=yellow!50,align=center,minimum width=15mm},
    add/.style={font=\Large, inner sep=0},
 ]
    \node[blue!50!black]       (in1)   at (0, 0) {${\bm{X}}_{\ell-1}$};
    \node[blkg] (l1) at (-2,4em)  {$\operatorname{MultiHead}$};
    \node[add]  (add) at (0,8em) {$\oplus$};
    \node[blky, above=6em of add] (l2) { $\operatorname{LayerNorm}$ };
    \node[above=2em of l2] (Yell) { ${\bm{Y}}_{\ell} = \operatorname{LN} (~ {\bm{X}}_{\ell-1} + \operatorname{ATT} ({\bm{X}}_{\ell-1}) ~)$ };
    \node[above=2em of add] (Xell) { $ {\bm{X}}_{\ell-1} + \operatorname{ATT} ({\bm{X}}_{\ell-1})$ };
    \node[fill=orange!50] (out) at (-2,24em) {$\operatorname{MLP}$};
    \node[add]  (add2) at (0,27em) {$\oplus$};
    \node[blky, above=4em of add2] (l3) { $\operatorname{LayerNorm}$ };
    \node[blue!50!black, above=3em of l3]  (out2) {${\bm{X}}_{\ell} = \operatorname{LN} (~ {\bm{Y}}_{\ell} + \operatorname{MLP} ({\bm{Y}}_{\ell}) ~)$};
    
    
    \draw[->] (in1) -- (add) ;
    \draw[->] (add) -- (Xell) ;
    \draw[->] (Xell) -- (l2) ;
    \draw[->] (in1) -| (l1) {};
    \draw[->] (l1) |- (add) {};
    \draw[->] (l2) -- (Yell) node[right, pos=0.5] {};
    \draw[->] (Yell) -- (add2) {};
    \draw[->] (out) |- (add2) node[left, pos=0.5] {};
    \draw[->] (0,21em) -| (out) {};
    
    \draw[->] (add2) -- (l3) node[right, pos=0.5] {${\bm{Y}}_{\ell} + \operatorname{MLP} ({\bm{Y}}_{\ell})$};
    \draw[->] (l3) -- (out2);
 \end{tikzpicture}
\end{minipage}\hfill
\begin{minipage}{0.3\textwidth} 
 \begin{tikzpicture}[
    >=Stealth,
    blkg/.style={fill=green!30!gray!30,align=center,minimum width=15mm},
    blky/.style={fill=yellow!50,align=center,minimum width=15mm},
    add/.style={font=\Large, inner sep=0},
 ]
    \node[blue!50!black]       (in1)   at (0, 0) {${\bm{X}}_{\ell-1}$};
    \node[blky] (l2) at (-2,4em) { $\operatorname{LayerNorm}$ };
    \node[blkg, above=3em of l2] (l1)  {$\operatorname{MultiHead}$};
    \node[add]  (add) at (0,12em) {$\oplus$};
    \node[above=2em of add] (Yell) { ${\bm{Y}}_{\ell} = {\bm{X}}_{\ell-1} + \operatorname{ATT} (~\operatorname{LN} ( {\bm{X}}_{\ell-1} ) ~)$ };

    \node[blky] (out) at (-2,21em) { $\operatorname{LayerNorm}$ };
    \node[fill=orange!50, above=3em of out] (l3) { $\operatorname{MLP}$ };
    \node[add]  (add2) at (0,29em) {$\oplus$};
    
    \node[blue!50!black, above=4em of add2]  (out2) {${\bm{X}}_{\ell} = {\bm{Y}}_{\ell} + \operatorname{MLP} (~ \operatorname{LN} ( {\bm{Y}}_{\ell} ) ~)$};


    \draw[->] (in1) -| (l2) {};
    \draw[->] (l2) -- (l1) {};
    \draw[->] (l1) |- (add) node[left, pos=0.5] {};
    \draw[->] (in1) -- (add) {};
    \draw[->] (out) -- (l3) {};
    \draw[->] (l3) |- (add2) node[left, pos=0.5] {};
     \draw[->] (add) -- (Yell) node[right, pos=0.5] {  };
     \draw[->] (Yell) -- (add2) node[right, pos=0.5] {  };
    \draw[->] (add2) -- (out2) { };
    \draw[->] (0,18em) -| (out) {};
 \end{tikzpicture}
\end{minipage}
\hfill \mbox{ \ }
\caption{$\operatorname{Post-LayerNorm}$ and $\operatorname{Pre-LayerNorm}$ Transformer blocks. In the $\operatorname{Post-LayerNorm}$ form, normalization is applied after each residual addition, whereas in the $\operatorname{Pre-LayerNorm}$ form, it is applied before each sublayer. The comparison highlights that $\operatorname{Pre-LayerNorm}$ preserves an exact identity path along the residual stream, which is one reason it is favored in modern LLMs.}
\end{figure}

A residual unit, with identity mapping, can be expressed in the general form
\begin{align}
{\bm{Y}}_\ell &= g({\bm{X}}_{\ell-1}) + {\mathcal{F}} ({\bm{X}}_{\ell-1};\Theta_\ell),
\label{equa7.11}
\\
{\bm{X}}_\ell &= f({\bm{Y}}_\ell),
\label{equa7.12}
\end{align}
where \({\bm{X}}_{\ell-1}\) and \({\bm{X}}_\ell\) are the input and output of the \(\ell\)-th sublayer, respectively, and $\mathcal{F}$ is a residual function with parameters \(\Theta_\ell\). In the Transformer setting, the relevant case is the identity skip connection $g(\bm{x})=\bm{x}$.

Therefore, the essential residual form is
\begin{equation}
{\bm{X}}_\ell = {\bm{X}}_{\ell-1} + {\mathcal{F}} ({\bm{X}}_{\ell-1};\Theta_\ell) ,
\label{equa7.13}
\end{equation}
possibly followed or preceded by normalization depending on whether one uses Post-LayerNorm or Pre-LayerNorm.
Equation \eqref{equa7.13} shows that a Transformer sublayer is not designed to overwrite the current state, but to add a learned correction to it. 
This is the identity-plus-correction viewpoint.

\begin{figure}
\begin{subfigure}{\textwidth}
\centering
 \begin{tikzpicture}[
    >=stealth,
    blkg/.style={fill=green!30!gray!30,align=center,minimum width=15mm},
    blky/.style={fill=yellow!50,align=center,minimum width=15mm},
    add/.style={font=\Large, inner sep=0},
 ]
    \node       (in1)   at (0, 0) {${\bm{X}}_{\ell-1}$};
    \node[blkg] (l1)    at (2,-1) {${\mathcal{F}}$};
    \node[add]  (add)   at (4, 0) {$\oplus$};
    \node[blky, right=3em of add] (l2) { $\operatorname{LayerNorm}$ };
    \node[blue!50!black, right=2em of l2] (out) {$\operatorname{LayerNorm} (~{\bm{X}}_{\ell-1} + {\mathcal{F}} ({\bm{X}}_{\ell-1})~)$};
    
    \draw[->,thick] (in1) -- node[right, pos=0.8] {} (add);
    \draw[->,thick] (add) -- (l2) node[above, pos=0.5] {};
    \draw[->,thick] (l2) -- (out);

    \draw[->,thick] (in1) |- (l1);
    \draw[->,thick] (l1) -| node[right, near end] {} (add);

    
 \end{tikzpicture}
\caption{Post-LayerNorm: normalize after residual addition.}
\label{figure:postLayer}
\end{subfigure}
\begin{subfigure}{\textwidth}
\centering
\begin{tikzpicture}
    [block/.style={draw,minimum width=#1,minimum height=1em},
    block/.default=10em,high/.style={minimum height=3em},
    node distance=2em, > = stealth]

    \node (n0) {${\bm{X}}_{\ell-1}$};
    \node[fill=yellow!50,align=center, minimum width=15mm, right=3em of n0] (n1) {$\operatorname{LayerNorm}$};
    \node[fill=green!30!gray!30, align=center, minimum width=15mm, right=3em of n1] (n2) {${\mathcal{F}}$};
    \node[right=3em of n2, font=\Large, inner sep=0, ] (n3) {$\oplus$};
    \node[blue!50!black, right=3em of n3, label={}] (n4) {${\bm{X}}_{\ell-1} + {\mathcal{F}} (~\operatorname{LayerNorm} ({\bm{X}}_{\ell-1})~)$};


    \foreach \i [count=\j from 1] in {0,...,3} 
    \draw[->, thick] (n\i) -- (n\j);
    \draw[->, thick] (n0) --++(0,-1) -| (n3);
\end{tikzpicture}
\caption{Pre-LayerNorm: normalize before the sublayer.}
\label{fig1:Pre-norm_residualA}
\end{subfigure}
\caption{Generic residual unit in Post-LayerNorm and Pre-LayerNorm form. In both cases the sublayer contributes a learned correction $\mathcal{F}$ to the incoming state, but the location of normalization differs: Post-LayerNorm applies normalization after the residual sum, whereas Pre-LayerNorm applies it before the sublayer. 
}
\end{figure}
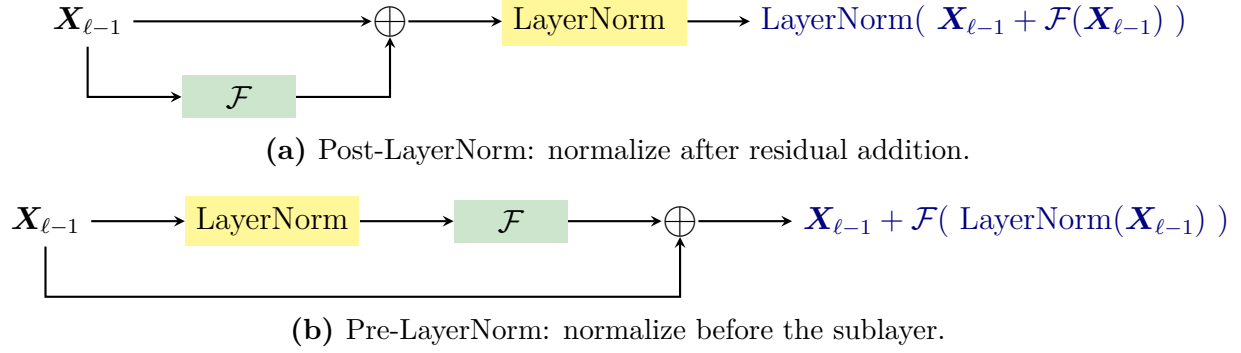

In the Pre-LayerNorm setting, layer normalization is treated as part of the sublayer rather than as a post-processing step after residual addition.
For interpretive clarity, the residual-stream equations in this subsection are written at the level of a single token representation, so that ${\bm{x}}_{\ell}, {\bm{a}}_{\ell}$, and ${\bm{m}}_{\ell}$ denote the residual-stream vector, the attention contribution, and the MLP contribution for one token position at depth $\ell$, respectively.
Here, the transformer block \(\ell\) consists of alternating layers of ATT and MLP stages, i.e.,
\begin{align}
{\bm{a}}_\ell &= \operatorname{ATT} \bigl(\operatorname{LN} ( {\bm{x}}_{\ell-1})\bigr),
\label{equa7.14}
\\
{\bm{m}}_\ell &= \operatorname{MLP}\ \bigl(\operatorname{LN}( {\bm{x}}_{\ell-1} + {\bm{a}}_\ell)\bigr),
\label{equa7.15}
\\
{\bm{x}}_\ell &= {\bm{x}}_{\ell-1} + {\bm{a}}_\ell + {\bm{m}}_\ell .
\label{equa7.16}
\end{align}
These equations show that the vectors computed in the attention and MLP modules are added to the residual stream at each layer. 
Hence, the residual stream ${\bm{x}}_\ell$ represents an accumulated sum of contributions from the attention and MLP modules together with the incoming stream,
\begin{equation}
{\bm{x}}_\ell
=
{\bm{x}}_0
+
\sum_{i=1}^{\ell} {\bm{a}}_i
+
\sum_{i=1}^{\ell} {\bm{m}}_i .
\label{equa7.17}
\end{equation}

This is one of the most important structural equations in the subsection.
It shows that the hidden state at depth $\ell$ is an accumulated sum of writes into a shared state space. 
This is why the term residual stream is so apt: information is not stored in isolated compartments, but is repeatedly written into and read from a common sequence representation.

\subsection{The Transformer block as a composite operator}
The discussion so far has introduced the main components of a Transformer block---self-attention, the feed-forward sublayer, normalization, and residual connections. 
We now assemble these ingredients into one mathematical object. The Transformer block becomes a composite operator acting on the sequence representation. With this view, we can write the block as a composition of structured maps, distinguish clearly between its linear and nonlinear parts, and examine the order in which normalization, attention, the MLP, and residual addition are applied. Depth then has a simple interpretation: we repeatedly compose these maps and refine the representation layer by layer.

\subsubsection{One block as a composition of structured maps}
Each Transformer block \(\ell\) is a structured map
\begin{equation}
{\mathcal{T}}_{\ell}(\,\cdot\,;\Theta_\ell):\mathbb{R}^{n\times d_{\mathrm{model}}}\to\mathbb{R}^{n\times d_{\mathrm{model}}},
\label{equa10.1}
\end{equation}
acting on the residual-stream matrix. Its input-output relation is
\begin{equation}
{\bm{X}}_\ell = {\mathcal{T}}_{\ell}({\bm{X}}_{\ell-1};\Theta_\ell),
\qquad
\ell=1,\dots,n_{\mathrm{blocks}},
\label{equa10.2}
\end{equation}
with
\begin{equation}
{\bm{X}}_\ell \in \mathbb{R}^{n\times d_{\mathrm{model}}}.
\label{equa10.3}
\end{equation}

As discussed earlier, each Transformer block consists of two distinct stages or sublayers, with layer normalization in each stage and residual connections. The first stage operates across the sequence of tokens and is the multi-headed self-attention sublayer. The second stage operates across the features of each token and is the multilayer perceptron (MLP).

Accordingly, a Transformer block is not best viewed as one undifferentiated nonlinear map. Rather, it is a composition of several structured operations: row-wise normalization, token-mixing attention, residual addition, row-wise normalization again, feature-mixing MLP transformation, and a second residual addition.

\subsubsection{What is linear and what is nonlinear?}
The structured nature of the block becomes clearer if one separates its linear and nonlinear pieces. 
In the attention stage, the projections
${\bm{Q}}_\ell = {\bm{X}}_{\ell-1}{\bm{W}}_Q^{(\ell)}$, ${\bm{K}}_\ell = {\bm{X}}_{\ell-1}{\bm{W}}_K^{(\ell)}$, ${\bm{V}}_\ell = {\bm{X}}_{\ell-1}{\bm{W}}_V^{(\ell)}$
are linear in the input matrix. The score matrix $ {\bm{S}}_\ell $ defined in \eqref{eq:Sl}
is bilinear in the projected states, while the attention matrix
${\bm{P}}_\ell = \operatorname{softmax}({\bm{S}}_\ell)$ is nonlinear because of the row-wise softmax. 

In the MLP stage, the standard dense form is
\begin{equation}
\operatorname{MLP}({\bm{Y}}_\ell)
=
\bigl(
\operatorname{ReLU}({\bm{Y}}_\ell {\bm{W}}_{\mathrm{up}}^\ell + {\bm{b}}_{\mathrm{up}}^\ell)
\bigr){\bm{W}}_{\mathrm{down}}^\ell + {\bm{b}}_{\mathrm{down}}^\ell,
\label{equa10.14}
\end{equation}
which is linear before and after the activation, but nonlinear overall. 
Layer normalization,
\begin{equation}
\operatorname{LN}(\bm{x})
=
\frac{ {\bm{x}}- \mu(\bm{x}) }{\sqrt{ \sigma^2( \bm{x} )+\varepsilon}}
\odot \boldsymbol{\gamma} + \boldsymbol{\beta} ,
\label{equa10.15}
\end{equation}
is likewise nonlinear because the centering and scaling depend on the input row $\bm{x}$.


\subsubsection{Residual addition and operator ordering}
%
We add residual updates twice in a Transformer block: once after attention and once after the MLP. These additions matter. They specify when, and in what order, each component writes information into the residual stream.
For the Pre-LayerNorm block, the ordering is
\begin{equation}
{\bm{X}}_{\ell-1}
\;\xrightarrow{\;\operatorname{LN}\;}\;
\operatorname{ATT}
\;\xrightarrow{\;+\;{\bm{X}}_{\ell-1}\;}
{\bm{Y}}_\ell
\;\xrightarrow{\;\operatorname{LN}\;}\;
\operatorname{MLP}
\;\xrightarrow{\;+\;{\bm{Y}}_\ell\;}
{\bm{X}}_\ell.
\label{equa10.16}
\end{equation}

This ordering matters. In general, attention followed by MLP is not the same as MLP followed by attention, because the attention stage mixes across token positions whereas the MLP stage mixes across feature coordinates within each token. Likewise, normalizing before a sublayer is not the same as normalizing after adding its residual contribution.
The Transformer block is therefore a noncommutative composition of structured maps. Its behavior depends not only on the ingredients it contains, but also on the order in which those ingredients act.

\subsubsection{Progressive representation refinement}
By composing these transformer blocks, the Transformer progressively adjusts token embeddings so that they do not merely encode individual words but instead contextual meaning. 
Therefore, depth should be understood as progressive representation refinement.

At shallow layers, the rows of \({\bm{X}}_\ell\) remain relatively close to the input token-plus-position embeddings. At deeper layers, repeated attention and MLP updates integrate more contextual information, reshape features, and propagate task-relevant structure across the sequence. Thus, the residual stream evolves from an embedding-level representation into a highly contextualized matrix whose rows support next-token prediction and other emergent capabilities of large language models.

A Transformer block is a composite operator acting on the residual-stream matrix ${\bm{X}}_{\ell-1}$. 
It combines row-wise normalization, multi-head self-attention, residual addition, a positionwise feed-forward map, and a second residual addition. 
In Pre-LayerNorm form, the block is described by Equations
\eqref{equa4.6}--\eqref{equa7.8}, and a full decoder stack is obtained by composing such blocks across depth as in \eqref{equa4.5}.
From a linear algebra viewpoint, the significance of the Transformer block is that it organizes a sequence of matrix-valued operations into a stable identity-plus-correction architecture. The resulting depth is best understood as repeated refinement of a shared residual-stream state rather than as recurrence in time.

\subsection{Internal workings of decoder-only LLMs}
At the input layer, \(\ell=0\), textual prompts undergo tokenization and are combined with positional encoding to create an initial high-dimensional embedding. As in Sections~\ref{sec:transformer_architecture}--\ref{sec:single_head}, the initial hidden-state matrix is
\begin{equation}
{\bm{X}}_0 = {\bm{U}} {\bm{W}}_E + {\bm{P}} \;\;\in \mathbb{R}^{n\times d_{\mathrm{model}}},
\label{equa11.1}
\end{equation}
where ${\bm{U}}$ is the token matrix, ${\bm{W}}_E$ is the embedding matrix, and ${\bm{P}}$ encodes positional structure. These embeddings of the $n$ tokens are stacked row-wise to form the matrix ${\bm{X}}_0$, which then passes through $n_{\mathrm{blocks}}$ Transformer blocks.

Hence, the output of a Transformer with $n_{\mathrm{blocks}}$ transformer blocks is the composition of \(n_{\mathrm{blocks}}\) functions \({\mathcal{T}}_{\ell}\)
shown in Equation~\ref{equa4.5} 
In a decoder-only large language model, these layerwise updates progressively transform the initial token-plus-position embeddings into contextual hidden states that summarize the visible prefix at each token position.

In the final representation, an additional layer normalization is applied,
\begin{equation}
{\bm{X}}_{n_{\mathrm{blocks}}}
\leftarrow
\operatorname{LN}({\bm{X}}_{n_{\mathrm{blocks}}}),
\label{equa11.3}
\end{equation}
where the final hidden states are passed into a bias-free linear layer to obtain logits. The layer projects each output vector ${\bm{x}}_i^{(n_{\mathrm{blocks}})}\in\mathbb{R}^{d_{\mathrm{model}}}$ into a larger vector called a logits vector ${\boldsymbol{\ell}}_i \in \mathbb{R}^{|{\mathcal{V}}|}$, i.e.,
\begin{align}
{\boldsymbol{\ell}}_i &= {\bm{x}}_i^{(n_{\mathrm{blocks}})} {\bm{W}}_U  \;\;\in \mathbb{R}^{|\mathcal{V}|},
\label{equa11.4}
\intertext{where}
{\bm{W}}_U &\in \mathbb{R}^{d_{\mathrm{model}}\times |\mathcal{V}|}
\label{equa11.5}
\end{align}
is the unembedding weight matrix. The softmax layer then turns these scores into probabilities,
\begin{equation}
{\bm{p}}_i = \operatorname{softmax} ( {\boldsymbol{\ell}}_i ) \;\;\in [0,1]^{|\mathcal{V}|},
\label{equa11.6}
\end{equation}
and the token with highest probability may be selected as the output for that time step.

The defining architectural feature of modern LLMs is that they are predominantly decoder-only Transformers trained autoregressively. In this setting, the attention operator is causally masked, so the hidden state at position \(i\) may depend only on tokens \(1,\dots,i\). Thus the row \({\bm{X}}_\ell(i,:)\) of the residual-stream matrix evolves as a contextual representation of the visible prefix rather than of the full sequence.

This is the central difference between the generic Transformer architecture and the decoder-only LLM setting. The block equations remain the same at the local level, but the self-attention operator is constrained by causality. Hence decoder-only LLMs should be understood as deep stacks of masked Transformer blocks acting on a shared residual stream.

\subsubsection{The residual stream as working memory}
A useful way to describe the internal workings of a decoder-only LLM is to view the residual stream as its working memory.
At layer \(\ell\), the model state is the matrix
\begin{equation}
{\bm{X}}_\ell \in \mathbb{R}^{n\times d_{\mathrm{model}}},
\label{equa11.7}
\end{equation}
and each row corresponds to one token position. Through repeated attention and MLP updates, each row is gradually enriched by information written from earlier positions and by nonlinear feature transformations applied locally.

Using the residual-stream viewpoint from Section~\ref{sec:residual_1}, one may write
\begin{equation}
{\bm{x}}_\ell
=
{\bm{x}}_0
+
\sum_{i=1}^{\ell} {\bm{a}}_i
+
\sum_{i=1}^{\ell} {\bm{m}}_i,
\label{equa11.8}
\end{equation}
where the ${\bm{a}}_i$ terms denote attention contributions and the ${\bm{m}}_i$ terms denote MLP contributions. Thus, the hidden state at depth $\ell$ is not a fresh representation computed from scratch, but an accumulated superposition of writes into a shared vector space.

In decoder-only models, this working-memory interpretation is especially natural: at token position $i$, the row ${\bm{X}}_\ell(i,:)$ stores what the model currently \emph{knows} about the prompt prefix up to $i$, and this row is progressively refined across depth.

One reason large language models are effective is that earlier token information
is not discarded after one layer. Instead, information extracted by attention at
one depth can be preserved in the residual stream, transformed by later MLP
blocks, and read again by deeper attention heads. In this sense, \emph{context
  is reused both across token positions and across layers.}

This reuse happens through two coupled mechanisms. First, attention allows a
token at position $i$ to gather information from earlier positions $j \le i$.
Second, the residual stream preserves the result of that gathering so that later
layers can build further computations on top of it.  The model therefore
supports iterative context refinement rather than one-shot contextualization.

\subsubsection{Decoder-only LLM examples: GPT-3, Llama-3, and Gemma}
Decoder-only LLMs instantiate the same broad Transformer template at very different scales. For example, GPT-3 175B uses 96 blocks, while Llama-3 405B uses 126 blocks \cite{brown2020gpt3}.
For GPT-3 175B, one may write
\begin{equation}
n_{\mathrm{blocks}} = 96,\qquad
d_{\mathrm{model}} = 12{,}288,\qquad
d_{\mathrm{mlp}} = 49{,}152,\qquad
n_{\mathrm{heads}} = 96.
\label{equa11.9}
\end{equation}
For Llama-3 405B, one may write \cite{dubey2024llama3}
\begin{equation}
n_{\mathrm{blocks}} = 126,\qquad
d_{\mathrm{model}} = 16{,}384,\qquad
d_{\mathrm{mlp}} = 53{,}248,\qquad
n_{\mathrm{heads}} = 128.
\label{equa11.10}
\end{equation}
GPT-3 uses a standard dense MLP with ReLU-style design, whereas Llama-style models use gated activations such as SwiGLU and grouped-query style attention variants.
These architectural differences are summarized in Table~\ref{table:keyParameters}.
%
\begin{table}[htb]
\begin{center}
\begin{tabular}{c|c|c|c} \toprule
                & & \textbf{GPT-3 175B} & \textbf{Llama-3 405B} \\ \midrule
Vocabulary Size & $n_{\texttt{vocab}} \equiv |\mathcal{V}|$ & \num[group-separator={,}]{50257} & \num[group-separator={,}]{128256} \\
Layers          & $n_{\texttt{blocks}}$ &  96  & 126 \\
Model Dimension & $d_{\texttt{model}}$ & \num[group-separator={,}]{12288} & \num[group-separator={,}]{16384} \\
MLP Dimension   & $d_{\texttt{mlp}}$ & \num[group-separator={,}]{49152} & \num[group-separator={,}]{53248} \\                          
Attention Heads & $n_{\texttt{heads}}$ & 96  & 128 \\
Key/Value Heads & &     & 8 \\
Activation Function & $\sigma$ & ReLU & SwiGLU \\ 
$d_{\texttt{key}}, d_{\texttt{query}}, d_{\texttt{value}}$ &  & 128, 128, 128 & \\ \bottomrule
\end{tabular}
\end{center}
\caption{Comparison of key architectural hyperparameters for GPT-3 175B and Llama-3 405B. The table shows that, although both models follow the decoder-only Transformer template, they differ substantially in vocabulary size, depth, model width, feed-forward dimension, head configuration, and activation design.}
\label{table:keyParameters}
\end{table}

This is already enough to show that modern LLM families are not identical instantiations of one decoder-only template. They share the same broad operator structure, but differ in width, depth, feed-forward design, positional mechanism, and attention parameterization.

Gemma provides another useful point of comparison \cite{geminiteam2024gemini}. It may be viewed as an efficiency-oriented decoder-only family that remains architecturally close to the Transformer backbone while emphasizing deployability and long-context engineering.

\subsubsection{Parameter counting as a worked example}
A useful way to make the architecture concrete is through parameter counting.
For GPT-3 175B, the model may be broken into MultiHead parameters, MLP parameters, embedding and unembedding parameters, and normalization parameters, yielding a total of $175{,}191{,}908{,}352$
%
%
parameters. 
For Llama-3 405B, the analogous breakdown yields $405{,}872{,}984{,}064$
%
%
parameters.

These parameter-count examples are valuable because they illustrate one of the recurring themes of the section: the MLP blocks dominate the parameter budget in dense decoder-only models. In GPT-3, the MLP contribution is larger than the multi-head attention contribution, and in Llama-3 the MLP share is even more dominant because of the gated three-matrix design.
This breakdown is shown explicitly in Tables~\ref{table:parameterCount_GPT3} and \ref{table:parameterCount_Llama3} for GPT-3 175B and Llama-3 405B, respectively.
%
\begin{table}[t]
\begin{center}
\begin{tabular}{c|clr} \toprule
 & \textbf{GPT-3 175B} & & \# parameters
 \\ \midrule
$\operatorname{MultiHead}$  & $( 3 d_{\texttt{model}} d_{\texttt{head}} +  d_{\texttt{head}} d_{\texttt{model}} ) \times n_{\texttt{heads}} ~ \times n_{\texttt{blocks}} $
& = &  \num[group-separator={,}]{57982058496}
\\
$\operatorname{MLP}$  & $( 2 d_{\texttt{mlp}} d_{\texttt{model}} + (d_{\texttt{mlp}} + d_{\texttt{model}}) ) ~ \times n_{\texttt{blocks}}$
& = &  \num[group-separator={,}]{115970015232} 
\\
$\bm{W}_E, \bm{W}_U$  &  $(2 d_{\texttt{model}} n_{\texttt{vocab}} )$ & = &  \num[group-separator={,}]{1235116032} 
\\
$\boldsymbol{\gamma}, \boldsymbol{\beta}$ & $2 \times (2 d_{\texttt{model}}) \times n_{\texttt{blocks}}$ & = & \num[group-separator={,}]{4718592}
\\ \midrule
 & \textbf{Total} & = &  \num[group-separator={,}]{175191908352}  \\ \bottomrule
\end{tabular}
\end{center}
\caption{Parameter breakdown of GPT-3 175B by major architectural component. The table decomposes the total parameter count into attention, MLP, embedding/unembedding, and normalization contributions, illustrating that the dense MLP blocks dominate the parameter budget.}
\label{table:parameterCount_GPT3}
\end{table}

\begin{table}[t]
\begin{center}
\begin{tabular}{c|clr} \toprule
 & \textbf{Llama-3 405B} & & \# parameters
 \\ \midrule
$\operatorname{MultiHead}, {\bm{W}}_K, {\bm{W}}_V$  & $2 \times d_{\texttt{head}} d_{\texttt{model}} \times n_{\texttt{blocks}} $
& = &  \num[group-separator={,}]{4227858432}
\\
$\operatorname{MultiHead}, {\bm{W}}_Q, {\bm{W}}_O$  & $2 \times d_{\texttt{model}} d_{\texttt{model}} \times n_{\texttt{blocks}} $
& = &  \num[group-separator={,}]{67645734912}
\\
$\operatorname{MLP}$  & $( 3 d_{\texttt{mlp}} d_{\texttt{model}} + (2 d_{\texttt{mlp}} + d_{\texttt{model}}) ) ~ \times n_{\texttt{blocks}}$
& = &  \num[group-separator={,}]{329788440576} 
\\
${\bm{W}}_E, {\bm{W}}_U$  &  $(2 d_{\texttt{model}} n_{\texttt{vocab}} )$ & = &  \num[group-separator={,}]{4202692608} 
\\
$\boldsymbol{\gamma}, \boldsymbol{\beta}$ & $2 \times (2 d_{\texttt{model}}) \times n_{\texttt{blocks}}$ & = & \num[group-separator={,}]{8257536}
\\ \midrule
 & \textbf{Total} & = &  \num[group-separator={,}]{405872984064}  \\ \bottomrule
\end{tabular}
\end{center}
\caption{Parameter breakdown of Llama-3 405B by major architectural component. The table decomposes the total parameter count into attention, MLP, embedding/unembedding, and normalization contributions, illustrating that the gated MLP blocks account for the dominant share of parameters in this model.}
\label{table:parameterCount_Llama3}
\end{table}

\subsubsection{A modern sparse variant: gpt-oss}
A final useful example is the gpt-oss family, because it shows how the decoder-only LLM template can be modified by sparse Mixture-of-Experts design. In this setting, the overall architecture remains decoder-only and residual-stream based, but the dense feed-forward stage is replaced by a routed sparse expert module \cite{agarwal2025gptoss}.

This example is especially useful for the next section, because it shows that modern LLMs need not be dense all the way through. The same decoder-only residual-stream architecture can be combined with sparse expert routing, changing the scaling behavior without abandoning the Transformer block structure.



\subsection{Mixture-of-Experts Transformers}
Mixture-of-Experts (MoE) Transformers modify the feed-forward stage of the Transformer by replacing a single dense MLP with a routed collection of expert MLPs. This change preserves the basic Transformer architecture while making the channel-mixing stage sparse and input-dependent: different tokens may activate different experts at the same layer. The result is a model whose total parameter count can be made very large, while the computation used for any given token remains much smaller \cite{shazeer2017outrageously}.

\subsubsection{From dense feed-forward blocks to experts}
In a standard dense Transformer, the feed-forward block applies the same per-token nonlinear map to every token representation at a given layer. 
Thus, although attention is input-dependent, the MLP stage remains dense and shared across all tokens.

Mixture-of-Experts (MoE) Transformers replace this single dense feed-forward block with a collection of expert blocks together with a routing mechanism that selects only a small subset of experts for each token. In this way, the model can have a very large total parameter count while activating only a much smaller number of parameters for each token during computation. This is the central scaling idea behind sparse expert language models such as Switch Transformer \cite{fedus2022switch}, Mixtral \cite{jiang2024mixtral}, and OpenAI’s gpt-oss family \cite{agarwal2025gptoss}.

Let
\begin{equation}
{\bm{x}}_i^\ell \in \mathbb{R}^{d_{\mathrm{model}}}
\label{equa12.1}
\end{equation}
denote the representation of token position \(i\) at layer \(\ell\). 
In a dense Transformer, the MLP stage applies one map
\begin{equation}
{\bm{x}}_i^\ell \mapsto {\operatorname{MLP}}_\ell ( {\bm{x}}_i^\ell ) ,
\label{equa12.2}
\end{equation}
where ${\operatorname{MLP}}_\ell$ is the shared positionwise feed-forward transformation of layer \(\ell\).


In an MoE layer, the single dense map \(\operatorname{MLP}_\ell\) is replaced by a family of expert maps \cite{fedus2022switch}
\begin{equation}
E_{\ell, 1} ,  \; E_{\ell, 2} , \; \dots , \; E_{\ell, M} , 
\label{equa12.3}
\end{equation}
where $E_{\ell, m} = \operatorname{MLP}_{\ell,m}$ denotes the $m$-th expert operator at layer $\ell$. 
These experts are combined by a routing mechanism to define the full MoE transformation $\operatorname{MoE}_\ell$, which at a high level may be written as
\begin{equation}
\operatorname{MoE}_\ell({\bm{x}}_i^\ell)
=
\sum_{m \,\in\, \mathcal{S}_\ell({\bm{x}}_i^\ell)}
\pi_{\ell,m}({\bm{x}}_i^\ell) \, E_{\ell,m} ({\bm{x}}_i^\ell) ,
\label{equa12.4}
\end{equation}
where $\mathcal{S}_\ell({\bm{x}}_i^\ell)$ is the set of selected experts and \(\pi_{\ell,m}({\bm{x}}_i^\ell)\) are the corresponding routing weights, which satisfy
\begin{equation}
\pi_{\ell,m} ( {\bm{x}}_i^\ell ) \ge 0,
\qquad
\sum_{m \,\in\, \mathcal{S}_\ell ( {\bm{x}}_i^\ell ) }
\pi_{\ell, m} ( {\bm{x}}_i^\ell ) = 1 .
\label{equa12.5}
\end{equation}
Thus the channel-mixing stage becomes a sparse, input-dependent operator rather than a single shared dense operator.

\subsubsection{Routing, sparse expert computation, and operator interpretation}
The essential new ingredient is the router.
Given a token representation \( {\bm{x}}_i^\ell\), it computes a score vector over the $M$ experts. A standard formulation is
\begin{equation}
{\bm{r}}_i^\ell
=
{\bm{x}}_i^\ell {\bm{W}}_R^\ell + {\bm{b}}_R^\ell
\;\;\in\; \mathbb{R}^{M},
\end{equation}
where ${\bm{W}}_R^\ell \in \mathbb{R}^{d_{\mathrm{model}}\times M}$ and ${\bm{b}}_R^\ell \in \mathbb{R}^{M}$.
Applying softmax gives routing probabilities
\begin{equation}
{\bm{p}}_i^\ell
=
\operatorname{softmax}( {\bm{r}}_i^\ell ) ,
\label{equa12.6}
\end{equation}
with
\begin{equation}
\sum_{m=1}^{M} p_{i,m}^\ell = 1 .
\label{equa12.7}
\end{equation}
Thus, $p_{i,m}^\ell$ denotes the $m$-th component of the routing vector ${\bm{p}}_i^\ell$, and corresponds to the earlier high-level routing weight $\pi_{\ell,m} ( {\bm{x}}_i^\ell )$.

However, the key point is that one typically does not evaluate all experts for every token. Instead, only the top-\(k\) experts are selected according to the router scores. If \({\mathcal{T}}_i^\ell \subseteq \{1,\dots,M\}\) is the set of selected experts for token \(i\), then \( | {\mathcal{T}}_i^\ell|=k \), with \(k\) usually much smaller than \(M\). 
This is what makes the channel-mixing computation sparse.
Here ${\mathcal{T}}_i^\ell$ denotes the set of selected experts for token position $i$ at layer $\ell$; it is the same object previously denoted by $\mathcal{S}_\ell({\bm{x}}_i^\ell)$.

Once the selected experts have been chosen, the full MoE transformation for token $i$ is formed by a weighted combination of the selected expert outputs:
\begin{equation}
\operatorname{MoE}_\ell ( {\bm{x}}_i^\ell )
=
\sum_{m \,\in\, {\mathcal{T}}_i^\ell}
p_{i, m}^\ell\, E_{\ell, m} ( {\bm{x}}_i^\ell ) .
\label{equa12.8}
\end{equation}
This is the defining MoE operation: a token-dependent sparse combination of expert maps.
In the top-1 case, this reduces to a single expert output multiplied by its routing weight. In the top-2 or top-4 case, several expert outputs are combined. Thus the dense feed-forward stage ${\bm{x}} \mapsto \operatorname{MLP}_\ell( {\bm{x}} )$
is replaced by a routed family of maps
\begin{equation}
{\bm{x}}
\mapsto
\sum_{m=1}^{M} \pi_{\ell, m} ( {\bm{x}} )\, E_{\ell, m} ( {\bm{x}} ),
\label{equa12.10}
\end{equation}
where the gating coefficients $\pi_{\ell, m} ( {\bm{x}} )$ are sparse and input-dependent.

The main appeal of Mixture-of-Experts is that it decouples total parameter count from the number of parameters activated for a given token.
Suppose each expert contains approximately \(N_E\) parameters and there are \(M\) experts. Then the expert pool contributes roughly
\begin{equation}
N_{\mathrm{experts}} \approx M N_E
\label{equa12.11}
\end{equation}
parameters in total. 
But if only \(k\) experts are active for a given token, then the token-level active parameter budget is only
\begin{equation}
N_{\mathrm{active}} \approx k N_E.
\label{equa12.12}
\end{equation}

This is the central scaling advantage of MoE. 
One can increase capacity by increasing the number of experts \(M\), while keeping the active compute much closer to that of a dense model as long as \(k\) remains small.

MoE introduces a difficulty that dense Transformers do not have: the router may send too many tokens to only a few experts. In that case, some experts become overloaded while others are rarely used. This is both an optimization problem and a systems problem.
If \(n_m^\ell\) denotes the number of tokens routed to expert \(m\) in layer \(\ell\), then one would like the vector $(n_1^\ell,\dots,n_M^\ell)$
to remain reasonably balanced rather than sharply concentrated. 
For this reason, MoE models typically introduce auxiliary load-balancing losses or capacity constraints.
Thus the MoE design problem concerns both representational power and the distribution of computation across experts in a way that is statistically useful and computationally feasible.
These routing and sparsity principles appear in several modern MoE architectures.

%
Switch Transformer \cite{fedus2022switch} is an early and influential sparse Transformer. Its routing is simple: choose the top-1 expert. This design shows how sparse expert models can grow to very large total parameter counts while keeping per-token computation manageable. Mixtral \cite{jiang2024mixtral} gives a modern open-weight case.
In Mixtral $8\times 7B$, each MoE layer contains 8 experts, and each token is routed to 2 experts. That makes Mixtral a useful example of MoE deployment in a high-performance open LLM. The gpt-oss family gives a GPT-style MoE case \cite{agarwal2025gptoss}. Here, the model stays decoder-only. Each MoE block contains many experts, but the router activates only a few experts for each token.
This illustrates how sparse routing can be combined with the standard residual-stream architecture of decoder-only Transformers.
These examples show that Mixture-of-Experts is not a single architecture but a design family within the Transformer framework. What remains common is the routed sparse-expert mechanism; what varies is the number of experts, the number of active experts, the routing rule, the balancing mechanism, and the scale of the model.

\medskip
From a linear-algebra viewpoint, MoE is especially interesting because it changes the nature of the MLP stage. In a dense Transformer, each layer contains one shared per-token feed-forward operator acting on all token rows. In an MoE Transformer, each token sees a token-dependent sparse combination of expert operators. Thus MoE extends the Transformer’s design principle of data-dependent computation. Attention already makes token mixing input-dependent by computing a data-dependent stochastic matrix. MoE makes the channel-mixing stage input-dependent as well by selecting which expert operators are applied to each token. In this sense, the dense feed-forward stage ${\bm{X}} \mapsto \operatorname{MLP}({\bm{X}})$
is replaced by a routed family of sparse per-token maps ${\bm{X}}
\mapsto
\operatorname{MoE}({\bm{X}})$,
whose active components depend on the current token states through the router. MoE should therefore be viewed as a genuine architectural generalization of the Transformer block rather than as a mere optimization device.



\subsection{Training dynamics}
Training a decoder-only large language model combines a simple probabilistic objective with a highly structured computational pipeline.
At the probabilistic level, the model is trained to predict the next token from the visible prefix. At the computational level, this objective is implemented through masked parallel operations on full sequences, followed by gradient-based optimization through depth. The section develops these training dynamics, then turns to the later fine-tuning and alignment stages used in modern chat-oriented models.

\begin{figure}[tb]
\begin{center} 
  \parbox{0.98\textwidth}{
  \hrule   
  \noindent \vrule \hfill
  \begin{minipage}{0.45\textwidth}
    \includegraphics[width=0.99\textwidth]{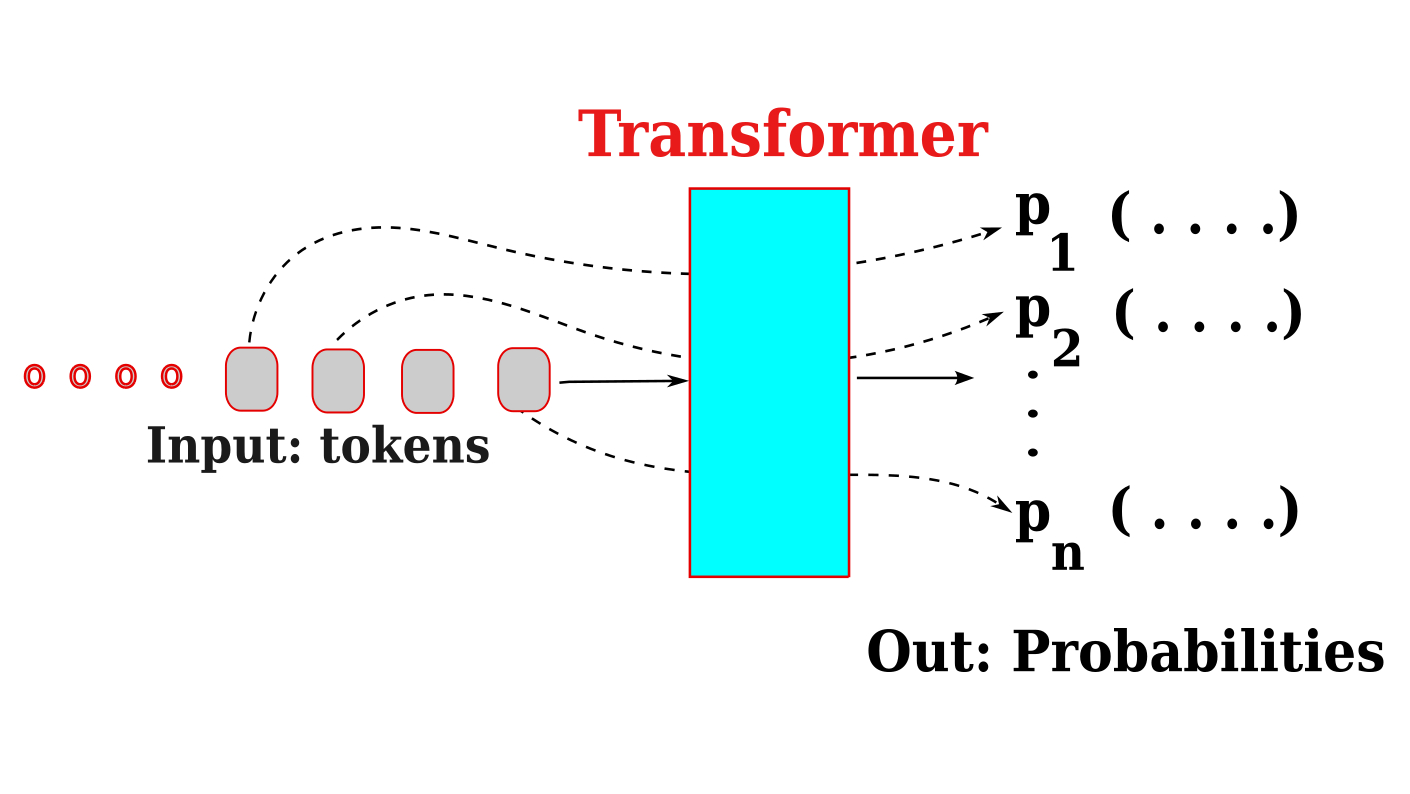}
    \centerline{
      \parbox{0.85\textwidth}{Training: get weights s.t. output
        probabilities for next token match target
      }}
       \mbox{ \ \ } \\ \mbox{\ }
  \end{minipage}  \hfill \vrule \hfill
  \begin{minipage}{0.42\textwidth}
    \includegraphics[width=0.99\textwidth]{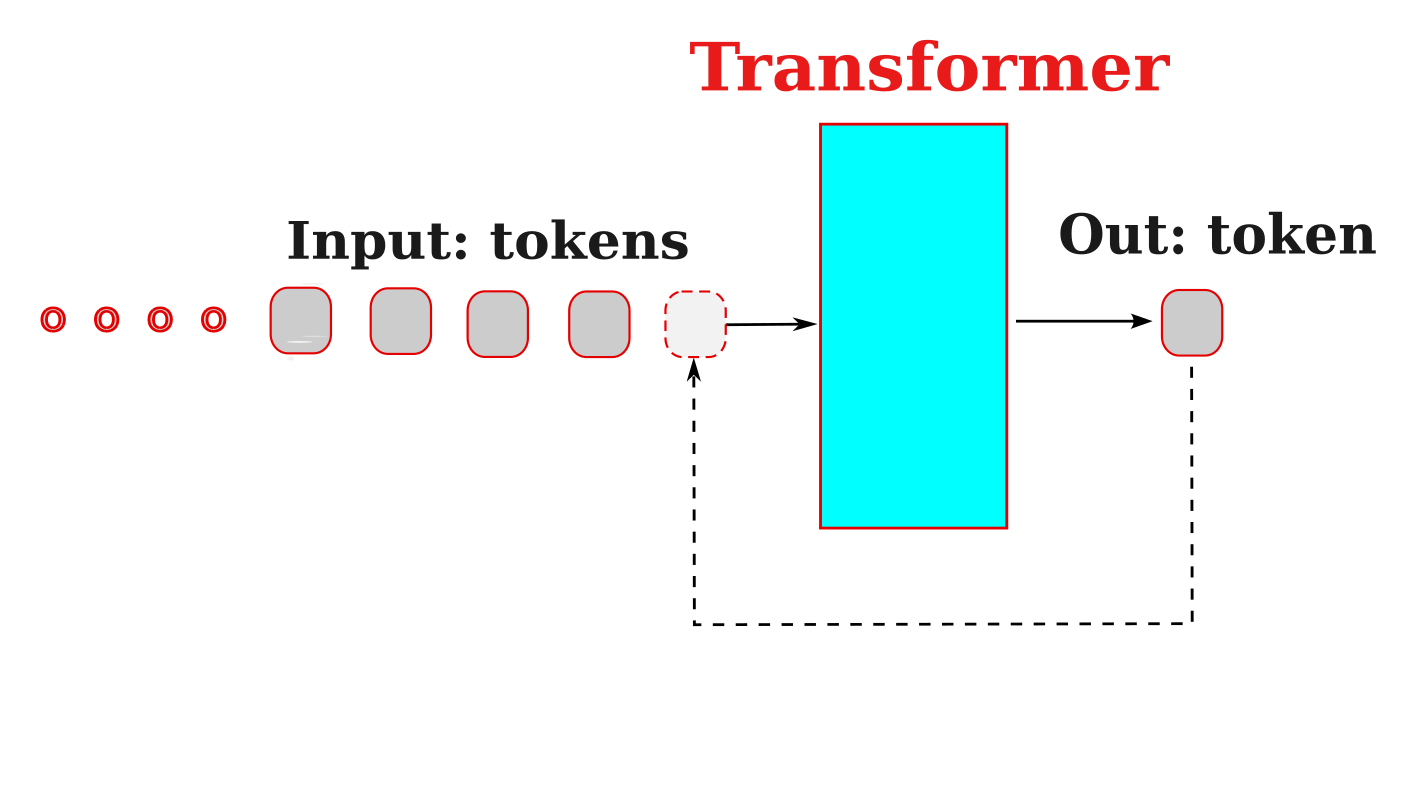}
   \centerline{ \parbox{0.85\textwidth}{ Generation: get next token.
   Re-inject as input. Repeat} }  \\        \mbox{ \ \ } \\ \mbox{\ }
\end{minipage}
\hfill \vrule \hrule  \mbox{ \ }
}
\end{center}
\caption{Training and inference in a decoder-only Transformer. During training, the model processes an entire token sequence in parallel and adjusts its weights so that the output probabilities match the target next tokens at all positions. During inference, by contrast, generation is autoregressive: the predicted next token is fed back as input, and the process is repeated sequentially.}
\label{fig:bigpicture}
\end{figure}

\subsubsection{Autoregressive training, masking, and optimization}
The training objective for autoregressive language modeling is the cross-entropy loss, equivalently the negative log-likelihood of the training sequence, given by
\begin{equation}
\mathcal{L}_\Theta
=
\frac{1}{n}
\sum_{t=1}^{n}
-\log p_{\mathrm{LM}}(u_t \mid u_1,\dots,u_{t-1}) .
\label{equa13.1}
\end{equation}
This is the standard next-token prediction objective used in autoregressive Transformer language models, including GPT-style models.
The distinction between parallel training and sequential autoregressive generation is illustrated schematically in Figure~\ref{fig:bigpicture}.

If \({\bm{X}}_L \in \mathbb{R}^{n\times d_{\mathrm{model}}}\) denotes the final hidden-state matrix after \(L\) blocks, and if
\begin{equation}
{\bm{W}}_U \in \mathbb{R}^{d_{\mathrm{model}}\times | {\mathcal{V}} |}
\label{equa13.2}
\end{equation}
is the unembedding matrix, then the logits are
\begin{equation}
{\bm{Z}} = {\bm{X}}_L {\bm{W}}_U \in \mathbb{R}^{n\times | {\mathcal{V}} |} .
\label{equa13.3}
\end{equation}
Applying softmax row-wise yields a distribution over the vocabulary at every token position. Thus the training signal is attached to every visible position in the sequence rather than only to the final token.

A crucial point is that autoregressive language modeling is sequential in definition but highly parallel in training. During training, the full token sequence \(u_1,\dots,u_n\) is known, so the model can process all token positions simultaneously, provided that each position is prevented from seeing future tokens. This is achieved by causal masking
  as described earlier in Section \ref{sec:masking}
  which defined the masked attention matrix~\eqref{eq:MaskedAtt}. 
Recall that entries corresponding to forbidden future positions are set
to \(-\infty\) before the row-wise softmax.
Thus the forward pass remains parallel at the matrix level even though the probabilistic model is autoregressive. 
This masked-parallel computation is one of the main reasons the Transformer architecture scales so effectively relative to recurrent models.

The distinction between training and inference is straightforward and important. 
During training, all rows of the hidden-state matrix can be evaluated in parallel under the causal mask. During inference, by contrast, future tokens are not available, so generation proceeds one token at a time. After producing token \(u_t\), the model appends it to the context and recomputes or updates the relevant hidden state needed to predict \(u_{t+1}\).

Thus the model is autoregressive in both training and inference, but only inference is inherently sequential in wall-clock time.
Training uses full sequences with masked parallel computation; inference uses partial sequences with iterative decoding.

Training proceeds by repeated application of the chain rule through the block composition
\begin{equation}
{\bm{X}}_L
=
{\mathcal{T}}_L \circ {\mathcal{T}}_{L-1}\circ \cdots \circ {\mathcal{T}}_1({\bm{X}}_0) .
\label{equa13.6}
\end{equation}

In a single attention stage, the forward computation has the form represented by
Equations~\eqref{equa5.3}, \eqref{eq:Sl}, \eqref{eq:MaskedAtt},
and \eqref{eq:AlDef}.
%
During backpropagation, gradients must therefore flow through the linear projections, the bilinear score formation, the row-wise softmax, and the final value transport. 
The attention stage remains differentiable. It is also more structurally intricate than a standard dense layer.

Residual structure matters here.
In Pre-LayerNorm form, we wrote the block equations
in the form: \eqref{equa7.7} and \eqref{equa7.9}.
%
The skip path contributes an explicit identity map, so gradients keep a direct route across depth. This helps explain why Pre-LayerNorm Transformers train more stably than Post-LayerNorm variants in deep LLM regimes.

The original Transformer paper used the Adam optimizer with
\begin{equation}
\beta_1 = 0.9 , \qquad \beta_2 = 0.98 , \qquad \varepsilon = 10^{-9} ,
\label{equa13.12}
\end{equation}
together with the warmup-plus-decay schedule
\begin{equation}
\mathrm{ l\_rate }
=
d_{\mathrm{model}}^{-1/2}
\min \bigl(
\mathrm{step\_num}^{-1/2},
\mathrm{step\_num}\cdot \mathrm{warmup\_steps}^{-3/2}
\bigr) .
\label{equa13.13}
\end{equation}
%
Under this schedule, we first increase the learning rate linearly during warmup. After warmup, we decrease it in proportion to the inverse square root of the step number.

Modern LLM training pipelines vary in their engineering choices. The overall recipe, however, is still recognizable. We train autoregressively at large scale. We usually use Adam-like optimizers, learning-rate warmup, mixed precision, and data parallelism. We also rely heavily on systems engineering to control memory use and improve throughput.

\subsubsection{Fine-tuning, alignment, and comparison with recurrent training}
For chat-oriented LLMs, we usually do not stop at pretraining.
A common next step is supervised fine-tuning (SFT) on instruction-response data. The objective is still autoregressive. The data changes. We continue optimizing the model on curated instruction-following examples rather than generic web-scale text.
A further stage often introduces preference alignment. 
A standard pipeline begins from a pretrained model, then performs supervised fine-tuning on human demonstrations, trains a reward model from ranked outputs, and finally updates the model using reinforcement learning from human feedback (RLHF)\cite{ouyang2022training, rafailov2023direct}.
These later stages do not replace the Transformer architecture. The block map
\begin{equation}
{\bm{X}}_\ell = {\mathcal{T}}_{\ell}({\bm{X}}_{\ell-1})
\label{equa13.14}
\end{equation}
remains unchanged. 
What changes is the training distribution and, in preference-based stages, the optimization objective used to shape behavior.

It is useful to contrast Transformer training with recurrent training. In a recurrent neural network, the hidden state evolves through time as
\begin{equation}
{\bm{h}}_t = f( {\bm{x}}_t, {\bm{h}}_{t-1} ) , 
\label{equa13.15}
\end{equation}
and training requires backpropagation through time, often over long temporal chains. 
In a Transformer, by contrast, the state variable is the matrix \({\bm{X}}_\ell\), and causal dependence is enforced by masking rather than by recurrent state evolution.

Thus the move from recurrent models to Transformers changed both the architecture and the form of the training computation. Temporal dependence was retained at the probabilistic level, but implemented computationally through masked matrix operations rather than recurrent state evolution. This change is one of the central reasons Transformers are so well aligned with modern parallel hardware.


\subsection{High-dimensional geometry, JL, and intrinsic dimension}
\label{sec:geometry_JL}
The study of Transformers naturally leads to questions of geometry. Hidden states, residual streams, and learned operators all live in high-dimensional spaces, yet empirical evidence repeatedly suggests that their effective behavior is governed by lower-dimensional structure, concentration phenomena, and approximate geometric regularity.

The preceding sections described how Transformers and large language models compute. 
We next ask what geometric structure these models learn.
Four viewpoints will guide the discussion: high-dimensional concentration, approximate orthogonality, learned token-representation geometry, and the intrinsic dimension of the subspaces where the model operates most effectively.
The point is that many learned internal states and operators appear to occupy lower-dimensional or otherwise structured regions of the ambient spaces \(\mathbb{R}^{n\times d_{\mathrm{model}}}\) and \(\mathbb{R}^{d_{\mathrm{model}}}\).

This section introduces the geometric ideas that are most useful for thinking about large language models from a linear algebra viewpoint.

\subsubsection{Johnson–Lindenstrauss, and representation geometry}
Modern large language models operate in very high-dimensional representation spaces. 
For example, GPT-3 uses \(d_{\mathrm{model}}=12{,}288\), whereas Llama-3 405B uses \(d_{\mathrm{model}}=16{,}384\), see Table~\ref{table:JL_Lemma}. 
A natural question is how such spaces can encode many concepts, contextual directions, and partially overlapping features without severe interference.
%
%
\begin{table}[h]
\begin{center}
\begin{tabular}{c|c|c|c|c|c} \hline 
                & & \textbf{GPT-2} & \textbf{GPT-3} & \textbf{GPT-4} & \textbf{Llama-3 405B}
                \\ \hline 
Model Dimension & $d_{\texttt{model}}$ & 768 & \num[group-separator={,}]{12288} & $\num[group-separator={,}]{12288}$ & \num[group-separator={,}]{16384}
\\  \hline 
\end{tabular}
\caption{Representative model dimensions illustrating the scale of modern embedding spaces. The table lists the model dimension $d_{\text{model}}$ for GPT-2, GPT-3, GPT-4, and Llama-3 405B, showing that contemporary large language models operate in ambient spaces of dimension on the order of $10^4$. Such dimensions help explain why many randomly chosen directions are nearly orthogonal, as illustrated in Figure~\ref{fig:jl_lemm}.}
\label{table:JL_Lemma}
\end{center}
\end{table}
%
%

A compelling explanation emerges from the \emph{Johnson–Lindenstrauss (JL) Lemma}\footnote{The core intuition behind superposition, i.e., how language models encode overlapping concepts in high-dimensional spaces, is elaborated in recent pedagogical and technical resources. An accessible introduction is given by Grant Sanderson (minute 17:04 of the 3Blue1Brown video series)~\cite{3Blue1Brown_video_series}, while a more detailed treatment with code and visualizations appears in an online article by N. Yoder, see~\cite{yoder2024beyond}.}
Formally, given a set of $n$ points in high-dimensional space, they can be embedded into a lower-dimensional subspace of dimension $d={\cal{O}} ( \varepsilon^{-2} \log n )$ such that all pairwise distances are preserved within a factor of $(1 \pm \varepsilon)$, see~\cite{johnson1984extensions}, i.e.,
\begin{equation}
(1 - \varepsilon) \| \bm{u} - \bm{v} \|^2 \,\leq\, \| \varphi (\bm{u}) - \varphi (\bm{v}) \|^2 \,\leq\, (1 + \varepsilon) \| \bm{u} - \bm{v} \|^2 ,
\end{equation}
for all pairs of points $\bm{u}, \bm{v}$ in the dataset, and for any random projection $\varphi$, which maps those data points into a space of dimension $d={\cal{O}} ( \varepsilon^{-2} \log n )$, e.g. see~\cite{Chiong2016RandomPE}.
This property ensures that the relative distances between points are approximately preserved, making the technique valuable for algorithms that rely on distance computations.


Here, we do not use the JL lemma mainly as a dimensionality-reduction result. We use it for geometric intuition. In high dimension, a space can contain many directions that are almost orthogonal, or at least weakly interfering. That helps explain why a representation space of dimension $d_{\mathrm{model}}$ can carry many partially distinct features, even though exact orthogonality cannot hold for all of them at once.

A central insight behind the \emph{Johnson–Lindenstrauss (JL) lemma} is that in high-dimensional spaces, random vectors are nearly orthogonal with high probability~\cite{gastpar2024nearly}. 
This \emph{near-orthogonality} is fundamental to why random projections $\varphi$ preserve pairwise distances.
The squared Euclidean norm of any projected vector remains approximately invariant in expectation~\cite{Advani2020RandomPA}, i.e.,
\begin{equation}
{\mathbb{E}} [ \, \| \varphi (\bm{x} ) \|^2 \,] = \| \bm{x}  \|^2 
\quad \text{and} \quad
\texttt{Prob} \Bigl(\, \Bigl| \| \varphi (\bm{x} ) \|^2 - \| \bm{x}  \|^2 \Bigr| > \varepsilon \| \bm{x}  \|^2  \,\Bigr)  \,\leq\, 
2 \exp ( -c \cdot \varepsilon^2 \cdot d ) ,
\end{equation}
for some constant $c > 0$.
This reflects \emph{concentration of measure}\footnote{The \emph{concentration of measure} phenomenon refers to the fact that, in high-dimensional spaces, \emph{Lipschitz} functions of random variables are sharply concentrated around their expectation or median. Formally, for a \emph{1-Lipschitz function} \(f\) and a high-dimensional random vector $\bm{x}$ (e.g., standard Gaussian or uniform on the sphere), we have 
\begin{equation*}
\texttt{Prob}(\, |\, f(\bm{x}) - \mathbb{E}[f(\bm{x})] \,| \geq \varepsilon \,) \leq C \exp( -c \cdot \varepsilon^2 \cdot d ) ,
\end{equation*}
where $C, c > 0$ and $d$ is the dimension. This property underlies results like the Johnson-Lindenstrauss lemma and explains why random projections preserve distances with high probability \cite{ledoux2001concentration, vershynin2018high}.} in high-dimensional probability.
This implies that while the maximum number of \emph{perfectly orthogonal} directions in ${\mathbb{R}}^d$ is $d$, allowing a bounded distortion $\varepsilon$ permits encoding an exponential number of directions, on the order of $\exp ( \varepsilon \cdot d)$. 
Such approximate orthogonality significantly increases representational capacity, enabling models like GPT-3, GPT-4, Llama-3, etc., to embed a vast number of features or concepts into a single fixed-size space.

\begin{example}
To see this geometrically, consider that cosine similarity $\cos \theta$ measures angular closeness, i.e., two vectors at $90\degree$ are orthogonal ($\cos \theta = 0$).
In a 12,288-dimensional space, if we allow up to $\pm 1\degree$ deviation from orthogonality, i.e., vectors separated by angles in $[89\degree, 91\degree]$, the cosine similarity ranges roughly from $\cos 89\degree \approx 0.0174$ to $\cos 91\degree \approx - 0.0174$.
Vectors within this range are nearly orthogonal but can still be distinguished.
Permitting such angular tolerance increases the number of directions that can be accommodated exponentially. Specifically, the number of vectors with pairwise cosine similarity bounded by $| \cos \theta \,| < \delta$ scales as approximately $\exp( c \cdot \delta^2 \cdot d)$, for some constant $c$.
This dramatically expands representational capacity, in GPT-3’s case, even with a modest tolerance (e.g., $\delta = 0.05$), the space can support on the order of billions of distinguishable directions.
%
%
\begin{figure}[tb]
\centerline{
  \includegraphics[width=0.65\textwidth]{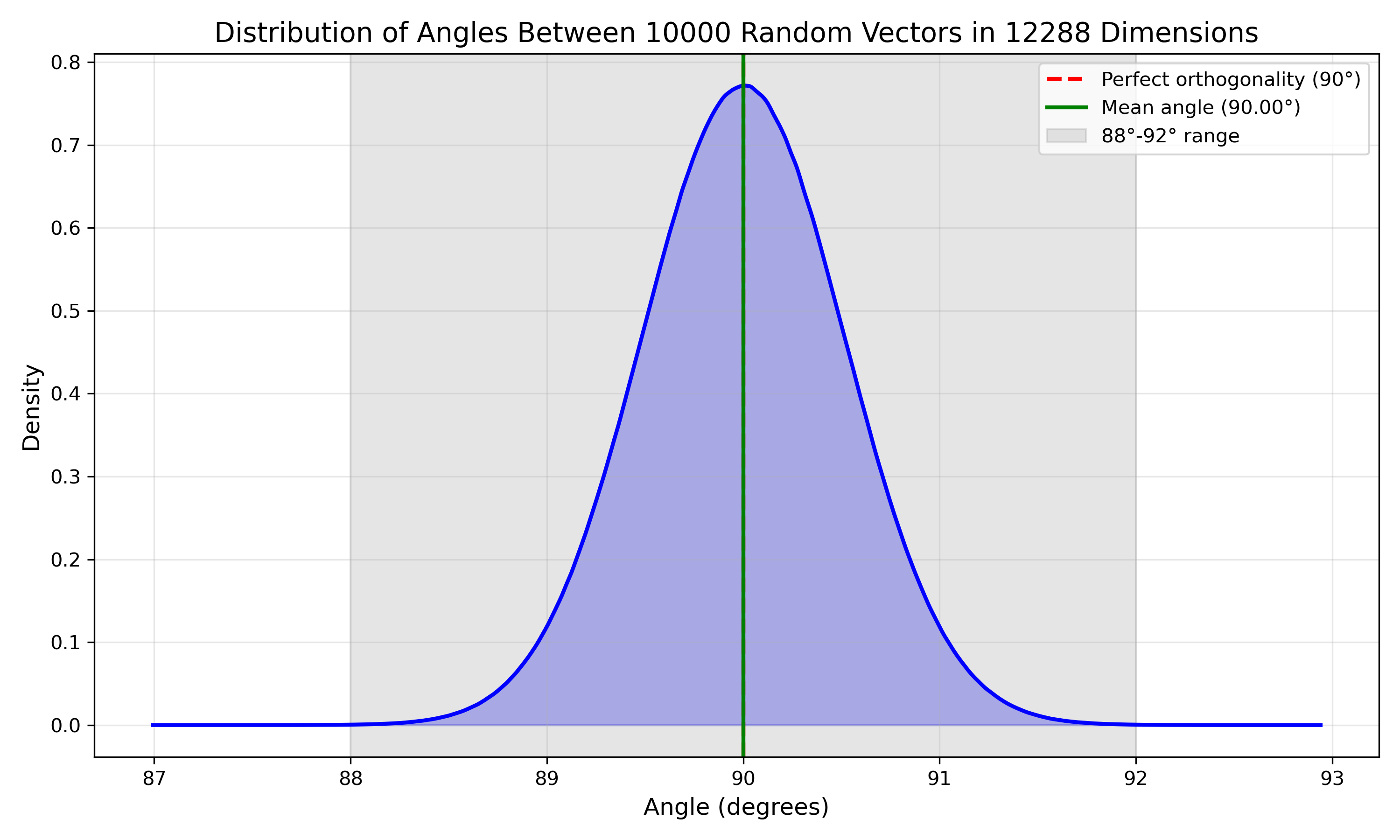}
}
\caption{Concentration of pairwise angles in high dimension. The figure shows the distribution of angles between 10{,}000 random vectors in a 12{,}288-dimensional embedding space. The sharp concentration around 90\degree, together with the narrow 88\degree--92\degree \,band, illustrates the near-orthogonality phenomenon that characterizes random directions in high-dimensional spaces.}
\label{fig:jl_lemm} 
\end{figure}
%
%

This phenomenon is central to understanding how foundation models efficiently compress rich semantic structure. It aligns with the emerging theory of superposition\footnote{Superposition refers to the reuse of embedding directions across different concepts or features, enabling models to represent a combinatorially large number of meanings within limited dimensions. This perspective has been discussed in recent interpretability studies and geometric analyses of embedding spaces; see~\cite{elhage2022toy, nanda2023progress, 3Blue1Brown_video_series, vershynin2018high}.}, where multiple concepts are encoded via overlapping directions in the embedding space, i.e., embedding features in intersecting subspaces rather than along isolated axes.
\end{example}


The JL viewpoint suggests a wider geometric picture. In a high-dimensional model space, we should not expect every semantically meaningful direction to be exactly orthogonal to every other one. That is too restrictive. But many such directions can still be nearly orthogonal, or at least separated enough to avoid strong destructive interference.
This gives a useful way to think about representational packing in large language models. The model does not need one perfectly isolated coordinate axis for each concept. Instead, it can store many directions or subspaces that overlap partially but remain separated enough inside a high-dimensional ambient space. This view will connect naturally to the later discussion of superposition and feature geometry.

\subsubsection{Intrinsic dimension in language models}
The intrinsic dimension (ID) of a model quantifies the minimal number of parameters or latent directions required for a neural network to perform a given task effectively \cite{li2018measuring}. Rather than relying on the nominal parameter count, intrinsic dimension captures the effective degrees of freedom in the learned representation space, offering insight into the model’s geometric and functional complexity \cite{janapati2024measuring, schioppa2024theoretical}. Originally introduced in the context of loss landscape analysis\footnote{Loss landscape analysis examines how the loss function varies across the neural network's parameter space. Key features --such as flat vs. sharp minima and saddle points-- shed light on optimization stability and generalization performance. Flat minima are typically associated with better generalization, while sharp minima may lead to overfitting. See~\cite{draxler2018essentially,li2018visualizing,HaeffeleVidal2017}.}, the concept has since gained traction in analyzing large language models and their fine-tuning behavior \cite{aghajanyan2020intrinsic, jin2024intrinsic}.

In the context of LLMs, intrinsic dimension provides a way to formalize the idea that learned representations and useful updates often lie in much lower-dimensional subspaces than the ambient parameter or embedding dimension suggests. 
Recent empirical studies suggest that modern language models operate effectively within much lower-dimensional subspaces than their total parameter counts would suggest.
For example, Aghajanyan et al.~\cite{aghajanyan2020intrinsic} demonstrated that \texttt{RoBERTa}\footnote{\texttt{RoBERTa} is a LLM that is available in multiple variants. The \texttt{RoBERTa-base} model has approximately 125 million parameters, using 12 transformer layers, a hidden size of 768, and 12 attention heads --mirroring the architecture of \texttt{BERT-base}. The \texttt{RoBERTa-large} model contains 355 million parameters, with 24 layers, a hidden size of 1024, and 16 attention heads. Unlike \texttt{BERT}, \texttt{RoBERTa} is trained with more data, longer sequences, and dynamic masking, yielding stronger performance across many NLP benchmarks, see~\cite{liu2019roberta}.} retains over 90\% of its full performance when optimized in a randomly chosen subspace of just 200 dimensions. Similarly, Subramani et al.~\cite{subramani2020measuring, subramani2022low} showed that controlling LLM outputs could be effectively achieved within subspaces as small as 384 dimensions, highlighting substantial redundancy.

Methods for estimating the \emph{intrinsic dimension} vary, including reparameterization into low-dimensional subspaces with projections back into full model space \cite{aghajanyan2020intrinsic}, and latent vector scaling via linear recoverability criteria\footnote{Linear Recoverability Criteria (LRC) estimate the intrinsic dimension of neural representations by assessing the ability to reconstruct high-dimensional features from low-dimensional projections via linear mappings. If representations are recoverable with minimal error from a small number of dimensions, this suggests that the model's functional capacity is concentrated in a low-dimensional subspace --offering insight into redundancy and efficiency in learned features, see~\cite{subramani2020measuring, subramani2022low}.}.
A standard reconstruction-based formulation is the following. Let
\begin{equation}
{\bm{h}} \in \mathbb{R}^{D}
\label{equa14.4}
\end{equation}
be a hidden representation, and let
\begin{equation}
{\mathcal{E}}_\phi \colon~ \mathbb{R}^{D} \to \mathbb{R}^{d},
\qquad \text{and} \qquad
{\mathcal{D}}_\theta \colon~ \mathbb{R}^{d} \to \mathbb{R}^{D}
\label{equa14.5}
\end{equation}
be an encoder-decoder pair with \(d\ll D\). One may then ask for the smallest \(d\) such that the reconstruction error
\begin{equation}
\min_{\phi, \,\theta}
\mathbb{E}_h \bigl[
\|\, {\mathcal{D}}_\theta (\, {\mathcal{E}}_\phi( {\bm{h}} ) \,) - {\bm{h}}  \,\|_2^2
\bigr]
\label{equa14.6}
\end{equation}
remains below a prescribed threshold corresponding to high fidelity (e.g., explaining 90\% of the variance or recovering 90\% of downstream task accuracy).
This gives a concrete operational meaning to intrinsic dimension: it is the dimension of a compressed latent space that still captures the essential information carried by the representation.
Subramani et al.~\cite{subramani2020measuring, subramani2022low} show that embeddings and model steering can be well approximated in subspaces of a few hundred dimensions (e.g., 384-768), highlighting the potential for parameter-efficient fine-tuning methods such as LoRa \cite{hu2021lora}, which optimize low-rank updates aligned with these intrinsic dimensions.

Janapati et al.~\cite{janapati2024twoNN} employed the Two Nearest Neighbors (TwoNN) estimator, see~\cite{facco2017estimating}, for its robustness and efficiency in measuring the \emph{intrinsic dimension} across neural representations, while Lee et al.~\cite{lee2025intrinsic} found that token embeddings often lie on low-ID manifolds, with low-ID tokens forming semantically coherent clusters. 
These findings align with the manifold hypothesis\footnote{The \emph{manifold hypothesis} asserts that high-dimensional data arising in real-world applications concentrate near low-dimensional manifolds embedded within the ambient input space \cite{fefferman2016testing, coifman2006diffusion}. This suggests that the effective dimensionality of data is significantly lower than its ambient dimension, enabling models to exploit geometric structure for efficient representation learning, improved generalization, and robustness. The \emph{manifold hypothesis} underpins many nonlinear dimensionality reduction techniques and motivates architectures designed to capture intrinsic data geometry, thus addressing the curse of dimensionality in large-scale learning.}, see~\cite{janapati2024twoNN}, which suggests that real-world data resides on low-dimensional manifolds, thereby constraining the space neural networks must explore.
Kataiwa et al.~\cite{kataiwa2025intrinsic} further quantified this redundancy, showing that LLM token embeddings exhibit redundancy ratios exceeding 98\%, reinforcing the idea that most of the parameter space is underutilized during learning. Such redundancy provides a theoretical basis for the empirical success of compression techniques, including pruning \cite{frankle2019lottery, gale2019sparsity}, distillation, and parameter-efficient fine-tuning approaches such as LoRa \cite{hu2021lora}. 
Notably, Jin et al.~\cite{jin2024derivativefreeoptimizationlowrankadaptation, jin2023_instance_aware} linked LoRa’s effectiveness to its alignment with intrinsic subspaces, allowing for robust adaptation with minimal parameter updates.

\begin{example}[Intrinsic Dimension in Fine-Tuning RoBERTa]
Consider a pre-trained large language model with parameter vector $\bm{w} \in \mathbb{R}^D$, where $D$ is very large (e.g., $D \approx 3.55 \times 10^8$ for \texttt{RoBERTa-large}).
Fine-tuning typically updates $\bm{w}$ to ${\bm{w}}' = \bm{w} + \Delta \bm{w}$ to minimize a task-specific loss ${\cal{L}} ( {\bm{w}}' )$.
\emph{Intrinsic dimension} assumes that $\Delta \bm{w}$ lies approximately in a low-dimensional subspace of dimension $d \ll D$, such that $\Delta {\bm{w}} = {\bm{U}}^T {\bm{z}}$, where ${\bm{U}} \in {\mathbb{R}}^{d \times D}$ is a fixed random projection matrix with orthonormal columns, and $\bm{z} \in {\mathbb{R}}^d$ are the parameters optimized during fine-tuning.

Empirically, Aghajanyan et al.~\cite{aghajanyan2020intrinsic} found that for \texttt{RoBERTa}, optimizing $\bm{z}$ with $d \approx 200$ recovers over 90\% of the full fine-tuning performance, despite \(d \ll D\). Formally,
\begin{equation*}
\mathbb{E}_{ \bm{x}, \bm{y} } \left[ \ell \bigl( f_{\Theta} ( \bm{x} \mid {\bm{w}} + {\bm{U}}^T {\bm{z}}^* ) , \bm{y} \bigr) \right] 
\approx 
0.9 \times \mathbb{E}_{ \bm{x},\bm{y} } \left[ \ell \bigl( f_{\Theta} ( \bm{x} \mid {\bm{w}} + \Delta {\bm{w}}^* ), \bm{y} \bigr) \right] ,    
\end{equation*}
where $\ell$ is the loss function, $f_{\Theta}$ is the model, and ${\bm{z}}^*$, $\Delta {\bm{w}}^*$ are the optimal updates in the subspace and full space respectively.
\end{example}

%
\begin{example}[Linear Recoverability and Intrinsic Dimension in LLMs]
Let $\bm{h} \in \mathbb{R}^D$ be the token embedding or hidden representation from a large language model, such as \texttt{BERT} or \texttt{RoBERTa}, e.g., $D = 768$ in \texttt{RoBERTa-base}.
Assume there exists an encoder ${\cal{E}}_{\varphi} \colon \mathbb{R}^D \rightarrow \mathbb{R}^d$ and a decoder ${\cal{D}}_{\theta} \colon \mathbb{R}^d \rightarrow \mathbb{R}^D$, such that:
\begin{equation*}
{\bm{h}}' = {\cal{D}}_{\theta} ( {\cal{E}}_{\varphi} ( \bm{h} ) ) 
=
{\bm{W}}_{\texttt{dec}}^T ( {\bm{W}}_{\texttt{enc}}^T {\bm{h}} ) ,    
\end{equation*}
where ${\bm{W}}_{\texttt{enc}} \in \mathbb{R}^{D \times d}$ and ${\bm{W}}_{\texttt{dec}} \in \mathbb{R}^{d \times D}$ are linear projection matrices.
To evaluate recoverability, we measure whether the reconstruction ${\bm{h}}'$ preserves key semantic information, typically using a classification or retrieval task.
For example, Subramani et al.~\cite{subramani2022low} define \emph{reconstruction accuracy} as:
\begin{equation*}
{\cal{L}}_{\texttt{recon}}
= 
\frac{1}{N} \sum\limits_{i=1}^{N} 
\| {\bm{h}_i} \,-\,  {\bm{W}}_{\texttt{dec}}^T ( {\bm{W}}_{\texttt{enc}}^T {\bm{h}_i} ) \|_2^2 ,
\end{equation*}
which checks if the predicted class (or nearest token) from the reconstructed embedding matches that of the original.
We define \emph{linear recoverability} as the smallest $d$ such that ${\cal{L}}_{\texttt{recon}} \leq \varepsilon \cdot {\cal{L}}_{\texttt{base}}$, where ${\cal{L}}_{\texttt{base}}$ is the loss with no dimensionality reduction (i.e., identity mapping), and $\varepsilon \in [0, 1]$ is a small threshold (e.g., 0.1).
If the model achieves $\texttt{Accuracy} \geq 90\%$ for a relatively small $d \ll D$, then we infer that the \emph{intrinsic dimension} of the representation $\bm{h}$ is approximately $d$. 
 For instance, Subramani et al. show that \texttt{BERT} sentence embeddings can be linearly recovered from $d=384$ dimensions (out of 768) with over 90\% accuracy, implying that \emph{intrinsic dimension} $\approx 384 \ll 768 = D$.
This suggests that the high-dimensional embedding lies on or near a low-dimensional manifold, validating the \emph{manifold hypothesis} and motivating techniques such as low-rank adaptation in fine-tuning.
\end{example}

%
\begin{example}[Singular Value Interpretation]
In~\cite{aghajanyan2020intrinsic},
although they did not explicitly compute SVD, this result implies that the gradient updates during fine-tuning largely lie in a low-rank subspace.
If we were to compute the SVD of the gradient Jacobian, we would likely find that the top 200 singular values account for $> 90\%$ of the total gradient energy, i.e.:
\begin{equation*}
    \frac{ \sum\limits_{i=1}^{200} \sigma_i^2 }{ \sum\limits_{i=1}^r \sigma_i^2} \ge 0.90 ,
\end{equation*}
where $\sigma_1 \ge \sigma_2 \ge \cdots \ge \sigma_r$ are the singular values of the gradient matrix.
\end{example}

\begin{example}[Gradient Subspace Analysis]
Let us consider a typical transformer-based model like \texttt{BERT-base} ($110\times 10^6$ parameters). 
Suppose we collect gradient vectors from 100 mini-batches during training.
We stack gradients as columns in a matrix $\bm{G} \in {\mathbb{R}}^{D\times T}$, where $D = 110\times 10^6$, and $T = 100$.
It is not feasible to store $\bm{G}$ directly, so let us analyze a \emph{compressed sketch} of $\bm{G}$, say from a small subnetwork or with randomized projections, by first computing its singular value decomposition, i.e, $\bm{G} = {\bm{U}} \Sigma {\bm{V}}^T$, where $\Sigma = \diag( \sigma_1, \sigma_2, \cdots, \sigma_T)$.
\begin{table}[htb]
    \centering
    \begin{tabular}{|c|c|c|} \hline
     Component $i$ & $\sigma_i$ (Energy) & Cumulative \\ \hline
         1 & 0.40 & 0.40 \\ \hdashline
         2 & 0.20 & 0.60 \\ \hdashline
         3 & 0.10 & 0.70 \\ \hdashline
         4 & 0.08 & 0.78 \\ \hdashline
         5 & 0.06 & 0.84 \\ \hdashline
      6-10 & 0.03 (each) & 0.99 \\ \hdashline
      11-100 & < 0.01 & $\sim$ 1.00 \\ \hline 
    \end{tabular}
    \caption{Illustrative low-rank concentration of gradient energy. The table shows a normalized singular-value distribution $\sum_i \sigma_i =1$, for which the cumulative energy exceeds 90\% within roughly 10 components. Despite the ambient parameter dimension $D = 110\times 10^6$, most of the gradient energy is concentrated in a very low-dimensional subspace, illustrating the motivation behind low-rank gradient methods.}
    \label{tab:SVD_Gradient}
\end{table}

%
The gradient matrix is, therefore  numerically low-rank.
Empirical studies suggest that gradients in large language models often reside in low-rank subspaces, with effective ranks typically ranging between 100 and 1,000 even for models with billions of parameters. 
Understanding and leveraging the low-rank structure of gradients is crucial for developing efficient training strategies for LLMs, balancing resource constraints with performance requirements, see \cite{zhao2024galore_icml}.
\end{example}

\paragraph{Geometric Evolution of Token Representations in LLMs.}
Recent work has identified a characteristic \emph{expansion-compression} pattern in the \emph{intrinsic dimension} of token representations across the layers of large language models.The \emph{intrinsic dimension} increases in the early layers, reaches a maximum in intermediate layers, and then gradually declines toward the output. 
This trend has been observed across diverse architectures and tasks, e.g., see~\cite{Cheng2024EmergenceOA, Valeriani2023TheGO, yin2024evolution}.

The increasing \emph{intrinsic dimension} in early layers reflects the expansion of input tokens into a high-dimensional working space, while the subsequent decline indicates compression onto structured, low-dimensional manifolds as representations become more abstract.
This \emph{expansion-contraction} pattern supports the \emph{diffusion-projection hypothesis} \cite{Song2025BridgingTD}, wherein token embeddings initially diffuse into a broad latent space before being compressed onto task-relevant submanifolds.
Balestriero et al.~\cite{balestriero2024characterizinglargelanguagemodel} provide a formal geometric analysis of multi-head attention, deriving closed-form expressions for the \emph{intrinsic dimension} of its outputs. They show that attention constrains representations to lie on manifolds whose dimensionality depends on factors such as the number of heads, attention rank, and input diversity.



\paragraph{Low-dimensional structure and adaptation.}
The observation that large language models operate effectively within low-dimensional subspaces has significant implications for scalability and fine-tuning.
If useful updates lie in a low-rank subspace, then one may represent the update to a weight matrix \(\bm{W}\) in factored form
\begin{equation}
\Delta {\bm{W}} = {\bm{B}} {\bm{A}},
\label{equa14.7}
\end{equation}
where \({\bm{B}}\) and \({\bm{A}}\) have much smaller inner dimension than \({\bm{W}}\) itself.
This replaces full-rank adaptation by optimization in a restricted subspace, matching the empirical observation that large ambient parameter spaces often have much smaller effective task-relevant dimension. In this sense, intrinsic-dimension arguments and low-rank adaptation are closely connected.

\medskip
In conclusion, high-dimensional geometry helps explain why large language models can support many approximately distinct feature directions in spaces of dimension $d_{\text{model}}$.
The Johnson–Lindenstrauss lemma provides a geometric lens on approximate distance preservation and weak interference among many directions. 
Intrinsic dimension makes precise the idea that useful representations and updates often occupy much smaller subspaces than the ambient dimension suggests. 
The geometry of token representations also appears to evolve systematically across depth, reflecting the repeated action of attention, MLPs, and residual accumulation.



\subsection{Interpretability through linear algebra}
A linear algebra view of interpretability begins from the observation that several of the Transformer’s most important internal objects are explicit matrices or linear maps. 
Attention matrices describe token-token transport, residual-stream updates admit additive decomposition, and unembedding projects hidden states into a fixed vocabulary coordinate system. 
This section develops these three viewpoints as complementary ways of probing the model’s internal computation.
\subsubsection{Attention, residual streams, and logit-space probes}
The previous section argued that Transformers and large language models exhibit spectral, rank, and geometric structure. A natural next question is whether these same structures can be used to interpret what the model is doing internally. In the Transformer setting, this question is especially natural because many of the main internal objects are already matrix-valued or vector-valued: attention matrices, residual-stream states, MLP activations, and vocabulary logits.

Accordingly, a linear algebra approach to interpretability seeks to understand the model in terms of operators and representations rather than only through input-output behavior.
The goal is not to claim that the model is globally linear, but to exploit the fact that many important internal steps are linear or piecewise linear once the relevant gates or attention patterns are fixed. 
This viewpoint is central to the transformer circuits program, which studies Transformers by decomposing them into interpretable linear and multilinear pieces, especially through the residual stream and attention heads\cite{elhage2021mathematical}.

One of the most visible internal objects of a Transformer is the attention matrix
$\bm{P}_\ell$ defined in Equation \eqref{eq:MaskedAtt}.
%
%
Since each row sums to one, \({\bm{P}}_\ell\) may be viewed as a row-stochastic token-token operator, and the attention output is $\bm{A}_\ell(\bm{X}_{\ell-1})$ as
given by Equation~\eqref{eq:AlDef}.
%
Thus the attention matrix explicitly describes how value vectors are mixed across token positions. 
For this reason, attention maps are often the first objects inspected when trying to interpret a model.

However, one must be careful. Standard attention weights do not automatically provide faithful explanations of model predictions and should not simply be treated as explanations in themselves. 
Thus attention matrices are best understood as explicit internal transport operators, not as guaranteed explanations of output behavior.
Hence attention is interpretable in one precise sense: it gives a visible matrix describing which token positions are being mixed. But that does not mean it fully explains why the model ultimately predicts what it predicts.

\begin{figure}[tb]
\centerline{ 
\includegraphics[width=1.0\textwidth]{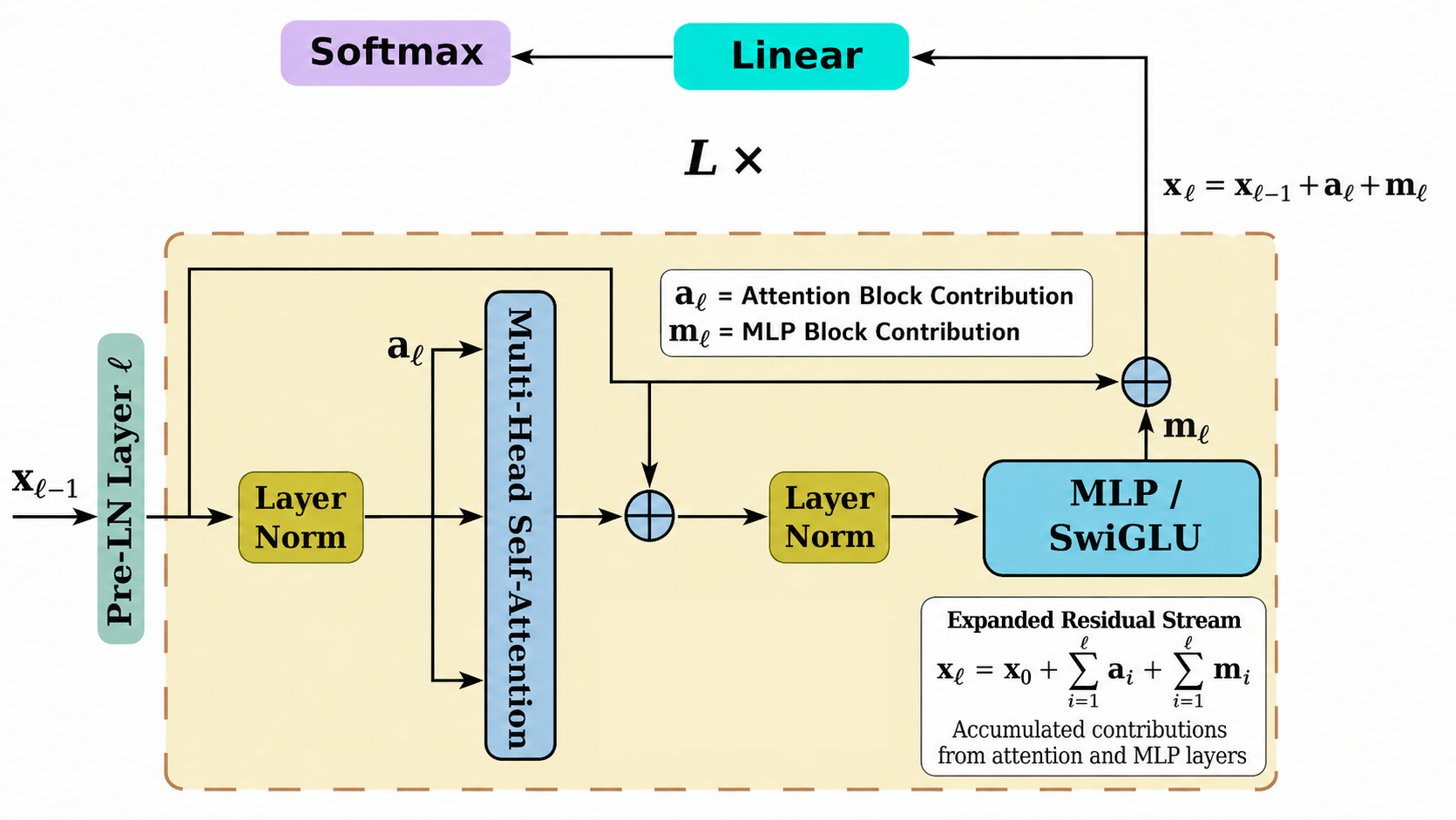}}
\caption{Pre-LN Transformer layer as residual-stream accumulation. The input residual stream ${\bm{x}}_{\ell-1}$ is updated by adding the attention-block contribution ${\bm{a}}_{\ell}$ and the MLP-block contribution ${\bm{m}}_{\ell}$, giving ${\bm{x}}_\ell = {\bm{x}}_{\ell-1} + {\bm{a}}_\ell + {\bm{m}}_\ell$. Across layers, the residual stream can be expanded as ${\bm{x}}_\ell = {\bm{x}}_0 + \sum_{i=1}^{\ell} {\bm{a}}_i + \sum_{i=1}^{\ell} {\bm{m}}_i$, showing how attention and MLP updates accumulate additively through the network.}
\label{fig:residual_stream}
\end{figure}

A more structurally faithful object for interpretation is the residual stream. As discussed earlier, in Pre-LayerNorm form one may write the layerwise updates schematically as

\begin{equation}
{\bm{x}}_\ell
=
{\bm{x}}_{\ell-1} + {\bm{a}}_\ell + {\bm{m}}_\ell ,
\label{equa15.3}
\end{equation}
or, expanded across depth,
\begin{equation}
{\bm{x}}_\ell
=
{\bm{x}}_0 + \sum_{i=1}^{\ell} {\bm{a}}_i + \sum_{i=1}^{\ell} {\bm{m}}_i .
\label{equa15.4}
\end{equation}
These equations already suggest a decomposition of the hidden state into accumulated contributions from attention heads and MLP blocks. 
This residual-stream accumulation viewpoint is illustrated in Figure~\ref{fig:residual_stream}.
This is one of the central ideas in the transformer-circuits framework: attention heads move information between positions through the residual stream, while later components can read and transform the resulting vectors.
%

From a linear algebra point of view, the residual stream is especially attractive because it is the common state space into which many different submodules write. This means that one can attempt to decompose a prediction or an intermediate representation into additive contributions from distinct components. 
Even when the resulting decomposition is not uniquely meaningful, it often provides a more faithful picture than attention weights alone because it reflects the actual data path through which information is stored and reused.

A second important interpretability idea is to examine hidden states through the vocabulary projection.
If ${\bm{W}}_U \in \mathbb{R}^{d_{\mathrm{model}}\times |{\mathcal{V}}|}$
is the unembedding matrix, then the final logits at layer \(L\) are ${\bm{Z}}_L = {\bm{X}}_L {\bm{W}}_U$.
A natural question is what happens if one applies this same unembedding map to earlier hidden states:
\begin{equation}
{\bm{Z}}_\ell = {\bm{X}}_\ell {\bm{W}}_U,
\qquad
\ell < L.
\label{equa15.7}
\end{equation}
This is the basic idea behind the logit lens: intermediate residual-stream states are projected directly into vocabulary space in order to see what the model ``already seems to believe'' at intermediate depths \citep[][]{nostalgebraist2020logitlens, belrose2023eliciting}.

A more refined version is the tuned lens, which inserts a learned affine map before the unembedding in order to better align intermediate states with the final output geometry. In the language of linear algebra, both methods interpret the evolving residual stream by projecting it into a fixed output coordinate system, namely the vocabulary basis induced by \({\bm{W}}_U\). This gives a coherent way to compare intermediate representations across layers.

\subsubsection{Head specialization, feature directions, and the limits of linear interpretability}
Multi-head attention naturally raises the question of whether different heads and layers serve different functions. If head \(h\) in layer \(\ell\) produces
\begin{equation}
\operatorname{Head}_h^\ell
=
{\bm{P}}_h^\ell {\bm{V}}_h^\ell,
\label{equa15.8}
\end{equation}
then one may ask whether certain heads implement recognizable operations, such as copying, induction, or delimiter matching.
This line of inquiry is central to the transformer circuits literature. In particular, induction-head analyses show that some heads can implement a copy-and-continue mechanism that supports in-context pattern continuation, and this can sometimes be derived explicitly using a mechanistic decomposition of the model’s linear pathways \cite{olsson2022induction}.
At the same time, specialization is not absolute. Some heads appear highly interpretable, while others seem redundant or distributed in function. Thus one should think of head and layer specialization not as a clean symbolic partition of labor, but as a structured and only partially disentangled decomposition of the model’s computation.

A deeper interpretability question concerns how concepts or features are represented in the residual stream. A naive view would assign one direction to each concept. But the high-dimensional geometry discussed in Section~\ref{sec:geometry_JL} suggests a more complicated picture: many features may be represented in overlapping directions rather than in a one-feature-per-axis basis.
This is the basic intuition behind superposition. The interpretability literature often studies the model as if meaningful features live in directions or subspaces of the residual stream, but there is no guarantee that these directions are mutually orthogonal or uniquely identifiable. Instead, the model may reuse coordinates and pack many partially overlapping features into the same ambient space.
This means that interpretability often becomes a problem of finding useful subspaces, basis changes, sparse decompositions, or directions with semantic stability, rather than simply identifying one neuron or one axis per concept \citep[][]{elhage2022superposition}.

The success of linear algebra methods should not be overstated. Attention maps are explicit operators, but they are not always faithful explanations. Residual decompositions are algebraically natural, but their semantic interpretation may be ambiguous. Lens methods project intermediate states into vocabulary space, but this projection may distort what the model is actually representing internally, especially in earlier layers.

The correct claim is not that linear algebra solves interpretability, but that it provides a disciplined language in which many internal computations can be expressed and compared. 
It gives us matrices, projections, subspaces, and additive decompositions that are often far more informative than black-box input-output analysis alone, while still leaving open the deeper question of which internal structures are causally or semantically meaningful.

\medskip
Interpretability through linear algebra proceeds along several related paths. Attention matrices may be studied as explicit token-token transport operators, though not as explanations by default. Residual-stream equations support additive decomposition of hidden states into attention and MLP contributions. Unembedding through the vocabulary matrix provides a common coordinate system for reading intermediate states, as in the logit lens and tuned lens. 
Headwise and layerwise analysis can reveal partial specialization, while feature directions and superposition suggest that concepts may occupy overlapping subspaces rather than isolated coordinates.

\subsection{Scaling laws and emergent phenomena}
Scaling laws provide one of the clearest indications that the behavior of large language models is not arbitrary, but follows regular trends as model size, data size, and compute budget increase. These empirical laws suggest that performance often improves in a predictable manner over broad regimes, while also raising deeper questions about which aspects of this behavior can be explained mathematically. At the same time, the rapid growth of model scale has brought attention to so-called emergent phenomena, namely qualitative changes in capability that appear only beyond certain ranges of scale. It is therefore natural to examine scaling laws and emergent behavior together, both as empirical facts about modern LLMs and as challenges for a more principled mathematical understanding.
\subsubsection{Empirical scaling laws and compute-optimal training}
The study of scaling laws in machine learning has roots in early analyses of generalization and network complexity, and gained renewed prominence with the advent of large-scale neural networks.
A key empirical finding demonstrated that language model performance follows a predictable power-law relationship with model size \(N\), data size \(D\), and compute budget \(C\). The general form of these laws is
\begin{equation}
L(N,D)
\approx
A N^{-\alpha} + B D^{-\beta} + E,
\label{equa17.1}
\end{equation}
where \(L\) denotes test loss, \(A\) and \(B\) are constants, \(\alpha,\beta>0\) are scaling exponents indicating diminishing returns,
and $E$ is a constant representing the \textit{irreducible loss}.

The seminal work of Kaplan et al.\ revealed that language model performance obeys predictable power-law scaling with respect to model size, dataset size, and compute. This empirical regularity provides a quantitative framework for extrapolating how performance improves as models, datasets, and training budgets increase \cite{kaplan2020scaling_laws}.
Scaling laws are important because they reveal a remarkable degree of regularity in systems that are otherwise extremely high-dimensional and nonlinear.
Even though Transformers are built from complicated compositions of attention, normalization, gating, and residual updates, their aggregate training behavior often obeys simple power-law trends over many orders of magnitude.
%
\begin{figure}[tb]
\centerline{
\includegraphics[width=1.0\textwidth]{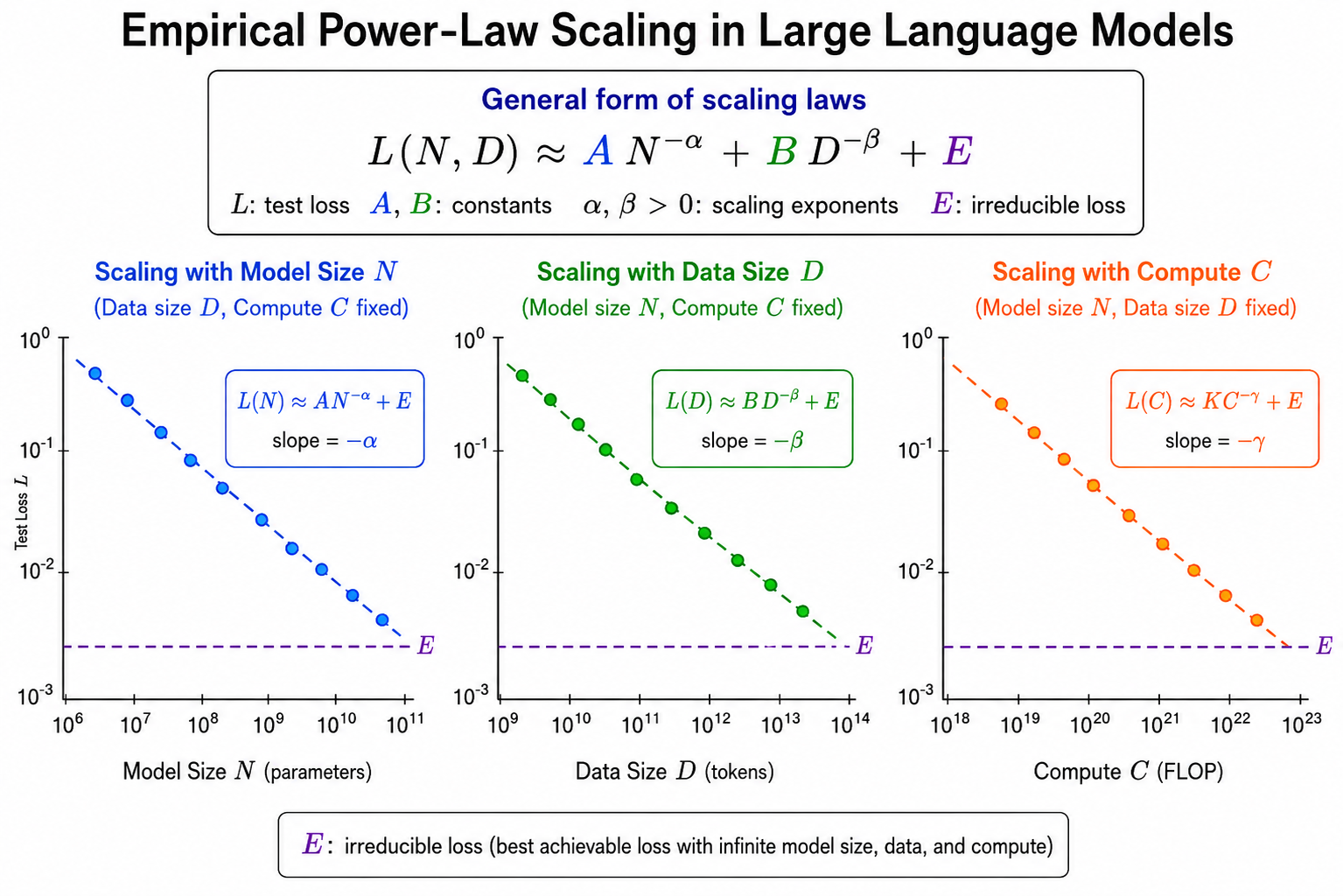}}
\caption{Empirical power-law scaling in large language models. Test loss decreases $L (N, D)$ predictably as model size $N$, dataset size $D$, and compute budget $C$ increase, following approximate power-law relationships. The schematic summarizes the scaling-law perspective that performance improves smoothly with scale, with diminishing returns captured by
  the exponents  in $L(N,D)$ defined in \eqref{equa17.1}} 
\label{fig:scaling_law}
\end{figure}
Subsequent work refined the original scaling paradigm by identifying trade-offs between model size and dataset size. Hoffmann et al.\ introduced compute-optimal scaling, showing that models trained with balanced allocations of parameters and tokens outperform those optimized along a single axis \cite{hoffmann2022compute}. 
This led to the Chinchilla scaling laws, which emphasize the importance of jointly scaling model size and data size under a fixed compute budget.
In more concrete terms, the lesson is that making a model larger is not enough by itself. If the dataset size is not increased appropriately, one may end up with a model that is undertrained relative to its parameter count. This observation has had a substantial practical impact on how large language models are designed and trained.
These developments provide concrete guidance for resource allocation in large-scale model development.
They replace crude trial-and-error scaling with a more disciplined picture in which parameters, data, and compute must be co-designed.

These results support the scaling hypothesis, namely that performance can be reliably improved by increasing model and data size even without fundamental architectural changes.
This principle has informed the design of modern foundation models, with GPT-3 serving as a canonical example of the capabilities unlocked at sufficient scale.
This point is historically important. GPT-3 did not introduce a radically new architecture relative to GPT-2; rather, it demonstrated that sufficiently scaling a decoder-only Transformer could produce qualitatively richer task behavior.
The scaling hypothesis is also a linear algebra challenge. It suggests that increasing depth, width, dataset coverage, and optimization time does not merely enlarge the model quantitatively, but may alter the geometry and operator structure of the learned representation space in qualitatively important ways.

\subsubsection{Emergent behavior, mathematical questions, and cautious interpretation}
Emergent abilities challenge the continuity assumed by traditional scaling laws. Whereas scaling laws predict smooth, gradual improvements in performance as model size increases, emergent abilities represent abrupt, nonlinear gains in capability that appear only after crossing specific thresholds in model size or training complexity. These behaviors are absent in smaller models and cannot be anticipated by simply extrapolating from their performance. 
Their unpredictability and qualitative novelty mark them as a distinct regime in the scaling behavior of large language models.

The phenomenon of emergent abilities gained significant traction with the release of GPT-3, which introduced in-context learning, i.e., the capacity to solve novel tasks solely through prompt-based conditioning, without gradient updates to model parameters \cite{brown2020gpt3}. 
This capability was not observed in earlier models such as GPT-1 or GPT-2 in the same way, pointing to the existence of threshold effects in which qualitatively new behaviors emerge once a model surpasses a critical scale.
Whether one interprets these effects as genuine phase transitions, evaluation artifacts, or some mixture of both, the main point remains: scale can change not only accuracy values but the apparent qualitative repertoire of the model \cite{wei2022emergent}.

One of the most interesting unresolved questions is what exactly changes mathematically as scale increases. At the architectural level, the equations of the Transformer block remain the same:
\begin{equation}
{\bm{X}}_\ell
=
{\mathcal{T}}_{\ell}({\bm{X}}_{\ell-1}) ,
\label{equa17.2}
\end{equation}
and the full model remains a composition
\begin{equation}
{\bm{X}}_L
=
{\mathcal{T}}_L \circ {\mathcal{T}}_{L-1}\circ \cdots \circ {\mathcal{T}}_1({\bm{X}}_0) .
\label{equa17.3}
\end{equation}
Yet the empirical behavior of the model changes dramatically as the number of layers, hidden dimension, training tokens, and optimization steps grow.
Several possibilities suggest themselves. The effective rank of learned operators may increase; the residual stream may support a richer family of approximately separable feature directions; optimization may discover more stable algorithmic circuits; and high-dimensional geometric effects may become more pronounced. But at present, these remain hypotheses rather than a settled mathematical theory of scale.
This suggests that scale concerns not only larger models, but also changes in the geometry, redundancy, and expressivity of the learned representation space.

Scaling behavior has also been observed beyond unimodal language models, including contrastive language-image pretraining and multimodal generative models. 
This suggests that scaling is not a peculiarity of text-only autoregressive models, but a more general empirical phenomenon associated with large neural operator systems trained on rich data distributions.

This widens the significance of the scaling discussion. The question is not only why decoder-only language models scale well, but also why structured deep operator systems built from learned projections and nonlinear mixing seem to obey such regularities across modalities.

At the same time, one should be careful not to overstate what scaling laws currently explain. They describe empirical regularities in loss or task performance, but they do not yet provide a deep mechanistic account of why those regularities arise from the internal operator structure of the model. Nor do they fully explain why some capabilities appear to improve smoothly while others seem to emerge abruptly.

The right conclusion is twofold. First, scaling laws are one of the strongest empirical regularities in modern deep learning. Second, they remain only partially understood from the perspective of linear algebra, operator theory, or representation geometry.

\medskip
Scaling laws show that language-model performance often follows predictable power-law trends with respect to model size, dataset size, and compute. Compute-optimal scaling refines this picture by emphasizing the joint scaling of model and data. At the same time, large models exhibit threshold-like or emergent behaviors that are not fully captured by a naive smooth-scaling interpretation. 
GPT-3 is a canonical example of this transition from scale as quantitative improvement to scale as qualitative change in behavior.

The key point is that scaling is not only an engineering story. It is also a mathematical story about how the behavior of a deep structured operator system changes as dimensionality, training data, and compute budget increase.


\section{Exploiting Spectral Characteristics of Transformers}
\label{sec:spectral} 
Applications of numerical linear algebra techniques 
can be found virtually everywhere in machine learning, from optimizing
matrix operations to better designs of models through compression.
In this section we briefly describe a few examples of instances where
clever uses of NLA played a major role in improving the performance of
deep learning models. It may be argued that the single most important
NLA tool exploited in deep learning consists of reducing the number of
variables through model compression or low-rank approximation techniques.

\subsection{Model Compression} 

In deep learning, compression refers to a class of techniques employed to reduce
the size of deep neural networks while maintaining or minimally affecting their
performance. Compression is especially helpful when deploying models on
resource-constrained devices (e.g., mobile phones, IoT devices). More generally
it aims at reducing memory requirements and improving computational efficiency.
Its effectiveness is due in part to the fact that modern deep networks are often
strongly overparameterized.  After training, many directions in parameter space
appear to have only a limited effect on the loss, a phenomenon often described
through low-curvature directions of the Hessian.  This helps explain why
pruning, quantization, distillation, and low-rank approximation can reduce
storage and computation with only limited loss in accuracy.  In this sense,
model compression may be viewed as an algorithmic expression of the principle of
parsimony.  Common approaches to model compression include pruning,
quantization, knowledge distillation, and low-rank approximation. Despite their
differences, all aim to reduce storage and computation while preserving accuracy
as much as possible.

\subsubsection{Pruning} 
This technique amounts to sparsifying the links
between layers in neural networks. Pruning consists of removing less important
weights or neurons to make the network sparse and reduce storage as well as
computational costs.  The idea suggested in \citep{han2015Pruning} consists of three
steps.  In the first step the connectivity of the network is learned by seeking
to determine which connections are important. In a second step, the weights are
`pruned' by removing connections between layers when they are deemed
unnecessary, essentially sparsifying the network. The third and final stage
trains the network to determine the weights of the remaining connections.

\subsubsection{Quantization.}
Quantization exploits hardware to enable compression. It consists of reducing
the numerical precision used to represent weights, activations, or both, in order to lower memory requirements and improve computational efficiency (e.g., from 32-bit floating point to 8-bit or lower),   
First, parameters are quantized, i.e., real values are mapped
into shifted and scaled integer representations. The scheme developed in
\citep{quantization18} uses integer-only arithmetic during inference and
floating-point arithmetic during training.  It uses an affine mapping from
integers to real numbers whereby a real number $r$ is represented as
$r = S(q-Z)$ where the scaling $S$ (real) and 
the `zero-point' or shift $Z$ (integer) are parameters. These parameters are the same within each activation array and for each weight array. Bias vectors may have longer integer representation.  The point is that the `quantized' value $q$ can be represented as an 8-bit integer - leading to big savings see \citep{quantization18}.  
With the quantization scheme just described, \citep{quantization18} describes how
inference, which relies mostly on matrix-matrix multiplication, can be carried
out using only (fixed-point) integer arithmetic.  Such methods are mostly
deployed for the inference phase and they target application on mobile devices
where memory and compute time need to be minimized.

\subsubsection{Knowledge Distillation}
This is a methodology whereby a small network called ``student network'' is
trained to mimic a larger network called the ``teacher network''.  The basic
idea was first described by Bucil\u{a} et al.  \citep{Bucila06} as a model
compression technique to transfer information from an ensemble of large models
to train a smaller model while maintaining a good accuracy.  Later Hinton et al. 
\citep{HintonAlDistill15} introduced the term `knowledge distillation' as a
specific model compression technique in deep learning.

One of the key arguments in \citep{HintonAlDistill15} is that the student model
will reproduce the performance of the teacher model when measured by its
capacity to generalize to unseen data.  It is hypothesized that the 
logits output by the teacher model contain much information besides
the choice of classification made from it.  Here a logit is the output of
the last layer before it is fed to the softmax function to provide `soft
targets'.
Following the notation of Section~\ref{sec:trainMLP} we will call
$z_{i}$ or $z_{i,:}$ one such logit, a (row) vector of
length $C$. Then  the class
probabilities produced by the network are computed by the softmax function:
\eq{eq:piqi1} p_i = \frac{ \exp(z_{i,:}/T) }{\sum_j \exp(z_{i,j}/T) } ,
\en
where by default the temperature $T$ is set to 1, see \eqref{eq:softmax}. 

\begin{figure}[tb]
  \centerline{\includegraphics[width=0.7\textwidth]{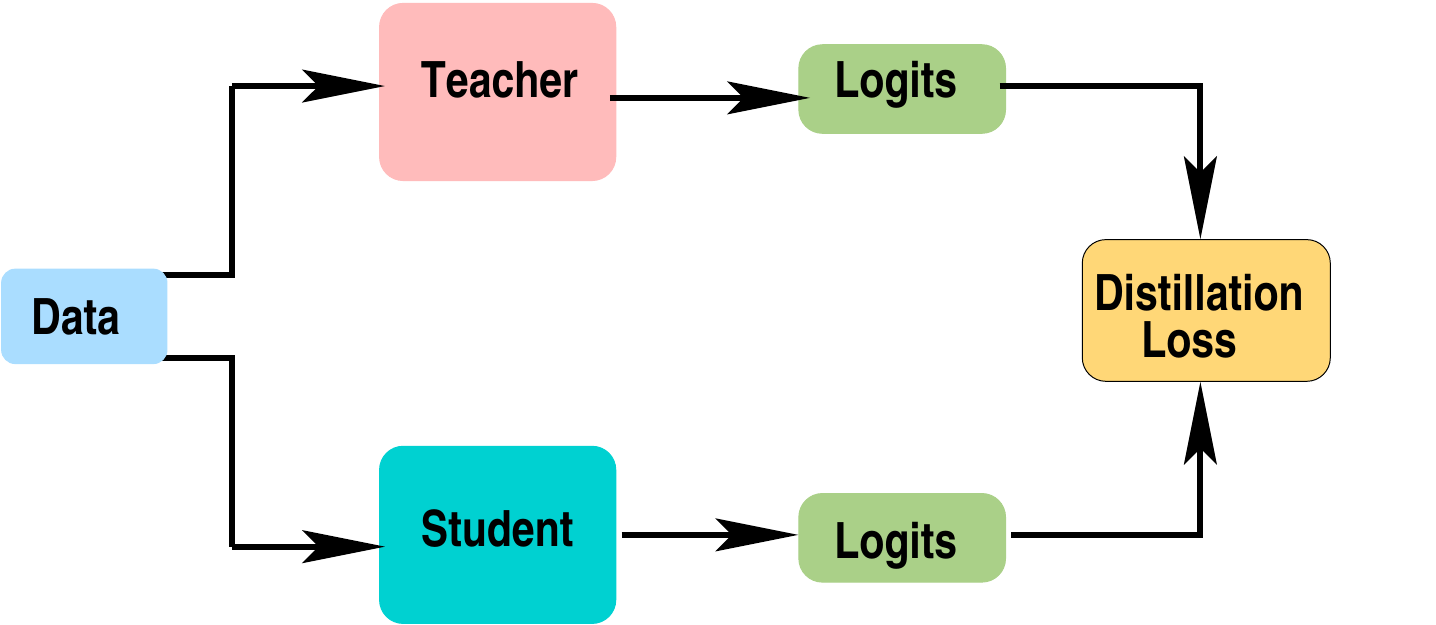}}
  \caption{Teacher–student knowledge distillation. A pretrained teacher model produces logits and softened output probabilities for the training data, and a smaller student model is trained to match these outputs while also fitting the original targets. The distillation loss transfers information from the teacher’s predictive distribution to the student, allowing the student to imitate the teacher’s behavior with a more compact model.}
  \label{fig:KnowDist}
  \end{figure}

In the classical setting, these probabilities are then used to classify item $i$
and compute the loss from the ground truth.  However, the argument made in
\citep{HintonAlDistill15} is that there is much more information encoded in $z_i$
and $p_i$ -- and this information dubbed `dark knowledge' can be essential in
ensuring good generalization. In order for the student model to achieve good
results it must aim at reproducing the probabilities $p_i$ of the teacher model.
Notice that a high temperature softmax function will produce a softer
probability distribution across the classes which means that the values in $p_i$
will not vary as much as when $T$ is small.  The idea is to select the same high
temperature $T$ for both the teacher and student model and then train the
student model so that it aims at reproducing the output of the teacher model.
The diagram shown in Figure~\ref{fig:KnowDist} illustrates the teacher-student
distillation process.

When the student model is trained in turn 
for the same data provided and the same temperature $T$, 
it will produce logit vectors $z_i\up{s}$  for $i=1:n$.
Applying the softmax to the output will yield the probability vectors: 
\eq{eq:piqi2}
q_i =\frac{ \exp(z_{i,:}\up{s}/T) }{\sum_j \exp(z_{i,j}\up{s}/T) } .
\en
The cost function is made up of two parts. The first consists of the usual Cross-Entropy
(CE) loss for the student model, see Equation \nref{eq:CEcost}, seen earlier.
This part is the standard CE calculated with the usual
temperature $T=1$ and is independent of the distillation process. We refer to
this cost as $\scl_{CE}$. The second part is the distillation cost which
measures the deviation between the teacher and student models based on the
probabilities computed in \nref{eq:piqi1} and \nref{eq:piqi2}.
For this we use the Kullback–Leibler
divergence measure: 
\eq{eq:distKL} \scl_{dist} = \frac{1}{n} \sum_{i=1}^{n}
D_{KL} \left(P(x_i) \,||\, Q(x_i) \right) = \frac{1}{n} \sum_{i=1}^{n} \left\langle p_i,
  \log ( p_i  \oslash q_i) \right\rangle \ .
\en
Here $p_i, q_i$ are the vectors of probabilities produced above
(equations (\ref{eq:piqi1}--\ref{eq:piqi2})), $\langle . , . \rangle$ is a dot
product of vectors, and $\oslash$ stands for a componentwise division of
vectors. As usual, the function $\log$ is applied componentwise to $p_i \oslash q_i$,
which is the componentwise division of $p_i$ by $q_i$.

  The final loss function combines the distillation loss and the student model
  loss as follows:
  \eq{eq:FinLoss}
  \scl_{mod} = \alpha  \frac{1}{T^2} \scl_{dist} + (1-\alpha) \scl_{CE}
  \en
  where $\alpha $ is a hyperparameter.  As can be seen, the goal of the 
  loss function is to ensure that the model is trained to achieve a compromise
  between  mimimizing the classical loss $\scl_{CE}$ while also trying to reproduce the teacher
  model. It is shown in~\citep{HintonAlDistill15} that
  the magnitudes of the gradients produced by soft targets scale like
  $1/T^2$ which
  motivates the scaling in the first term of \eqref{eq:FinLoss}.


  Note that in the original idea described in \citep{Bucila06} the teacher model
  is an ensemble of models instead of a single one. There is no difficulty in
  generalizing the principles just discussed to this situation and one can expect
  better improved results with such a setting. 

\subsubsection{Low-Rank Factorization methods}
These methods approximate weight matrices with lower-rank decompositions to
reduce the number of parameters. This is proposed in \citep{LowRankExpansion2014}
in the context of Convolution Neural Networks.  A more general technique known as LoRa - also rooted in low-rank approximations of the parameters is described
next.
 
\subsection{LoRa: Exploiting Low-rank structure in LLMs}
\label{section_LoRA}

The number of parameters required to represent a given model
can be enormous, e.g., in the trillions for the most powerful
LLMs, such as GPT-4. Starting in the mid 2015's, researchers
began exploring the nature of these parameters as well as that of the
Hessians encountered during optimization, see, e.g.,
\citep{DOSDNN19,HessianSpect17,IntrinsicDim19,IntrinsicDim20,LoSparse-paper23,
  huh2023the,LoRaPaper21,yaras2020law}, among many others.

Two important observations were made. First concerning the Hessians, it was
noted that as the iterates near convergence, the bulk of the eigenvalues of these
matrices tended to lie near the origin, with a few 
 outliers located away from this cluster 
\citep{SagunSpectHess16,HessianSpect17,DissectingHessians20,papyan19Hessian,DOSDNN19}.
In other words the Hessian is, approximately speaking, a low rank matrix but there was more to the
analysis that this simple fact. Among other things, the articles discussed the
impact of the number of layers on the spectrum. 
The second observation made was that in spite of large the number of parameters,
the dimension of the space of parameters tends to be small,
and this is particularly true near convergence
\citep{IntrinsicDim19,IntrinsicDim20,huh2023the,LoRaPaper21,yaras2020law}.
In fact, it was observed that higher depths, i.e., with
over-parameterization in DNN,  leads to lower dimensions
in the parameter space. The term ``Law of parsimony'' was coined to describe
this phenomenon  in \citep{yaras2020law}.

A  contribution related to this discovery which had a major practical impact
is the work on Low-Rank adaptation of LLMs \citep{LoRaPaper21}.
This paper addressed the problem of `fine-tuning' pre-trained models.
It is common practice to take an already trained model and perform a new
training to modify the weights in order to take into account new data
or to adapt the model for new tasks. This fine-tuning is time-consuming
and impractical. Indeed, it essentially requires retraining the model
with an initialization based on the previously obtained parameters. 
The idea of LoRa is rather simple to understand from a linear algebra view-point:
Given a set of pre-training parameters $\bm{W}_0 \in \RR^{d \times d_v}$,
e.g., associated with a certain layer,
we will update the parameters in the form
\eq{eq:LoraUpdt}
\bm{W} = \bm{W}_0 + \bm{A} \bm{B}
\en
where $\bm{A} \ \in  \ \RR^{d \times r} $  and $\bm{B} \ \in  \ \RR^{r \times d_v}$
are two low-rank matrices.
The parameter set $\bm{W}_0$ is untouched. What is trained is the set of
parameters contained in $\bm{A}, \bm{B}$ which is  much smaller.

\begin{figure}[tb]
\centerline{
 \begin{tikzpicture}[
    >=Stealth,
    blkg/.style={fill=green!30!gray!30,align=center,minimum width=15mm},
    blky/.style={fill=yellow!50,align=center,minimum width=15mm},
    add/.style={font=\Large, inner sep=0},
 ]
\definecolor{blue2}{rgb}{0,0,0.69}
\definecolor{cyan3}{rgb}{0,0.82,0.82}
\definecolor{red3}{rgb}{0.82,0,0}
\definecolor{mrn}{rgb}{0.867,0.667,0.533}
\pgfsetlinewidth{2pt}
\node[fill=red!20](W0)   at (0, 0) {$W_0$ \textbf{(Frozen)}};

\pgfsetstrokecolor{blue}
\pgfsetfillcolor{mrn}

\Large
\boldmath 

\filldraw[fill=green!30!gray!20,very thick] (-2,-1) -- (-2,2) -- (2,2) -- (2,-1) -- (-2,-1);

\node (w0) at (0.12,1.25) {$W_0 \ \textcolor{red}{(d \times d_v)}$};
\node (fr) at (0.15,0.25) {\textbf{(Frozen)}};
\node [rectangle,draw,fill=gray!10,very thick] (c) at (2.75,4.5)  {$h = x \times W_0 + x \times A B$};

\draw[very thick, black][->]  (0.0, 2.0) -- (2.5,3.5) ;
\draw[very thick, black][->]  (5.7, 2.4) -- (3.0,3.5) ;

\filldraw[fill=mrn!30,very thick] (3.5,-1) -- (3.5,2) -- (3.8,2) -- (3.8,-1) -- (3.5,-1);
\filldraw[fill=mrn!30,very thick] (3.9,2) -- (3.9,2.3) -- (7.5,2.3) -- (7.5,2) -- (3.9,2);
\node (rr) at (6.0,0.5) {\textbf{(Trained)}};
\node (aa) at (5.4,-0.8) {$A \ \textcolor{red}{ (d \times r)}$};
\node (bb1) at (7.8,2.2) {$B$};
\node (bb2) at (5.8,1.6) {$\textcolor{red}{(r \times d_v)}$};
\node (op) at (2.75,3.6) {$\oplus$};


\filldraw[fill=blue2!60,thick] (0.75,-2.5) -- (.75,-2.3) -- (4.75,-2.3) -- (4.75,-2.5) -- (.75,-2.5);
\node (xx) at (6.55,-2.405) {$x \quad \textcolor{red}{ (1 \times d)}$};

\draw[very thick, black][->]  (2.73, -2.2) -- (0.0,-1.05) ;
\draw[very thick, black][->]  (2.77, -2.2) -- (3.65,-1.05) ;

\draw[dashed,very thick, red]  (2.73, -1.5) -- (2.73,3.0) ;

\end{tikzpicture}
}
\caption{Forward step in LoRa. A frozen pretrained weight matrix $\bm{W}_0$ is augmented by a trainable low-rank update ${\bm{A}} {\bm{B}}$, where ${\bm{A}} \in {\mathbb{R}}^{d\times r}$, ${\bm{B}} \in {\mathbb{R}}^{r\times d_v}$, and $r \ll \min(d, d_v)$. Instead of updating the full weight matrix, LoRa keeps $\bm{W}_0$ fixed and learns only the small-rank factors $\bm{A}$, and $\bm{B}$, thereby reducing the number of trainable parameters while preserving the original forward structure.}
\label{fig:LoRaFig}
\end{figure}
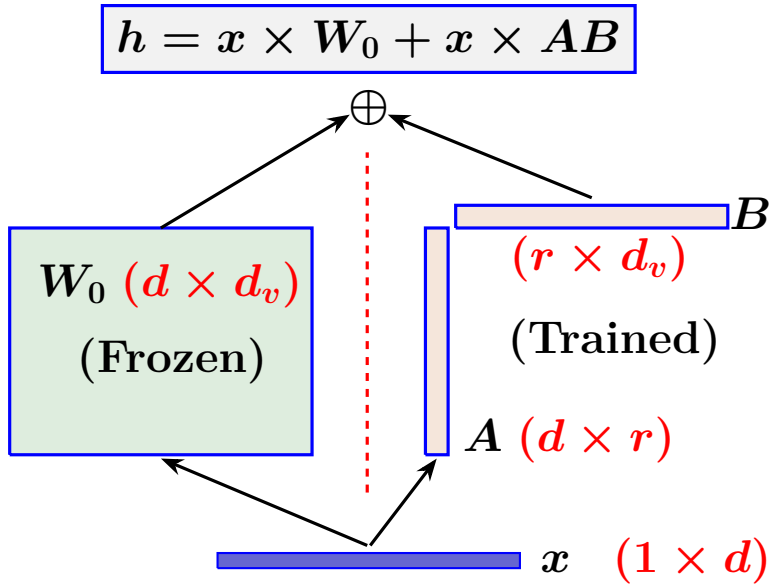
This can be done for adapting any neural network model.  For the case of
LLMs  specifically, the authors of \citep{LoRaPaper21}
suggest performing the adaptation only for the
self-attention weights, namely the weights denoted
by $\bm{W}_q, \bm{W}_k, \bm{W}_v $ in Equations~\nref{equa5.1} and \nref{equa5.3}.

In an illustrative example, the article showed that   {LoRa} was
able to reduce number of parameters needed for fine-tuning in GPT3
from 175B to $~ 17M$, which represents a gain of 10,000 fold.

On the theoretical side, many papers offered explanations of the
phenomenon, this bias toward parameters of  lower rank in DNN.
The paper \citep{huh2023the} discusses this with detail
and offers a theoretical explanation grounded in random matrix theory.

\subsection{The idea of `linear' transformers} \label{sec:LinTransf}
We now revisit the formula for computing the self-attention at stage $l$ of a
transformer, see Equations \nref{eq:AttDef} and \nref{eq:AlDef}, which we
rewrite here for convenience: 
\eq{eq:attnExpl}
\bm{A}_\ell(\bm{X}_{\ell-1}) = \operatorname{softmax}  \left(
  \frac{\bm{Q}_l \bm{K}_l^T}{\sqrt{d}} \right) \bm{V}_{\ell}  
\en
From a computational point of view,
the above formula will usually constitute a bottleneck. Indeed, assuming, as is
common practice, that each of the three matrices $\bm{Q}_l, \bm{K}_l$ and $\bm{V}_l$ is of
dimension $n \times d$,   where here $d$ stands
  for $d_k$, then we would normally compute the rows of $\bm{Q}_l \bm{K}_l^T$
for $i=1,2,\cdots,n$ and then apply the softmax operation to each row $i$.  This
would result in a row vector, which we call $z_i$ for $i=1:n$. The last step is
then to compute $z_i \bm{V}_l$ which is a $d$-dimensional (row) vector. The
computation of the rows of $\bm{Q}_l \bm{K}_l^T$ is an order $n^2$ process. Indeed, we
have to compute the rows of this matrix explicitly for $i=1:n$ in order to be
able to apply the softmax function to them. Recall that the whole row is needed because of
the normalization, see Equation~\eqref{eq:softmax}. 
Computing the rows of $\bm{Q}_l \bm{K}_l^T$
costs $ \sim 2 n^2 d$ operations. The softmax operation itself also costs
$O(n^2)$ operations since it is applied to each entry of the matrix $\bm{Q}_l \bm{K}_l^T$.

\begin{figure}[bth]
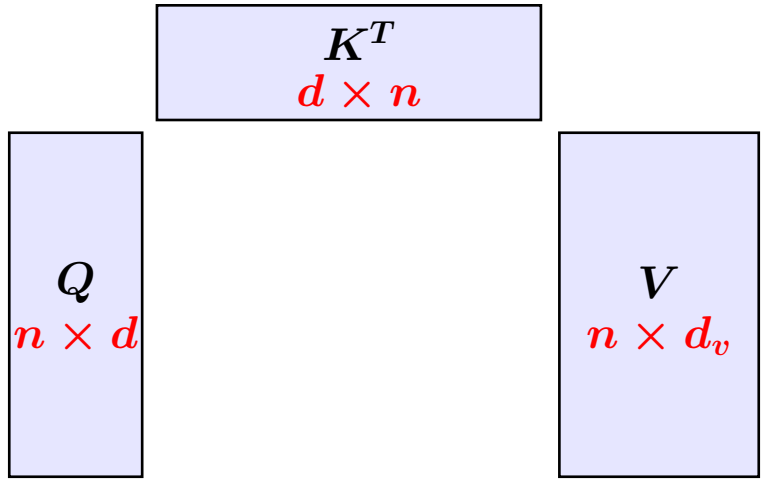

  \centerline{

\dimen1=3643sp
\dimen3=3143sp

\tikzpicture[x=+\dimen1, y=+\dimen3]

\tikzset{inner sep=+0pt, outer sep=+0pt}
\pgfsetlinewidth{1pt}
\pgfsetstrokecolor{black}
\filldraw[blue!10,draw=black] (1800,-2025) rectangle (2700,-4725);
\filldraw[blue!10,draw=black] (2800,-1025) rectangle (5400,-1925);
\filldraw[blue!10,draw=black] (5525,-2025) rectangle (6875,-4725);
\boldmath
\Large
\node (q) at (2250,-3200) {$Q$} ;
\node (vv) at (6200,-3200) {$V$} ;
\node  [below of = vv, node distance = 20pt]  {$\textcolor{red}{n \times d_v}$} ;
\node  [below of = q, node distance = 20pt]  {$\textcolor{red}{n \times d}$} ;
\node (kt) at (4170,-1300) {$K^T$} ;
\node  [below of = kt, node distance = 20pt]  {$\textcolor{red}{d \times n}$} ;
\endtikzpicture%
} 
  \caption{Illustration of linearized attention. If the softmax is temporarily omitted, the attention map takes the form ${\bm{A}}_{\ell}^{(\text{lin})} ( {\bm{X}}_{\ell-1} ) = {\bm{Q}}_{\ell} ( {\bm{K}}_{\ell}^T {\bm{V}}_{\ell} ) / { \sqrt{d_k} }$, which can be evaluated by first forming the smaller matrix ${\bm{K}}_{\ell}^T {\bm{V}}_{\ell} \in {\mathbb{R}}^{d\times d_v}$ and then multiplying on the left by ${\bm{Q}}_{\ell}$. This reordering avoids the explicit formation of the $n\times n$ score matrix and highlights the computational motivation behind so-called linear Transformers.}
  \label{fig:LinAtt}
\end{figure}

Assume for a moment that there is no softmax function involved.
Then the computation of this `linearized attention' would become: 
  \eq{eq:attnExplL}
 \bm{A}_l^{(lin)} (\bm{X}_{l-1}) =\left( \frac{\bm{Q}_l \bm{K}_l^T}{\sqrt{d}} \right) \bm{V}_l
=  \frac{1}{\sqrt{d}}  \bm{Q}_l \left( \bm{K}_l^T \bm{V}_l \right) . 
\en
This is illustrated in Figure~\ref{fig:LinAtt}.
  Evaluating  the expression as is suggested by the parentheses
  on the right side of \nref{eq:attnExplL} is an
  $O(n)$ process. Indeed, the cost of evaluating
  $\bm{K}_l^T \bm{V}_l$, which is a $d \times d$ matrix, is $2n d^2$. Once this matrix is
available we multiply $\bm{Q}_l$ by it on the right at the cost of another $2 d^2 n$ operations. 
  The scaling by $\sqrt{d}$ of the final result costs $n d$ multiplications. 

The idea of decoupling the softmax in \eqref{eq:attnExpl} in two seems to have been first
discussed in \citep{ShenEffAtt21}. The authors used a very simple remedy to the difficulty
discussed above. 
They describe their method for the case when the scaling  by $\sqrt{d}$ in \eqref{eq:attnExpl}
is omitted. 
With this modification they approximate the  matrix in \eqref{eq:attnExpl} as follows:
\eq{eq:ShenLinTrans}
\operatorname{softmax}\left(\bm{Q}_l \bm{K}_l^T \right) \approx
\operatorname{softmax}_Q \left( \bm{Q}_l  \right) \times
\operatorname{softmax}_K \left( \bm{K}_l  \right)^T ,
\en 
where $\operatorname{softmax}_Q \left( Y \right) $ applies the softmax function
rowwise to $Y$, while $\operatorname{softmax}_K \left( Y \right) $ applies it 
 columnwise to $Y$. With this way of writing the attention matrix,
it is clear that we can now apply the same idea as for the linear case where we have no
softmax function, i.e., $\bm{A}_l(\bm{X}_{l-1})$ is computed as in \eqref{eq:attnExplL},
yielding: 
\eq{eq:ShenLinTrans2}
\bm{A}_l(\bm{X}_{l-1}) \approx 
\operatorname{softmax}_Q \left( \bm{Q}_l  \right) \times
\left( \operatorname{softmax}_K \left( \bm{K}_l  \right)^T \bm{V}_l \right) .
\en 
Clearly the new approach results in an approximation of the original method but the
authors indicate that this approximation is fairly accurate in practice.

Another linearized attention method discussed in \citep{TansfAreRNNS} (see also
\citep{performer20}) consists of approximating the softmax operation in
\nref{eq:attnExpl} by an expression that decouples into a $\bm{Q}_l$ and a $\bm{K}_l$
part, so that something similar to the linearized version can be applied.  In
effect they replace $\operatorname{softmax}  (\bm{Q}_l \bm{K}_l^T) $ by $\phi(\bm{Q}_l) \phi( \bm{K}_l^T) $
where $\phi$ is a carefully selected function, leading to the following
approximation: 
\eq{eq:attnExplb} \widehat{\bm{A}}_l (\bm{X}_{l-1}) =\left( \frac{\phi(\bm{Q}_l)
    \phi(\bm{K}_l)^T}{\sqrt{d}} \right) \bm{V}_l = \frac{1}{\sqrt{d}} \phi(\bm{Q}_l) \left(
  \phi(\bm{K}_l)^T \bm{V}_l \right) .
\en

The softmax operation consists of a \emph{Kernel} that acts on a pair of
vectors to produce a scalar which serves the purpose of measuring the
similarity between these two vectors. It plays the role of an inner product
except that the bilinear form is replaced by a nonlinear operation. In this
particular case, the pairs of vectors are a row of $\bm{Q}_l$ and a row of $\bm{K}_l$.
The point is that this kernel can be well approximated by another kernel
which first transforms (nonlinearly) the two vectors and then takes a
standard inner product of the transformed vectors.

In the article \citep{TansfAreRNNS}, the function $\phi$ is set to
be equal to $\phi(t) = \texttt{elu}(t) +1 $ where the
\emph{Exponential Linear Unit} function \texttt{elu} is defined as: 
\eq{eq:ELU}
\texttt{elu}(t)
= \left\{\begin{array}{lcl}
           t & \text{if} & t > 0 \\
           \alpha (e^t -1) & \text{if} & t \le 0
         \end{array} \right.
       \en
       in which  $\alpha $ is a parameter.
The intriguing title of the article (`Transformers are RNNs ...')
comes from the observation made by the authors that when we enforce
causal masking, where a sample $i$ can be only be influenced by
samples with position $j\le i$, then the linear form of an
attention step resembles that of a recurrent neural network.

Since  attention constitutes one of the most expensives parts of a transformer,
it should not be surprising that there has been a flurry of activity dedicated to
developing other efficient alternative schemes.
The survey paper \citep{EfficientAttentionSurvey22} describes many of the
ramifications of the work based on exploiting a Kernel viewpoint and other  non-related
ideas. These are further discussed in  the next section. 

    
    \subsection{Randomization and random projections approaches} \label{sec:sketching}
     Randomized techniques have emerged in recent years as a powerful
    paradigm in numerical linear algebra, see, e.g.,
    \citep{Halko-al11,LibertyAl07,mahoney2011randomized,motwani1995randomized}
    among many references.  In a broad sense, the main ingredient of randomized
    algorithms is to exploit stochastic processes as a means to solve a given
    problem.  The term `Sketching' often refers specifically to  a class of methods
    where randomization is exploited  for the task of compressing  data.
    Given the effectiveness of randomized techniques in NLA, especially in a context
    where approximate methods suffice, their adoption in LLM should not be too surprising.

    The observation that has been made is that ``data tends to have a small rank
    character'' which means that the rank of real-world datasets when
    represented as matrices or tensors is usually small, much smaller than any
    of its intrinsic dimensions.  One of the reasons for this phenomenon is that
    real-world data is often correlated, among features or samples. This leads
    to redundancy, meaning that the effective rank is much lower than the actual
    dimensionality. Another advocated explanation is the so-called `manifold
    hypothesis' \citep{ManifoldHypothesisBengio13} which suggests that
    high-dimensional data near a low-dimensional manifold lies within the original
    high-dimensional space.  When matrices have a  low rank, they can easily be
    approximated by randomized techniques and this has been exploited repeatedly in
    neural networks.





    A technique that is similar to the one described in the previous section is 
    the algorithm known as `linformer' \citep{wang2020linformer}.  Linformer uses
    low-rank approximations to reduce the cost of the attention layers in
    transformers and is one of several methods that aim at reducing the cost of
    attention mapping from $O(n^2)$ to $O(n)$, see
    \citep{EfficientAttentionSurvey22} for a survey.  The key idea of linformer
    is to approximate the attention matrix $\bm{P}_\ell$
    shown in Equation~\eqref{eq:AttDef},
    and denoted by $P$ in the paper, by a low rank matrix
    \citep{wang2020linformer}.  This matrix $P$ is termed a context mapping
    matrix which, using the words of the authors, `captures the input context
    for a given token, based on a combination of all tokens in the
    sequence'. The authors invoke the Johnson–Lindenstrauss lemma discussed in
    Section \ref{sec:geometry_JL} to prove that with high probability $P$ can be
    approximated by a low-rank matrix.

In fact the matrix $P$ is not approximated directly as this is again costly. 
Instead, recall that in transformers the matrices $\bm{Q}_l $, $\bm{K}_l$, and
$\bm{V}_l$ all result
from multiplying the same input matrix $\bm{X}_{l-1}$ by parameter matrices $ \bm{W}_Q\up{l} $,
$ \bm{W}_K\up{l}$ and $ \bm{W}_V\up{l}$ respectively, see equation~\nref{equa5.3}.
In the following we omit the layer index $l$. 
The idea is to project the matrices $\bm{K} $ and $ \bm{V}$ into
$k \times d$ - dimensional projected key and value matrices. This is achieved by
\emph{random projectors} $\bm{E}\in \ \RR^{k \times n} $
for $\bm{K} $ and
$\bm{F} \in \ \RR^{k \times n} $ for $\bm{V} $.  With this the approximation
to  the product \eqref{eq:AlDef} is given by:
\eq{eq:LinFormAtt}
\operatorname{LinFormATT} (\bm{Q},  \bm{K}, \bm{V}) =
\operatorname{softmax} \left(\frac{ \bm{Q} (\bm{E} \bm{K})^T}{\sqrt{d}} \right) (\bm{F} \bm{V}),
\en
which is inexpensive to evaluate since $k \ll n$.
Note that this operation is performed for each
head in the multihead attention layer and that $\bm{E}, \bm{F}$ are specific to each attention head. 

Another   line of research that has exploited randomization is represented by the 
\texttt{performer} algorithm \citep{performer20}.
This work is a good example of an idea from classical Machine Learning that is
adapted to DNNs.
Indeed, the authors were inspired by randomized schemes to train
kernel Support Vector Machines (SVM) with large training data \citep{RandomFeatures}.

We observed earlier that the attention matrix in \eqref{eq:AttDef} can be viewed as a
Kernel matrix and the methods seen in Section~\ref{sec:LinTransf} aimed at approximating
this kernel in the form $K(x,y) \approx \phi(x)^T \phi(y)$, see Equation \eqref{eq:attnExplb}
and comments that follow. 
In \citep{performer20} this is explored from a different angle.
The original attention matrix is written in the form
\begin{equation}\label{eq:performer1}
  A = D\inv \hat A \quad \mbox{where} \quad
  \hat A = \left[\exp \left(\frac{\bm{Q}_i \bm{K}_j^T }{\sqrt{d}}\right) \right]_{i,j=1:n}  \quad
  D = \diag{(\hat A \ \One) } .  
\end{equation}
Here $\bm{Q}_i$ is a the $i$-th row of $\bm{Q}$ and  $\bm{K}_j$ is a the $j$-th row of $\bm{K}$, both
having $d$ entries. 

The exponential of the scaled inner product $\bm{Q}_i \bm{K}_j^T / \sqrt{d}$ is decomposed by using the
equality $x^T y = \half(\|x\|_2^2 +\| y \|_2^2 - \| x - y|_2^2)$ which yields: 
\[
  \exp\left(\frac{x^T y}{\sqrt{d}}\right) = \exp\left( \frac{\| x \|_2^2}{2\sqrt{d}} \right)  \cdot 
    \exp\left( \frac{\| x -y  \|_2^2}{2\sqrt{d}} \right) \cdot  \exp\left( \frac{\| y \|_2^2}{2\sqrt{d}} \right) .   \]
  Hence, with $r = 2\sqrt{d}$ and setting
  \begin{align}
    D_Q &= \diag \left( \exp( \|\bm{Q}_1\|_2^2/r), \cdots, \exp( \|\bm{Q}_n\|_2^2/r) \right) 
          \label{eq:DQ}\\
    D_K &= \diag \left( \exp( \|\bm{K}_1\|_2^2/r), \cdots, \exp( \|\bm{K}_n\|_2^2/r) \right)
          \label{eq:DK}
     \end{align}
     we obtain:
\begin{equation}\label{eq:bij}
  \hat A = D_Q B D_K  , \quad B \in \RR^{n \times n}, \quad 
  B_{ij} = \exp \left(- \| \bm{Q}_i - \bm{K}_j\|_2^2/r \right) . 
\end{equation} 
The main point  of the method is to find an \emph{unbiased stochastic approximation} of the matrix
$B$. Note that the matrix $B$ defined in \eqref{eq:bij} is the Gaussian kernel
$\mathcal{K}_{gauss}\up{\sigma} $ with $\sigma = d^{1/4}$ evaluated at the
vectors $\bm{Q}_i^T, \bm{K}_j^T$, that is:
\[
  B_{ij} = \mathcal{K}_{gauss}\up{\sigma} (\bm{Q}_i^T, \bm{K}_j^T) \equiv 
  \exp \left( - \frac{\| \bm{Q}_i - \bm{K}_j\|_2^2}{ 2 \sigma^2} \right)
\]
We can  obtain a low-rank approximation of $B$ using \emph{random features}
\citep{RandomFeatures}. 
Random features evaluate  a kernel $\calK$ from the expectation
\begin{equation}\label{eq:RF}
  \calK(x,y) = \EE (\phi (x)^T \phi (y) ) 
\end{equation} 
where $\phi: \RR^d \to \RR^M$.
In the above equation $\phi$ is a Random Feature (RF) map selected at random and
the expectation in \nref{eq:RF} is with respect to these maps.
Random feature maps can be generated for a large class of Kernels and they are
typically of the following form where $f$ is a function $f : \RR \to \RR$:
\begin{equation}\label{eq:RFf}
  \phi(x) = \frac{c}{\sqrt{M}} f(\bm{W} x + b) ,
  \quad \bm{W} \in \ \RR^{M \times d} \ ,  b \ \in \ \RR^{M } ,
\end{equation}
where $c>0$, and     the rows $\bm{W}_i$ of $\bm{W}$ and the entries $b_i$ of $b$ are drawn
independently, each from the same distribution.
For a  Gaussian, $c = \sqrt{2}$,  $ f(t)  \equiv \cos(t)$ and: 
$\bm{W}_i, \bm{W}_2, \cdots, \bm{W}_M \sim_{iid} \calN (0,\sigma^2 I)$, 
$b_1, b_2, \cdots, b_M \sim_{iid} \text{Unif} (0, 2\pi)$. 

With this we define the intermediate matrices:
\begin{equation}\label{eq:hQhK}
  \widehat{\bm{Q}}^T  =  \frac{c}{\sqrt{M}} f(\bm{W} \bm{Q}^T +b) \ ,\qquad
  \widehat{\bm{K}}^T  =  \frac{c}{\sqrt{M}} f(\bm{W} \bm{K}^T +b) \qquad 
\end{equation} 
and take the expectation of $\widehat{\bm{Q}} \widehat{\bm{K}}^T$ to obtain $\calK$ according to
\eqref{eq:RF}. This will give the matrix $B$ in \eqref{eq:bij}.
Then the desired matrix $\hat A$ in \eqref{eq:bij} is equal to
\begin{equation}\label{eq:RFB}
  B = \EE (D_Q \widehat{\bm{Q}} \widehat{\bm{K}}^T D_K ) . 
\end{equation}
A few clarifications are necessary. First, while  the above expression is an equality 
in practice we will only compute an approximation taking one set of samples, i.e.,
one pair $\widehat{\bm{Q}}, \widehat{\bm{K}}$ given by \nref{eq:hQhK} which corresponds to $M$ 
random feature functions. In other words, $B$ is approximated by
\begin{equation}\label{eq:RFsamples}
  \hat B \approx D_Q  \widehat{\bm{Q}}  \ (D_K \widehat{ \bm{K}})^T .
\end{equation}
Our second clarification is that we do not actually even compute the
matrix $\hat B$ shown above. The actual computation of the approximation to
$A  V$        where $A$ is shown in \eqref{eq:performer1} and
$\bm{V} \in \RR^{n\times d}$, is performed as follows.
First compute the matrices $D_Q, \bm{Q}_K$ at the outset from  \eqref{eq:DQ}
\eqref{eq:DK}         and form $C = [V, \One_n]  \in \RR^{n \times (d+1)}$. 
Then compute $D_Q \widehat{ \bm{Q}}$ and $D_K \widehat{\bm{K}}$ from \eqref{eq:hQhK} and
then $\bm{X} = (D_K \widehat{\bm{K}})^T C$ and $\bm{Y} =  (D_Q \widehat{\bm{Q}}) \bm{X}$.
Finally, with $ D = \diag (Y(:,d+1)) $  compute
the desired result $D\inv \bm{Y}(:,1:d)$ which corresponds to $D\inv \hat A \bm{V}$ from 
Equation~\eqref{eq:performer1}.

The above approximation is called Fast Attention via Orthogonal Random
features(FAVOR) by its authors. 
The complexity of FAVOR is similar that of the linear attention methods presented
in Section~\ref{sec:LinTransf} because the innermost calculation is of the same type.
The time complexity of the algorithm is $O(n M d)$ and the memory complexity
is $O(Md +nd + Mn)$. 
It is also important to note that the resulting approximation to the matrix
$A$ is of rank $M$. We also note that the above description is for the bi-directional
attention where there is no masking but the authors also present an algorithm for the
uni-directional case.



\section{Advanced optimization for Transformers}\label{sec:opt}

We saw  in Section~\ref{sec:TrainDNN} a few  optimization techniques 
commonly used to  train neural networks. The methods we discussed, including
the Stochastic Gradient Descent (SGD) algorithm, AdaGrad, and Adam, are 
mostly of the first-order type. As a reminder first order methods in
optimization refer  to  iterative algorithms that primarily utilize gradient
(or subgradient) information of a function to find its minimum. 
In contrast,  second order
methods aim at incorporating information related to second derivatives, as
represented by the Hessian matrix, in an effort to provide faster convergence, often at a
much higher computational cost per step.
  
The question at this point is whether or not higher order methods can be put to
good use in training DNNs.
This question is addressed in the  review article \cite{BottouCurtisNocedal18}.
To find the minimun of a function $\phi(w)$ that is
twice differentiable, we could use the local quadratic approximation
\eq{eq:LocQuad}
\phi(w + \delta) \approx \phi(w ) + \nabla \phi ^T \delta +
\frac{1}{2} \delta^T H \delta
\en
where $ \nabla \phi $ is the gradient of $\phi$ at $w$ and $H$ its Hessian.
When $H$ is positive definite, as is the case when $\phi$ is convex, then the
minimum of the above quadratic approximation is reached at
$\delta = - H \inv \nabla \phi$. This leads to Newton's
iteration:
\eq{eq:Newt} w_{k+1} = w_k - H_k\inv \nabla \phi_k ,
\en
where $H_k$ and $\nabla \phi_k $ are, respectively, the Hessian and the
gradient, at the current iterate $w_k$.  The advantage of Newton's method in a
classical optimization context is that when it converges then it
does so   quadratically. 
However, this advantage comes
at a high cost since the Hessian is expensive to compute and store and we now
need to  solve the  related Newton systems
$H_k \delta_k = -\nabla \phi_k$ at each step.  A common alternative is to
replace $H_k$ by a simpler matrix, say $P_k$, which is easy to form and for
which solving the systems $P_k \delta_k = -\nabla \phi_k$ is not costly. For example
$P_k$ can be just a diagonal matrix or a low-rank modification to a diagonal
matrix as is the case in Quasi-Newton matrix
\cite{Dennis-Schnabel,fletcher-book,Nocedal-Wright-book}.  In the machine
learning literature $P_k$ is often termed a `preconditioner'.\footnote{This is an
  improper terminology from a numerical linear algebra viewpoint where a
  preconditioner is an approximation to the coefficient  matrix of a linear system
  which  makes the solution of this system easier to solve by
  an iterative procedure.}.
We should note if we   replace $H$ by $\eta\inv I$ in \eqref{eq:LocQuad} then the quadratic term
acts as a  penalty term and the solution of the resulting problem would lead to
the Cauchy gradient step:
\eq{eq:GradStep1}
w_{k+1}  = w_{k} - \eta \nabla \phi_k . 
\en

There are several difficulties with second order methods for Deep Learning.
First among these is the lack of convexity of the objective functions invoked in
this context. As a result all methods that feature a second order character,
such as a Quasi-Newton approach, will have both theoretical and practical
difficulties.  A Quasi-Newton approach will incur substantial additional cost
(memory and arithmetic) per step and yet it may result in no or little
acceleration, if not in a breakdown caused by the non SPD nature of the Hessian.
Another problem with second order methods is the stochastic nature of the
underlying optimization procedures in use, dictaded by practical
considerations. As was explained in section~\ref{sec:SGD}, the full gradient is
rarely invoked due to computational cost issues. Instead, this gradient is
`sampled' at each step from a batch of gradients of functions that constitute
the `finite sum' that defines the loss function, see \nref{eq:fw}.  This
randomness means in effect that the function changes at each step. The mean of
the sampled functions converges in a probabilistic way to the full gradient but
the erratic nature of the samples makes it difficult to build a good
approximation to the Hessian in a cost-effective manner. A third disadvantage of
second order methods is their computational cost. Just from the point of view of
memory, storing a few directions as is often required by Quasi-Newton type
methods would be prohibitive in general.

For all these reasons, methods that are genuinely of second order type have seen
little use in deep learning. On the other hand, a number of ideas have recently
emerged that try to imitate the performance of second order methods.  These
techniques are labeled second-order only because they attempt to use second
order information in an inexpensive way, but one should not expect them to
attain quadratic or superlinear convergence as is the case in the classical
optimization context.

\subsection{Natural Gradients}\label{sec:NatG}
A key idea  in this context is that of \emph{Natural Gradients}
introduced by Amari in \cite{amari98}. In this paper, the author argued that the
standard Euclidean distance may not be the most suitable metric to use for a gradient descent-type
algorithm. He suggested to replace the standard Euclidean distance with
a general Riemannian metric where the length of an infinitesimal increment is
defined generally from a quadratic of the form
\eq{eq:Amari1}
\| dw \|_G^2    = \sum_{ij} g_{ij}(w)  dw_i dw_j .
\en
If the parameter space $S$ to which $w$ belongs is a (curved) manifold, a length
on the manifold needs to be written in this form where the matrix $G(w)$, called
the \emph{Riemannian tensor matrix}, depends on $w$.
If we are given a cost function $L(w)$ to mimimize around a current $w$,
the question asked here is: What is, to first order approximation, the
change $\delta$ that minimizes the first order
approximation $L(w+\delta)$ under the constraint
with $\| \delta \|_G = \epsilon $? The author showed that the answer to the question is a vector that
lies in the direction  of the negative of the \emph{Natural Gradient in the Riemannian space}
defined as follows:
\eq{eq:NatG} \widetilde{ \nabla }\phi(w) \equiv G(w)\inv \nabla
\phi(w).
\en
Thus, a \emph{Natural Gradient Descent} (NGD) algorithm should take the form 
\eq{eq:NatGD} w_{k+1} = w_k - \eta_k G(w_k)\inv \nabla \phi(w_k) , \en
where $\eta_k$ is a step-length similar to the one used for steepest descent.
Remarkably, note  that when $\eta_k \equiv 1$ we are simply replacing the
Hessian in Newton's method with the Riemannian tensor matrix.

When $G(w) = I$, we recover Cauchy's classical steepest descent algorithm,
see Equation~\eqref{eq:GradStep1} so
this is a generic way of generalizing gradient descent and it is implied that
the standard metric, namely $G(w)=I$  used by Cauchy's classical steepest descent,
 is not necessarily the best one to use for
optimizing $\phi(w)$ by an iterative procedure.
Thus, the natural gradient is motivated from the perspective of the
geometry of the parameter space:
It defines the direction in this parameter space that gives the largest
change in the objective per unit of change in the model, as measured by the
metric $G(w)$. In contrast, the standard gradient  is the direction that
gives the biggest change in the objective function per unit of change in the
parameters, where  this change is measured by the Euclidean distance.

We now adopt a viewpoint laid out in \cite{HeskesNGD00} to further motivate the
idea of natural gradient introduced above.
Suppose we have a model for a given parameter set $w$,
recalling that $w$ stands for a vector that holds all parameters.  The model
will provide an estimate $f(x,w) $ of a target $y$ for a given input $x$ and we
can compute the `distance' or `loss' $L(y,f(x,w))$ between the output $f(x,w)$ of the
model for input $x$ and the corresponsing target $y$ (thus $x,y$ is a pair 
in the training set).   Here $L$ is rarely a proper distance
and it may  not be symmetric.  For a different parameter set $w'$ we would
get  the loss $L(y,f(x,w'))$.
A  standard regularized gradient approach  corresponds
to setting as a new iterate $\hat w$ the minimizer in:
\eq{eq:NGwh} \min_{w'}
\left[ L(y,f(w',x)) + \frac{1}{2 \eta} \|w'-w\|_2^2 \right] .
\en
This can be viewed as a sort of trust-region approach where the quadratic
penalty term forces the new iterate to remain close to the current one.  It is a
penalized version of the Amari idea discussed above for the case when
$G(w) = I$, where we recall that in Amari's case we attempt to minimize $L(w+\delta)$ under the
constraint $\| \delta \|_G = \eta$.  Setting the gradient in \nref{eq:NGwh} to
zero yields the gradient descent iterate:
\eq{eq:NGwh1} \hat w = w - \eta \frac{\del
  L(y,f(x,w))}{\del w} .  \en

Using the Euclidean distance in the penalty term of \nref{eq:NGwh} seems somewhat
arbitrary. Instead, a more natural choice is a distance that compares the
outputs of the new model (with $w'$) with the old one (with $w$) or
$L(f(x,w'), f(x,w))$.
In the end we define
\eq{eq:NGwh2} L(w',w) \equiv L(f(w',x),f(x,w))  \ .
\en
For now we are considering one single pair $x,y$ of an input $x$ and its
associated target $y$. This particular scenario is referred to as the `on-line
learning' case. In `batch-mode learning' one will need to replace all quantities
involving the pairs $x,y$ by averages over batches as will be seen shortly.

Consider minimizing the following analogue to \nref{eq:NGwh}
\eq{eq:NGwh3}
\min_{w'} \left[ L(y,f(w',x)) + \frac{1}{ 2 \eta} L(w',w) \right] .
\en
If we write $w'=w +\delta$ then
$ L(w',w) = L(w+\delta,w) \approx \delta ^T F(w) \delta $ where $F(w)$ is the
Hessian of $L$:
\eq{eq:HessD} F(w) =
\left[\frac{\del^2 L(w+\delta,w) }{\del \delta^2}\right]_{\delta=0} .
\en
The matrix $F(w)$ captures the quadratic approximation of the distance
$L(w+\delta,w)$ and is the Fisher matrix in the particular case to be discussed
next.
Minimizing the quadratic approximation 
$L(y,f(w',x)) + \frac{1}{ 2 \eta} \delta^T F(w) \delta$ to \nref{eq:NGwh3}
yields the iterate: 
\eq{eq:NatGradIt} \hat w = w - \eta F(w) \inv \frac{\del  L(y,f(x,w))}{\del w} .
\en
Therefore, the natural gradient direction for this choice of metric is:
$F(w) \inv \nabla_w  L(y,f(x,w)) $.

Next we will consider the particular situation where $L$ is replaced by
the Kullback-Leiber divergence in the `batch-learning mode' where optimization
is performed in a classical probabilistic manner by exploiting a large data set
as opposed to a single pair $(x,y)$.
We assume that the input vectors $x$ are drawn independently from a
distribution $Q_x$ with density $q(x)$ and the corresponding (target) outputs
$y$ from a conditional target distribution $Q_{y|x}$  with density function $q(y|x)$.
The goal is to minimize of the KL divergence from the target
joint distribution $Q_{x,y}$ to the learned distribution $P_{x,y}(w)$.
A simple argument \cite{martens20} shows that this divergence is equal to: 
\eq{eq:KLd1}
\EE_{Q_x} \left[KL (Q_{y|x} || P_{y|x}(w)) \right] . 
\en

In practice, the idealized expectation with respect to $Q_x$ in \nref{eq:KLd1}
may not be available or it may be difficult to compute. It is therefore replaced by
its \emph{empirical} version $E_{\hat Q_x}$ leading to:
\eq{eq:KLd2}
\EE_{\hat Q_x} \left[KL (\hat Q_{y|x} || P_{y|x}(w)) \right]
= - \frac{1}{|S|} \sum_{(x,y) \in \ S} \log p(y|x,w) ,
\en
where $p(y|x,w)$ is the output of the model for the pair $(x,y)$ and
the equality is up to a non-significant constant.
Here $S$ is the training sample. 

Consider the more general situation where  the cost function $L(y,f(x,w)) $
has been put into  the form of a probability, which we write as
$L(y,f(x,w)) = p_w(y|x)$, the probability of predicting $y$ from input $x$,
with the parameter $w$.
Let $p_w(y|x), p_{w'} (y|x)$ two distributions where we write $w'=w+\delta$.
We measure the closeness of the second distribution from the first by
comparing their predictive distributions using the Kullback-Leibner
divergence:
\eq{eq:DKL}
  D_{KL} (p_w (y|x), p_{w'}(y|x))
  = \EE_{p_w(x)} \left[\log \frac{p_w (y|x)}{p_{w'} (y|x)} \right]
  = \EE_{p_w(x)} \left[\log {p_w (y|x)} - \log {p_{w'}(y|x)}  \right] . 
\en
Writing as before $w'=w+\delta$, we now approximate this divergence using
a second order Taylor series expansion:
\eq{eq:DKL1}
  D_{KL} (p_w (y|x), p_{w'}(y|x))
  \approx  - \delta^T \EE \left[\nabla_w \log {p_w (y|x)}\right]
  - \frac{1}{2} \delta^T  \EE \left[\nabla^2_w \log {p_w (y|x)} \right] \delta  . 
\en
The first  term, $\nabla \log {p_w (y|x)}$, is the derivative of the log-likelihood,
known as \emph{score}, and its expectation is easily shown to be zero
\cite[Lemma-3.3.1]{murphyBook_advanced} and so we end up
with
\eq{eq:DKL3}
D_{KL} (p_w (y|x), p_{w'}(y|x)) =
- \frac{1}{2} \delta^T  \EE \left[\nabla^2_w \log {p_w (y|x)} \right] \delta .
\en
Going back to Equations~\nref{eq:NGwh3} and
\nref{eq:HessD} suggests that this provides the desired metric.
The matrix $F(w) = E \left[\nabla^2_w \log {p_w (y|x)} \right]$ is known as the
\emph{Fisher Information Matrix} or just `the Fisher'.

A useful expression for the Fisher matrix which can be
shown~\cite{murphyBook_advanced} is the following:
\eq{eq:Fisher}
F(w) = \EE \left[ \nabla_w \log {p_w (y|x)} \nabla_w \log {p_w (y|x)}^T \right] . 
  \en
  This important expression indicates that $F(w)$ 
 is simply the mean of the rank-one matrices $g g^T$ where $g$ is the gradient
  of the loss function $ \log {p_w (y|x)}^T $.

  \medskip\paragraph{Example}
  We illustrate the advantage of using Fisher metrics with
  a 2-dimensional example discussed in \cite{sohlNG12}, see also
  \cite[6.4.3]{murphyBook_advanced}.
  Consider a 2D Gaussian distribution of the form:
  \eq{eq:FishExp}
  q(x;w) = \frac{1}{2 \pi } \exp \left[ -\half \| x - A w\|_2^2 \right] \quad
  \text{with} \quad A = \begin{bmatrix} \tau & \frac{1}{\tau} \\ \frac{1}{\tau} & 0
  \end{bmatrix} . 
  \en 
  Note that $x = [x_1, x_2]^T, w = [w_1, w_2]^T$ and the mean of $x$ for this
  distribution is  $A w$, while its covariance matrix is the identity.
  We set $\tau = 2$ (in contrast to \cite{sohlNG12} where $\tau \equiv  3$).
  The problem is poorly parameterized due to the nature of the transformation
  $A$ which tends to amplify the first coordinate relative to the second. 
  Assume that the cost function is the log-likelyhood of $q(x;w)$
  under some observed data distribution $p(x)$.
  \eq{eq:FisherCost}
  L(w) = -\EE_{p(x)}  [ \log q(x;w) ] . 
  \en
  Incidentally, since \[
    D_{KL} (p || q) = \int p(x)  \log \frac{p(x)}{q(x;w)} \ dx
    =  - \int p(x)  \log q(x;w) \ dx
    + \int p(x)  \log p(x) \ dx ,
  \]
 we see that  $L(w)$ is nothing but the KL divergence $D_{KL} (p||q) + c$
  where $c$ is a constant.

\begin{figure}[tb]
    \begin{minipage}[b]{0.4\textwidth}
      \includegraphics[width=0.90\textwidth]{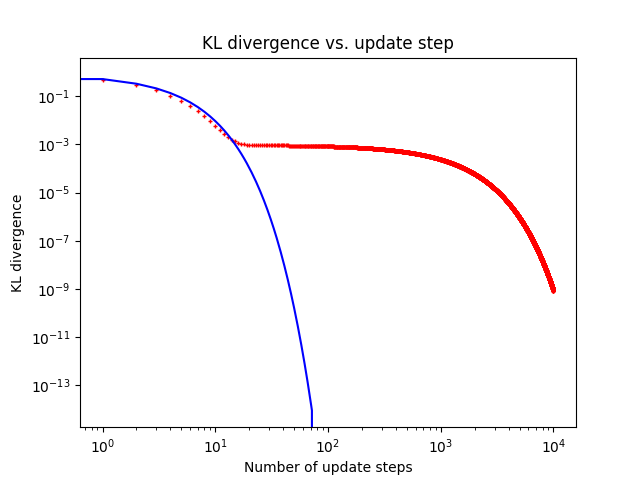}
      \vfill
      \bigskip        \bigskip        \bigskip 
      \mbox{  \ }  
    \end{minipage}
    \begin{minipage}[b]{0.6\textwidth}
      \includegraphics[width=0.99\textwidth]{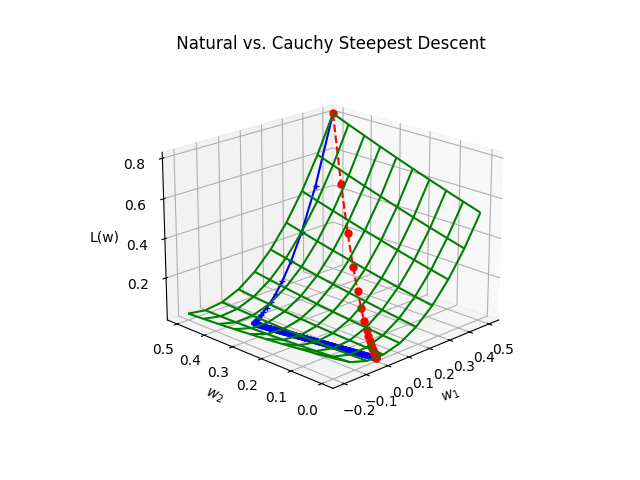}
    \end{minipage}
    \caption{Gradient descent versus natural gradient descent on a two-dimensional Gaussian model with direction-dependent geometry. The left panel shows the decay of the objective $L(w)$ over the iterations, while the right panel displays the corresponding trajectories on the loss surface. Standard gradient descent is slowed by the distorted geometry of the parametrization, whereas natural gradient descent accounts for this geometry through the Fisher metric and converges much more rapidly. Standard (Cauchy) steepest descent is shown by blue plus markers, while natural gradient descent is shown by red circles.}
    \label{fig:NatGD} 
  \end{figure}

  We start at the point $w = [\frac{1}{2}, \frac{1}{2}]$ and perform a
  maximum of 10,000 steps of either gradient descent or natural
  gradient descent -- but stop the iteration when an accuracy of $10^{-12}$
  is reached.
  The left side of Figure~\ref{fig:NatGD} shows the resulting
  values of the objective function as the iterations proceed.
  As can be seen NGD converges in less than 100 iterations.
  In contrast Gradient Descent did not reach the optimal solution (which is
  located at the origin) in 10,000 steps.
  This is rather easy to explain in this case. The right panel of the figure
  illustrates the iterations  in a 3D plot. The $(x,y)$ coordinates are those
  $w_1, w_2$ of the parameters. The vertical coordinate shows the value of the
  objective function. As can be seen, the natural gradient takes immediatly
  the direction toward the solution. The standard descent algorithm takes
  first a direction almost parallel to the $w_1$ axis toward the $w_1 = 0$
  line (vertical axis in 2-D). Once it comes close to  this line it starts a slow descent along a valley toward the origin. At every point the
  gradient is $ g = (A^T A) w $ while the natural gradient is
  $ \tilde g = (A^T A)\inv  g = w$. The direction of the natural gradient is
  the ideal one in this situation: we could converge in exactly one step if
  we took the optimal step size which is equal to $\eta = 1$: 
  \[
    w_1 = w_0 - 1 \times (A^TA)\inv g_0 = w_0 - (A^T A)\inv (A^T A) w_0 = 0 . \]
  
  \subsection{Connection with the Gauss-Newton method}
  In the linear case example shown at the end of the previous section we saw that
  the Hessian equals the matrix $A^T A$ of the normal equations. The general nonlinear
  case leads to a similar observation. 
  In the Gauss-Newton approach, one is interested in solving a system of
  equations $r(w) = 0 $ where $r$ is a mapping from $\RR^n$ to $\RR^m$ by
  a least-squares approach.
  If we call $r_i$, $i=1:m$ the components of $r(w)$ then the least-squares
  method amounts to minimizing the objective function:
  \eq{eq:GN1}
  \phi(w) = \half \| r(w) \|_2^2  = \half \sum_{i=1}^m r_i(w)^2 .
  \en
  The gradient of $\phi$ can be seen to be equal to
  $\nabla \phi(w) = J(w)^T r(w) $ where
  $J(w)$ is the Jacobian of $r$ while its Hessian is
  (\cite[p. 246]{Nocedal-Wright-book}):
  \begin{align} 
    \nabla^2 \phi(w) &=\sum_{i=1}^m \nabla r_i(w) \nabla r_i(w)^T +
                       \sum_{i=1}^m r_i(w) \nabla^2 r_i(w) \label{eq:GN2} \\
    &=   J(w)^T J(w) + \sum_{i=1}^m r_i(w) \nabla^2 r_i(w) . \label{eq:GN3} 
  \end{align} 
  In the linear case where $r(w) = b - Aw$, the second term in
  \nref{eq:GN2} and  \nref{eq:GN3} is zero. This is the situation we
  encountered in the example at the end of Section \ref{sec:NatG} where
  we found that $\nabla^2 \phi = A^T A$. If we want to use a Newton approach
  to minimize \nref{eq:GN1}, we would normally need to solve linear systems
  with the Hessian calculated in \nref{eq:GN3}. This Hessian is expensive
  (and impractical) to compute.
  The Gauss-Newton approach amounts to a Newton method applied to
  \nref{eq:GN1} where the Hessian in \nref{eq:GN3}
  is approximated by the Gauss-Newton matrix
  \eq{eq:GN4} G = J(w)^T J(w) \en in \nref{eq:GN3}.
  Note that when we are near the 
  solution the terms $r_i$ are small and this justifies this approximation.

  A remarkable observation for the case where $r_i(w) = f(x_i,w) - y_i$
  is that the matrix $J(w)^T J(w) $ is nothing but the Fisher as can be seen
  from the first part of the alternative expression \nref{eq:GN2} since 
  in this case  $\nabla r_i (w) \equiv \nabla f(x_i,w) $. 
  In LLMs, objective functions are not usually of the form  \eqref{eq:GN1}.
The expression~\eqref{eq:GN3} can be generalized to 
the more common situation where $\phi(w) = L(y,f(x,w))$
and we are minimizing the empirical expectation:
\eq{eq:GGN1}
  \frac{1}{|S|} \sum_{x,y \, \in \, S} L(y,f(x,w))
\en
where  $S$ is the training sample. An approximation of the expression that extends 
\eqref{eq:GN3} to this case  \cite{martens20}  will then yield  the following
Generalized Gauss-Newton matrix which is the
analogue of \eqref{eq:GN4}:
\eq{eq:GGN4}
  \frac{1}{|S|} \sum_{x, y \, \in \, S} J_f^T H_L J_f 
  \en
  where $J_f$ is the Jacobian of $f(x,y)$ with respect to the parameters $w$,
  evaluated at the sample  $(x,y)$ and $H_L$ is the Hessian of $L(x,z)$ with respect
  to $z$ evaluated at $f(x,y)$ for sample $(x,y)$.
  Details can be   found in \cite{martens20}. Note that the particular case 
  $L(x,z) = z-y$ corresponds to the situation when $r_i(w) = f(x_i,w)-y_i$ mentioned
  above and in this case the Hessian $H_L$ becomes the identity matrix. 
  The GGN matrix is not always equal to the Fisher matrix but it is for
  the broad class of objective functions that belong to the `exponential' family
  of models~\cite{martens20}.

  \subsection{The K-FAC approach}\label{sec:KFAC}
  The  `Kronecker-factored Approximate Curvature' (K-FAC) preconditioner introduced in \cite{KFAC-paper15}
  exploits the layer structure of standard machine learning models to define an approximation
  of the Fisher matrix which is then used as a preconditioner. 
  For a loss function $L(y,f(x,\theta))= -\log p(y|x,\theta)$ the authors define the notation:
  \eq{eq:delL}
  \calD v = \frac{d L(y,f(x,\theta))}{d v} = - \frac{d \log p(y|x,\theta))}{d v} .
  \en
  Here $\theta$ is the vector of all parameters of the model. 
  The variable $v$ in the above equation is any set of variables that occurs in
  the computational graph - i.e., those  variables we called $z_i, a_i$ in
  Section~\ref{sec:BProp}.
  It is important to note that $ \calD \theta$ is just the desired gradient of $L$ with respect
  to $\theta$. This gradient is computed
  recursively from the back-propagation algorithm using subvariables $v$ encountered in the
  computational graph.   If we call $W_1, W_2, \cdots, W_L$ the parameters at
  each of the $L$ layers, then it can be seen that
  \[\calD \theta = \left[\vvec (\calD W_1)^T, \vvec (\calD W_2)^T,\cdots, \vvec (\calD W_L)^T \right]^T  .
    \] 
    The whole Fisher matrix is $\EE [\calD  \theta  \ \calD \theta^T]$ and this is
    too costly to compute. However, we can approximate it by exploiting the specific structure of the
    parameters. In the particular case of MLP, the authors rewrite \nref{eq:zl} as:
    \begin{align}
      s_i & = W_i \bar z_{i-1}  \label{eq:Kfac-Si} \\
      z_i & = \sigma(s_i)  \label{eq:Kfac-Zi}
    \end{align}
    where the notation has changed as follows. First the matrix $W_i$ is transposed. Then
     homogeneous coordinates are used: 
    the bias $b_i$ is lumped together with $W_i$  by appending it as its last column and the vector
    $z_i$ now has  appended to it  an additional entry with value 1. 
    In the back-propagation algorithm, when we traverse layer $i$, we
    first compute
    \eq{eq:calDsi}
    g_i \equiv \calD s_i = \calD z_i \odot \sigma'(z_i)
    \en
    from which we get by using (\ref{eq:Kfac-Si}--\ref{eq:Kfac-Zi}): 
    \eq{eq:calDwz}
    \calD W_i = g_i \bar z_{i-1}^T \ ,\qquad \calD z_{i-1} = W_i^T g_i . 
    \en
    We are interested in the form of each  $\calD W_i$ which is  $\calD W_i = g_i \bar z_{i-1}^T $.
    The Fisher matrix has a block structure and its 
     $(i,j)$-th  block Fisher involves taking the expectation of the small matrices: 
    \[
      \tilde f_{ij} = \vvec (\calD W_i) \vvec (\calD W_j)^T = \vvec ( g_i \bar z_{i-1}^T)) \  \vvec ( g_j\bar z_{j-1}^T))^T . \]
    Observing that $\vvec(u v^T) = v \otimes u$, the Kronecker product of $v$ with $u$, this becomes:
    \[
      \tilde f_{ij} = (\bar z_{i-1} \otimes g_i) \ (\bar z_{j-1} \otimes g_j)^T = (\bar z_{i-1} \bar z_{j-1}^T) \otimes (g_i g_j^T) \]
   thanks to basic rules of Kronecker products~\cite{GVL-book}. 
   Each block $f_{ij} $ of the Fisher matrix is the expectation of the sample-based $\tilde f_{ij}$:
   \eq{eq:fij}
   f_{ij}  =
   \EE [ \tilde f_{ij} ] =  \EE [ (\bar z_{i-1} \bar z_{j-1}^T) \otimes (g_i g_j^T) ] \approx
   \EE [ \bar z_{i-1}  \bar z_{i-1}^T ]  \otimes \EE [g_i g_j^T] . 
   \en
   The approximation made at the end of the above equation is not an equality in general. It is helpful in that it
   permits to approximate $F$ by the Kronecker product of two separate matrices namely,
   $\EE [ \bar z_{i-1}  \bar z_{i-1}^T ] $ and $ \EE [g_i g_j^T] $. As a consequence the inverse of the related Fisher
   matrix approximation is easily obtained using again standard rules or Kronecker products of
   matrices:  $(A \otimes B) \inv = A\inv \otimes B\inv$.

  \subsection{The Shampoo approach}\label{sec:Shampoo} 
  A number of approaches followed the KFAC idea. Among these is a method named
  `Shampoo' in \cite{Shampoo-paper18} which aimed to address a weakness of KFAC,
  namely the dependence of this approach on the particular structure of the
  parameter set. The parameters in MLPs are organized in blocks related to the
  different layers and this is heavily exploited in the K-FAC approach.  In
  contrast, Shampoo is defined without any reference to these layers. Instead it
  needs to be only aware of the tensors involved in the optimization and their
  sizes.

  In the following description we assume that the parameters form a 2-mode
  tensor, i.e., that they are stored in a matrix $W \ \in \ \RR^{m\times n}$. In
  a gradient-descent approach, the parameter set after step $t$ is updated by
  the gradient $G_t = \nabla f_t(W_t)$ where $f_t$ is the loss function, which
  is typically based on a certain batch of samples. Here $G_t$ is an
  $m \times n $ matrix like $W_t$ and it is common to think in terms of
  `flattened' arrays in order to define a suitable preconditioner of size
  $mn \times mn$. Instead of this, Shampoo defines two smaller matrices
  $L_t \in \RR^{m\times m} $ (left side) and $R_t \in \RR^{n\times n}$ (right
  side) whose aim is to encapsulate second-moment information of the
  gradient. In this way the preconditioning operator acts with two small
  matrices one on the left and the other on the right of the current gradient
  $G_t$. Thus, the storage requirement is only $m^2 + n^2$.
  
\begin{algorithm}[htb]
  \centering
  \caption{Shampoo Preconditioning -- matrix case}\label{alg:Shampoo}
  \begin{algorithmic}[1]
    \State  \textbf{Input:} $W_0  = \textbf{0}_{m \times n},
    \quad L_{-1}= \epsilon I_{m\times m},
    \quad R_{-1} = \epsilon I_{n\times n},$
\For{$ t=0, \cdots,$}
\State Evaluate $G_t = \nabla f_t(W_t)$ 
\State Update Preconditioner Matrices:
\begin{align}
  L_t &= L_{t-1} + G_t G_t^T & \label{eq:shampoo1} \\ 
  R_t &= R_{t-1} + G_t^T G_t & \label{eq:shampoo2} 
\end{align}
\State Update Parameters:
\begin{align}
  W_{t+1} & = W_t - \eta L_t^{-1/4} G_t R_t^{-1/4} &  \label{eq:shampoo3} 
\end{align}
\EndFor 
\end{algorithmic}
\end{algorithm} 

A common variation is to precede the matrices $G_t G_t^T$ and $G_t^T G_t $ by scalars
leading to exponentiall moving averages instead of simple sums.
There is a striking resemblance between the updates (\ref{eq:shampoo1} --
\ref{eq:shampoo3}) in Shampoo and those of \nref{eq:ADAg} and
\nref{eq:ADAg1} of the ADAGrad optimizer discussed in
Section~\ref{sec:adagrad}. Shampoo can be seen as a generalization of ADAgrad.
ADAgrad employs the diagonal preconditioner \eq{eq:AdagP} D = \text{Diag}
\left[\epsilon + \sum_{i=0}^t g_i^2 \right]^{-1/2} \en where $g_i$ is the
flattened version of $G_i$ and we recall that all operations are component-wise.
Instead of an the $mn \times mn$ diagonal matrix $D$ of \nref{eq:AdagP} Shampoo
accumulates the left-sided (dense) matrices $G_i G_i^T$ and the right (dense)
matrices $G_i^T G_i$. It then scales the gradient $G_t$ by the accumulated
left-sided matrix to the power $-1/4$ on the left and by the accumulated
right-sided matrix to the power $-1/4$ on the right.

The matrices with the negative and fractional powers in (\ref{eq:shampoo1}
-- \ref{eq:shampoo3}) are not  expensive to compute because the corresponding 
operations are performed on relatively small matrices.  The authors of the
article show that their algorithm can converge considerably faster than commonly
used optimizers in standard deep learning models and that in spite of the
seemingly more complex update rule, the achieved runtume is much smaller than
that of simple methods such as vanilla SGD.
The superiority of Shampoo over AdamW has been demonstrated  as a distributed version of the algorithm
won  the 2024 \texttt{AlgoPerf}
Training Algorithms Competition
\footnote{See: \url{https://mlcommons.org/2024/08/mlc-algoperf-benchmark-competition/}.
  The announcement of the result stated:
  ``Non-diagonal preconditioning has dethroned Nesterov Adam, ...''.}. 

\subsection{Muon} \label{sec:MoenSoap} 
The Muon approach provides an interesting illustration of the use of advanced
Linear Algebra methods for developing effective optimization
algorithms in machine learning. The method was derived as an improvement over 
Shampoo with the specific aim to avoid its main drawback, namely its use of two scaling
matrices instead of one.

\begin{algorithm}[htb]
  \centering
  \caption{Muon Preconditioning -- matrix case}\label{alg:Muon}
  \begin{algorithmic}[1]
    \State  \textbf{Input:} $W_0  = \textbf{0}_{m \times n}$, $M_0 = 0$ 
\For{$ t=1,2, \cdots,$}
\State Evaluate $G_t = \nabla f_t(W_t)$ 
\State $M_{t} = G_t + \mu M_{t-1} $
\State $O_{t} = \text{Newton-Schulz} (M_{t}) $
\State $W_{t+1} =  W_t - \eta_t O_{t} $ 
\EndFor 
\end{algorithmic}
\end{algorithm}

If we were to set $O_t = M_t$ in Line~5, the result would be nothing but a form
of gradient descent where the gradient is replaced by its average or momentum $M_t$.
The Newton-Schultz operation aims at scaling $M_t$ by approximating the actions
of $L_t^{-1/4}$ and $R_t^{-1/4}$ on the left and right of $G_t$ in equation
\eqref{eq:shampoo3} of Shampoo.
In the simplest situation when $G_t$ is square and $L_t =  G_t G_t^T$, 
$ R_t = G_t^T G_t$  then the matrix
$L_t^{-1/4} G_t R_t^{-1/4}$ becomes the semi-orthogonal matrix $U V^T$ where
$G_t = U \Sigma V^T$ is the SVD of $G_t$. It does appear that computing
$U V^T$ via the fractional powers of the matrices $ G_t G_t^T$ and
$G_t^T G_t $ is far from optimal both in terms of memory and
computations. The Muon approach approximates this
matrix indirectly by an iterative procedure - except that now
$U, V$ are the left and right factors of the the SVD of $M_t$, not $G_t$.
The iterative procedure named Newton-Schulz is a form of polynomial iteration.
Starting with $X_0 = M_t/\|M_t\|_F$, we define the sequence of matrices:
\eq{eq:muonIt}
X_{k+1} = a X_k + b (X_k X_k^T)X_k + c (X_k X_k^T)^2 X_k
\en
with $a = 3.4445, b = -4.7750, c = 2.0315 $.

\section{Concluding remarks} 

We begin our conclusion by circling back to the exceptionally fast pace of innovation
in AI mentioned in our introduction.
As we were writing this paper we were often confronted with  a rapid phase of
proliferation of ideas  related to a certain theme, e.g., `efficient transformers',
which made it difficult to review interesting developments in an exhaustive manner.
As an indication of the vigorous nature of the field, consider that the number of
articles submitted to Neurips, a major conference in AI, has evolved as follows in the years
2021 to 2025: 9,122, 10,411, 12,345, 15,671, 21,575.
The number for 2025 was a major jump (of 40.6\%) over the previous year which
itself saw a jump of $\approx 27$\% over the year before it.  Thus, there has
been a signigicant recent acceleration of the number of articles submitted
leading one to wonder if this pace is sustainable\footnote{See the site
  https://www.ctol.digital/news/ai-research-summit-neurips-2025-receives-record-breaking-27000-paper-submissions/
  for information. The article in this site quoted a member of the academic
  community stating that with the current growth of 26.3\%, we could in theory
  see one submission for each person on earth in 59 years.}.  In 2025, over 5,290
papers were accepted to NeurIPS, alongside thousands of additional publications
from other AI research venues.
This makes it extremely challenging to keep abreast of new
developments.  In particular, some of the topics discussed in survey articles
such as this one become quickly obsolete.


As a second remark, we want to emphasize the main difference between research in
AI and in classical Numerical Linear Algebra.  As was seen, statistical
intuition and experience is important in developing new methods, more so than in
traditional approaches.  This is because methods founded on rigorous
mathematical principles alone do not necessarily work due to the exceedingly
complex nature of the problem at hand. A new contribution would typically entail
having some intuition on a method that may work followed by a great deal of
testing. These tests are often conducted on big parallel machines. The
combination of big teams and big machines, is mandatory for truly impactful
innovations such as those that lead to the idea of Transformers for example.
This puts traditional NLA teams at a disadvantage.


In spite of these considerations, it is essential for NLA specialists to get involved  in AI
research. As we have tried to show, many of the truly innovative ideas, especially those that deal
with `efficient variants' utilize mainly NLA concepts. It would be infortunate for NLA specialists
to keep working exclusively on traditional topics while exciting developments in AI, many of
which are based on NLA ideas, are shaping a new chapter in science. It is true that a certain
amount of learning is needed before starting to participate, but one must realize that there are
mitigating factors. For example learning is greatly facilitated by the availability of a large
number of resources and with the help of AI itself. 

\paragraph{Acknowledgment.} The authors are indebted to an anonymous referee for
her/his very careful reading of an earlier version of this manuscript.

\medskip
\bibliographystyle{siam}  
\bibliography{biblioTransformers,YSbib}
\end{document}